\documentclass[twoside,11pt]{article}

\usepackage{import} 

\usepackage{thmtools}
\usepackage{thm-restate}

\usepackage{amsmath}

\usepackage{jmlr2e} 
\usepackage{cleveref}
\usepackage{soul,color} 
\usepackage[english]{babel} 

\ShortHeadings{Unifying Attribution-Based Explanations Using Functional Decomposition}{Gevaert, Saeys}
\firstpageno{1}

\begin{document}

\title{Unifying Attribution-Based Explanations\\Using Functional Decomposition}

\author{\name Arne Gevaert \email arne.gevaert@ugent.be \\
	\addr Department of Applied Mathematics, Computer Science and Statistics\\
	Ghent University\\
	Data mining and Modeling for Biomedicine (DaMBi)\\
	VIB Inflammation Research Center\\
	Technologiepark-Zwijnaarde 71, 9052 Ghent, Belgium
	\AND
	\name Yvan Saeys \email yvan.saeys@ugent.be \\
	\addr Department of Applied Mathematics, Computer Science and Statistics\\
	Ghent University\\
	Data mining and Modeling for Biomedicine (DaMBi)\\
	VIB Inflammation Research Center\\
	Technologiepark-Zwijnaarde 71, 9052 Ghent, Belgium}

\editor{}

\maketitle

\begin{abstract}
	The black box problem in machine learning
	has led to the introduction of an ever-increasing set of explanation methods for complex models.
	These explanations have different properties,
	which in turn has led to the problem of \textit{method selection:}
	which explanation method is most suitable for a given use case?
	In this work, we propose a \textit{unifying framework} of attribution-based explanation methods,
	which provides a step towards a rigorous study of the similarities and differences of explanations.
	We first introduce \textit{removal-based attribution methods} (RBAMs),
	and show that an extensively broad selection of existing methods can be viewed as such RBAMs.
	We then introduce the \textit{canonical additive decomposition} (CAD).
	This is a general construction for additively decomposing any function based on the central idea of removing (groups of) features.
	We proceed to show that indeed every valid additive decomposition is an instance of the CAD,
	and that any removal-based attribution method is associated with a specific CAD.
	Next, we show that \textit{any} removal-based attribution method
	can be completely defined as a game-theoretic value or interaction index
	for a specific (possibly constant-shifted) cooperative game,
	which is defined using the corresponding CAD of the method.
	We then use this intrinsic connection to define formal descriptions of specific behaviours of explanation methods,
	which we also call \textit{functional axioms,}
	and identify sufficient conditions on the corresponding CAD and game-theoretic value or interaction index of an attribution method
	under which the attribution method is guaranteed to adhere to these functional axioms.
	Finally, we show how this unifying framework can be used to develop new, efficient approximations for existing explanation methods.
\end{abstract}


\section{Introduction}
In recent years, a vast number of explanation methods has been proposed
in an attempt to tackle the \textit{black box problem} \citep{arrieta2019,rudin2019,molnar2022},
\textit{i.e.}~the problem that many popular machine learning models are far too large and/or complex
for humans to interpret.
This black box problem leads to issues in trust \citep{tonekaboni2019,holzinger2021c}, debugging \citep{sculley2015} and deployment of machine learning models.
Many of the proposed explanation methods provide \textit{feature attributions},
which is an explanation in the form of an attribution or importance score for each feature.
This score is supposed to reflect how ``important'' or ``influential'' each feature is for a given model.

With the introduction of such a diverse range of attribution methods, a new problem arises:
which one of the many available methods is the most appropriate for a given use case?
This question has led researchers to develop a wide range of metrics designed to measure the quality of a given explanation or explanation method 
\citep{hedstrom2023,hedstrom2023b,ancona2018,yeh2019}.
However, recent work has shown that these quality metrics disagree about which explanation or method is the best one \citep{tomsett2020,gevaert2024}.
Another approach that has recently been explored is the so-called \textit{axiomatic approach},
in which a set of desirable properties or \textit{axioms} is assumed,
and an explanation method is defined such that it satisfies these desirable properties \citep{lundberg2017}.
However, research has shown that multiple methods can often satisfy the same set of axioms while still contradicting each other in practice,
even if the axioms are supposed to ``uniquely define'' a specific method in theory \citep{kumar2020,sundararajan2020}.
In conclusion, objective selection and/or evaluation of feature attribution-based explanation methods remains an open problem.
These developments highlight the need for a unifying framework to understand the differences between methods.
In this work, we introduce such a framework, based on the core concepts of cooperative game theory and additive functional decomposition.

First, we formally define the scope of our unifying framework in the form of \textit{removal-based attribution methods} (RBAM).
These methods are completely defined by three formal, mathematical choices.
The defining choices of a RBAM correspond to 1) the behaviour of the model that is explained,
2) how features are removed from the model,
and 3) how the behaviour of the model after features are removed is summarized into an explanation.
We show that a significant number of existing methods are indeed instances of RBAMs,
including Shapley value-based methods \citep{lundberg2017,lundberg2019,merrick2020,sundararajan2020},
permutation-based methods \citep{breiman2001a,strobl2008,zeiler2014},
higher-order attribution methods \citep{sundararajan2020a},
variance-based explanations \citep{sobol2001,song2016} and others \citep{ribeiro2016}.
A non-exhaustive overview of methods that can be viewed as RBAMs is given in \Cref{tbl:fdfi-shapley-methods,tbl:fdfi-non-shapley-methods}.

\begin{table}
	\centering
	\begin{tabular}{||l || r | r |}
		\hline
		Method                & Behaviour & Removal \\ [0.5ex]
		\hline\hline
		IME-Retrain \citep{strumbelj2009} & Local output   & Retrain   \\
		IME-Marginalize \citep{strumbelj2010} & Local output & Uniform \\
		QII \citep{datta2016} & Local output & PM \\
		Conditional SHAP \citep{lundberg2017} & Local output & Conditional \\
		KernelSHAP \citep{lundberg2017} & Local output & Marginal \\
		TreeSHAP \citep{lundberg2019} & Local output & TD \\
		Causal SHAP \citep{heskes2020} & Local output & Interventional \\
		JBSHAP \citep{yeh2022} & Local output & JBD \\
		RJBSHAP \citep{yeh2022} & Local output & RJBD \\
		LossSHAP \citep{lundberg2019a} & Local loss & Conditional \\
		Shapley Effects \citep{owen2014} & Variance & Conditional \\
		Shapley Net Effects \citep{lipovetsky2001} & Dataset loss & Retraining \\
		SPVIM \citep{williamson2020} & Dataset loss & Retraining \\
		SFIMP \citep{casalicchio2019} & Dataset loss & Marginal \\
		SAGE \citep{covert2020} & Dataset loss & Conditional \\
		\hline
	\end{tabular}
	\caption{Examples of Shapley-based RBAMs and their corresponding removal operators and behaviour mappings.
	All of these methods are covered by the unifying theory proposed in this work.}
	\label{tbl:fdfi-shapley-methods}
\end{table}

\begin{table}
	\centering
	\begin{tabular}{||l || r | r |}
		\hline
		Method                & Behaviour & Removal \\ [0.5ex]
		\hline\hline
		Occlusion \citep{zeiler2014} & Local output & Single baseline   \\
		ASV \citep{frye2020} & Local output & Conditional \\
		PredDiff & Local output & Conditional \\
		CXPlain & Local output & Single baseline \\
		RISE & Local output & Single baseline \\
		BANZHAF \citep{karczmarz2022} & Local output & TD\\
		WeightedSHAP \citep{kwon2022a} & Local output & Conditional \\
		PFI \citep{breiman2001a} & Dataset loss & Marginal \\
		Conditional PFI \citep{strobl2008} & Dataset loss & Conditional \\
		LOCO \citep{kohavi1997} & Dataset loss & Retraining \\
		Univariate Predictors \citep{guyon2003} & Dataset loss & Retraining \\
		\hline
	\end{tabular}
	\caption{Examples of RBAMs that use aggregation coefficients other than the Shapley aggregation coefficients,
	and their corresponding behaviour and removal mappings.
	All of these methods are covered by the unifying theory proposed in this work.}
	\label{tbl:fdfi-non-shapley-methods}
\end{table}

\begin{figure}
	\centering
	\includegraphics[width=\textwidth]{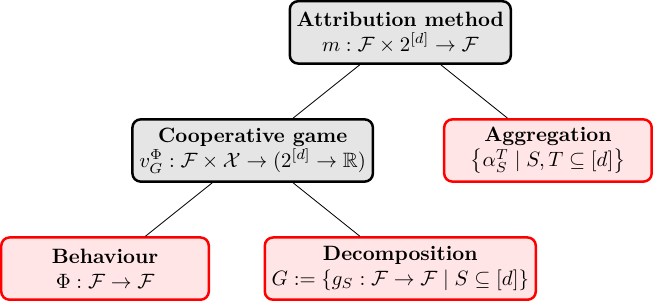}
	\caption{Summary of the unifying framework.
		Red nodes correspond to choices that need to be made by the user.
		Under mild conditions,
		any removal-based attribution method is fully determined by a choice of behaviour,
		functional decomposition,
		and a set of aggregation coefficients.
		The behaviour and decomposition together define a specific cooperative game
		for which the attribution method is a value or interaction index.}
	\label{fig:fdfi-three-choices}
\end{figure}

Next, we introduce the \textit{canonical additive decomposition} (CAD).
This is a general construction for additively decomposing any function
based on the central idea of removing groups of features.
We proceed to show that indeed every valid additive decomposition method is in fact an instance of the CAD.
We then establish a connection between the CAD and cooperative game theory
by showing that an additive decomposition can be used to construct a \textit{pointwise cooperative game.}
Finally, we show that
\textit{any removal-based attribution method 
        can be completely defined as a game-theoretic value or interaction index
	for a specific (possibly constant-shifted) cooperative game.}
This game-theoretic index is in turn completely defined by a set of \textit{aggregation coefficients.}
In summary, we are able to reduce the definition of a large set of attribution-based explanation methods
down to three formal, mathematical choices:
\begin{enumerate}
	\item The behaviour of the model $\Phi(f)$ to be explained.
	\item The manner in which $f$ is additively decomposed into a set of functions $\{ g_{S}(f) \mid S \subseteq [d] \}$.
	\item The constants $\{ \alpha_{S}^{T} \mid S,T \subseteq [d] \}$ that define a linear combination of the values $v(S)$,
	      where $v$ is a cooperative game generated by the behaviour and decomposition.
\end{enumerate}
A summary is given in \Cref{fig:fdfi-three-choices}.

This result has several implications.
First, it implies that \textit{any} removal-based attribution method
can be viewed as a game-theoretic explanation of a pointwise cooperative game.
This is surprising,
as the result also holds for various methods that were designed heuristically,
without any intended link to cooperative game theory.
Second, it implies that formal properties of explanation methods
can be studied through their corresponding additive decomposition.
This allows for a more rigorous and broad investigation of the behaviour of explanation methods,
whereas such investigations have happened mainly heuristically and on a case by case basis up to this point \citep{kumar2020,sundararajan2020,merrick2020}.
Finally, these results open up a clear path towards computationally efficient approximations for explanation methods.

The rest of this work is organized as follows.
In \Cref{sec:fdfi-related-work} we give an overview of existing attempts at 
unifying explanation methods
and links between explanation methods and additive decomposition.
\Cref{sec:fdfi-notation} provides a short overview of mathematical notation
and some basic concepts that will be used throughout this work.
The main contributions are described in \Cref{sec:removal-based-attribution,sec:additive-functional-decomposition,sec:unifying-framework}.
In \Cref{sec:removal-based-attribution} we formally define removal-based attribution methods.
\Cref{sec:additive-functional-decomposition} then introduces the canonical additive decomposition.
Next, the ideas from Sections \Cref{sec:removal-based-attribution,sec:additive-functional-decomposition}
are used to construct a general but formal unifying framework of
removal-based attribution methods in \Cref{sec:unifying-framework}.
We conclude with a discussion of the implications of our framework
and interesting directions for future research in \Cref{sec:conclusions-future-work}.
\Cref{sec:proofs} contains proofs of any propositions and/or theorems
for which the proof is not given immediately in the main text.

\section{Related Work}
\label{sec:fdfi-related-work}
Some attempts have already been made to provide unifying theories of attribution-based explanations and explanations in general.
\cite{merrick2020} introduce the \textit{Formulate, Approximate, Explain} framework, which is based on the specification of a reference distribution and the approximation of Shapley values using this distribution to remove features.
The authors show that a number of existing explanation techniques can be viewed as specific instances of this framework.
However, this framework only covers Shapley value-based explanations, and is unable to capture all existing Shapley value-based techniques, such as conditional Shapley values \citep{lundberg2017}
or causal Shapley values \citep{heskes2020}.

A more comprehensive framework is given by the \textit{Explaining by Removing} (XBR) framework \citep{covert2021}, which covers a wide range of removal-based explanation techniques.
Indeed, this framework introduces the three choices that our framework is also based on.
However, by focusing on attribution-based explanations specifically,
we are able to formulate the three choices in a more mathematically rigorous and precise way.
This allows us to perform a much more detailed theoretical analysis of the behaviour of explanation methods.
We do this by formally defining a set of \textit{functional axioms},
which are mathematically precise and intuitive guarantees of behaviour for removal-based attribution methods.
These functional axioms are designed to address known problems with the more commonly used \textit{game-theoretic axioms,}
which have been shown to be unintuitive as descriptors of method behaviour in practice \citep{chen2020,kumar2020}.
We then derive provably sufficient conditions for explanation methods to adhere to these functional axioms.
We also provide a formal link between removal-based explanations and additive functional decomposition,
which allows us to describe properties of attribution methods in terms of properties of additive decompositions.
This opens the path to further research by linking the existing literature on additive functional decomposition
to explainable machine learning, and vice versa.
Finally, we also extend the framework to cover higher-order attribution methods
such as the Shapley-Taylor interaction index \citep{sundararajan2020}.

Recent work has also uncovered a number of links between attribution-based explanation methods and additive functional decompositions.
\citet{owen2014} shows a connection between \textit{Shapley effects} and Sobol' indices \citep{sobol2001} of the functional ANOVA decomposition \citep{roosen1995}.
We demonstrate that a similar link exists for a more general class of Shapley-based methods.
\cite{hiabu2023} illustrate that SHAP and q-interaction SHAP \citep{tsai2023} can be linked in a similar fashion to partial dependence plots.
This link can be used to provide richer explanations and to perform model debugging and model editing.
\cite{bordt2023} introduce a bijective relation between additive functional decompositions and Shapley value-based explanations.
They use this relation to show an equivalence between the Shapley-Taylor interaction index, Faith-SHAP \citep{tsai2023} and their proposed n-Shapley values.
Finally, \cite{herren2022} also demonstrate how the functional ANOVA decomposition can be used to compute SHAP values, and use this link to show how numerical tests for feature interactions can be used to select coalitions in Shapley value sampling.

All of these works prove that there is a strong link between Shapley value-based techniques and additive functional decomposition.
In this work, we extend this link beyond Shapley value-based techniques,
and demonstrate that \textit{any} removal-based univariate or multivariate attribution method can be linked to additive functional decomposition in a similar fashion.

\section{Notation}
\label{sec:fdfi-notation}
In this section, we introduce some of the notation we will use in this work.
We will denote sets using the uppercase letters $S,T,U,V$, and the complement of a set $S$ as $\overline{S}$.
Random variables will be denoted using the uppercase letters $X,Y,Z$.
If a set has only one element $\{i\}$, we will declutter notation by denoting it simply as $i$ if it is clear from context that this should be a set.
For example: $S \cup i := S \cup \{i\}, S \setminus i := S \setminus \{i\}$.
For a given $d \in \mathbb{N}$, we introduce the shorthand notation $[d] := \{1, \dots, d\}$.

We will denote vectors using boldface letters $\mathbf{x} := ( x_1, \dots, x_d )$.
If $S \subseteq [d]$, we will use the notation $\mathbf{x}_S$ to signify the vector made by the elements of $\mathbf{x}$ that correspond to elements in $S$: $\mathbf{x}_S := ( x_i \mid i \in S )$.
For example, if $\mathbf{x} = (1,4,2,5,3)$, then $\mathbf{x}_{\{ 1,3,5 \}} = (1,2,3)$.
For two vectors $\mathbf{x}, \mathbf{y} \in \mathbb{R}^d$ and a subset $S \subseteq [d]$, we will denote the vector $( \mathbf{x}_S,\mathbf{y}_{\overline{S}} )$ as the vector constructed by combining $\mathbf{x}_S$ and $\mathbf{y}_{\overline{S}}$:
\begin{equation*}
	( \mathbf{x}_S,\mathbf{y}_{\overline{S}} )_i := \left\{
	\begin{array}{ll}
		x_i & \mbox{if } i \in S    \\
		y_i & \mbox{if } i \notin S
	\end{array}
	\right.
\end{equation*}
Let $f \in \mathcal{F}: \mathcal{X} \rightarrow \mathbb{R}$ be a function, where $\mathcal{X} \subseteq \mathbb{R}^d$.
We denote the $i$-th input variable of $f$ as $X_i$.
We will say that $X_i$ is an \textit{independent variable} of $f$ if
\begin{equation*}
	\mathbf{x}_{[d]\setminus\{i\}} = \mathbf{y}_{[d]\setminus\{i\}} \implies f(\mathbf{x}) = f(\mathbf{y}) \qquad \forall \mathbf{x}, \mathbf{y} \in \mathcal{X}
\end{equation*}
If the value of an independent variable $X_{i}$ of $f$ is changed without changing any of the other variables,
then this has no influence on the output of $f$.
For a set $S \subseteq [d]$, we will say that $f$ is \textit{independent of} $X_S$
if for each $i \in S$, $X_{i}$ is an independent variable of $f$.
$X_i$ is an \textit{additive variable} of $f$, or $f$ is \textit{additive in} $X_i$,
if there exist functions $g, h \in \mathcal{F}$ such that $f = g + h$,
where $g$ is independent of $X_{[d] \setminus i}$ and $h$ is independent of $X_{i}$.
Equivalently, $f$ is additive in $X_i$ 
if $f$ can be written as the sum of a univariate function of $X_i$
and a multivariate function that is independent of $X_i$.
This can be seen by defining $g': \mathcal{X}_{i} \rightarrow \mathbb{R}: x_{i} \mapsto g(x_{i},\mathbf{z}_{[d] \setminus i})$
for some arbitrary $\mathbf{z} \in \mathcal{X}$,
where $\mathcal{X}$ is the domain of $f$.
This is a univariate function that is independent of the choice of $\mathbf{z}$.

We will denote the set of permutations of a given set $A$, \textit{i.e.}~bijections of $A$ onto itself, as $\Pi(A)$.
For a given permutation $\pi \in \Pi(A)$, we will write $j \prec_\pi i$ if $j$ precedes $i$ in $\pi$.
Analogously, we write $j \preceq_\pi i$ if $j$ precedes $i$ in $\pi$ or $j$ is equal to $i$.
The set of predecessors of a given element $i$ in a permutation $\pi \in \Pi(A)$ will be denoted as $\text{pred}(i,\pi) := \{j \in A: j \prec_\pi i\}$.
For a given permutation $\pi \in \Pi([d])$, subset $A \subseteq [d]$, vector $\mathbf{x} \in \mathbb{R}^d$ and function $f: \mathbb{R}^d \rightarrow \mathbb{R}$,
we define the permuted subset, vector and function respectively as follows:
\begin{align*}
	\pi \mathbf{x}          & = ( x_{\pi(i)} | i = 1,\dots d ) \\
	\pi S                   & = \{\pi(i) | i \in S\}    \\
	(\pi f)(\pi \mathbf{x}) & = f(\mathbf{x})                  \\
\end{align*}
A function $f$ is called \textit{symmetric} if it is invariant to the order of its arguments:
\begin{equation*}
	\forall \pi \in \Pi([d]), \mathbf{x} \in \mathcal{X}: f(\pi \mathbf{x}) = f(\mathbf{x})
\end{equation*}
A subset of variables $X_{i}, i \in S \subseteq [d]$ is called
\textit{symmetric in $f$}
if $f$ is invariant to reorderings of the variables in $S$,
\textit{i.e.}~if
\begin{equation*}
	\forall \pi \in \Pi([d]), \mathbf{x} \in \mathcal{X}:
	(\forall i \in \overline{S}: \pi(i) = i) \implies f(\pi \mathbf{x}) = f(\mathbf{x})
\end{equation*}
A symmetric function is then simply a function that is symmetric
in all of its arguments.
In multiple proofs in this work, we will also use the following theorem:
\begin{theorem}[Inclusion-Exclusion Principle]
	Let $f$ and $g$ be two functions $2^{S} \rightarrow \mathbb{R}$ defined on subsets of a finite set $S$. Then:
	$$
		g(A) = \sum_{B \subseteq A} f(B) \iff f(A) = \sum_{B\subseteq A}(-1)^{|A|-|B|} g(B)
	$$
	\label{thm:fdfi-inclusion-exclusion-principle}
\end{theorem}
Although the principle is usually presented as an implication \citep{graham1995}, it is easy to verify that the reverse implication also holds.

Finally, we define the discrete derivative of a set function $v$ as follows.
\begin{definition}[Discrete derivative]
	Given a set function $v: 2^{[d]} \rightarrow \mathbb{R}$ and finite subsets $S,T \subseteq [d]$, the \textbf{$S$-derivative of $v$ at $T$} is recursively defined as:
	\begin{align*}
		\Delta_\emptyset v(T) & := v(T)                                  \\
		\Delta_i v(T)         & := v(T \cup i) - v(T \setminus i)        \\
		\Delta_S v(T)         & := \Delta_i[\Delta_{S \setminus i} v(T)]
	\end{align*}
\end{definition}
It is easy to prove by induction on $|S|$ that:
\begin{align*}
	\forall S,T \subseteq [d]                          & : \Delta_S v(T) = \Delta_S v(T \setminus S) = \sum_{L \subseteq S}(-1)^{|L|} v((T \cup S) \setminus L) \\
	\forall S \subseteq [d], T \subseteq [d] \setminus S & : \Delta_S v(T) = \sum_{L \subseteq S}(-1)^{|S|-|L|} v(T \cup L)
\end{align*}
Using the inclusion-exclusion principle, we can derive the following equality by setting $T = \emptyset$:
\begin{equation*}
	\forall S \subseteq [d]: \Delta_S v(\emptyset) = \sum_{L \subseteq S}(-1)^{|S|-|L|} v(L)
\end{equation*}
It is then easy to prove by induction that \citep{grabisch2000,fujimoto2006}:
\begin{equation}
	\label{eqn:fdfi-discrete-derivative-emptyset}
	\forall S \subseteq [d], T \subseteq [d] \setminus S: \Delta_S v(T) = \sum_{L \subseteq T}\Delta_{L \cup S}v(\emptyset)
\end{equation}

\section{Game theory}
\label{sec:game-theory}
In this section, we introduce some preliminary concepts from cooperative game theory.
We define cooperative games, values, and indices.
Next, we provide a brief overview of the different axioms related to values and indices.
Based on these axioms, we then introduce a basic taxonomy of game-theoretic values and indices.

Given a finite set of players $N := \{1,\dots,n\}$,
a \textit{cooperative game} is defined by a real-valued function
that assigns a \textit{worth} to each subset of players:
$v: 2^{N} \rightarrow \mathbb{R}$
with the added restriction that $v(\emptyset) = 0$.
The function $v$ is also called the \textit{characteristic function} of the game,
or in other words,
the game is represented in its \textit{characteristic} or \textit{coalitional form}.
In practice, the game is usually identified with the characteristic function $v$,
as $v$ is sufficient to describe all of the dynamics of the game.

A subset of players $S \subseteq N$ is also called a \textit{coalition}.
A cooperative game is called a \textit{game of transferable utility (TU-game)}
if the worth $v$ can be costlessly transferred between players of $N$,
\textit{i.e.}~a given quantity is ``worth'' just as much to one player as it is to any other.
TU-games will be the main object of study in the rest of this work.
Therefore, in the following descriptions we will also speak simply of \textit{games}
to denote cooperative games of transferable utility.

A \textit{value} for the game $v$, also called an \textit{imputation} \citep{shapley1988},
is a vector $\phi(v) \in \mathbb{R}^{n}$
where each entry contains the \textit{value} for a specific player in $N$.
As an example, consider the following game $v_{\text{ex}}$ with $N = \{ 1,2,3 \}$:
\begin{align*}
	v_{\text{ex}}(\emptyset) & = 0 & v_{\text{ex}}(N)         & = 8 \\
	v_{\text{ex}}(\{ 1 \})   & = 1 & v_{\text{ex}}(\{ 1,2 \}) & = 3 \\
	v_{\text{ex}}(\{ 2 \})   & = 2 & v_{\text{ex}}(\{ 2,3 \}) & = 7 \\
	v_{\text{ex}}(\{ 3 \})   & = 4 & v_{\text{ex}}(\{ 1,3 \}) & = 5
\end{align*}
The function $v_{\text{ex}}$ defines the \textit{worth} for each coalition $S \subseteq N$.
The problem of assigning a value to each of the players in $N$
can now be expressed as finding a vector $\phi(v_{\text{ex}}) \in \mathbb{R}^{3}$
that somehow quantifies each player's ``contribution'' to the outcome.
A simple definition for a value might for example be the average worth
of each coalition of which a given player is a member:
\begin{align*}
	\phi(v_{\text{ex}}) & = (\frac{1 + 3 + 5 + 8}{4}, \frac{2 + 3 + 7 + 8}{4}, \frac{4 + 7 + 5 + 8}{4}) \\
	                    & = (4.25, 5, 6)
\end{align*}
Although this is a valid definition for a value, we can easily identify some possible issues.
First of all, the values for the players do not add up to the worth of the total coalition.
This can be a problem in specific contexts:
for example, this implies that this value would not be an appropriate choice if the goal
is to somehow distribute the total produced worth $v_{\text{ex}}(N)$ among the players in $N$.
Another possible issue with this value is more subtle.
Consider player $1$.
If we inspect the worths of the coalitions containing player $1$ more closely,
we can see that adding player $1$ to any coalition $S \subseteq N \setminus \{ 1 \}$
increases the worth of the coalition by exactly $1$:
\begin{align*}
	v_{\text{ex}}(\{ 1 \}) - v_{\text{ex}}(\emptyset) & = v_{\text{ex}}(\{ 1,2 \}) - v_{\text{ex}}(\{ 2 \})     \\
	                                                  & = v_{\text{ex}}(\{ 1,3 \}) - v_{\text{ex}}(\{ 3 \})     \\
	                                                  & = v_{\text{ex}}(\{ 1,2,3 \}) - v_{\text{ex}}(\{ 2,3 \}) \\
	                                                  & = 1
\end{align*}
This can be interpreted as the player $1$ always ``contributing'' a value of $1$,
regardless of the other players in the coalition.
In other words, player $1$ does not really ``cooperate'' with the other players
in a meaningful way.
Now assume player $2$ becomes more productive, increasing the worth of all the coalitions
containing player $2$ by a value of $3$.
In this case, player $1$ still contributes the exact same value of $1$ to each coalition.
However, its value according to $\phi(v_{\text{ex}})$ would have increased significantly.
If the goal of the value is to \textit{fairly} distribute the worth produced by the total coalition
among the players according to their contributions,
then we could expect some protest from player $2$ if this definition would be used:
player $1$ is essentially ``freeloading'' off the extra productivity provided by player $2$.
This example illustrates that we need some way to define what it means for a value
to be ``fair''.
As we will see later in this section, this is typically done
by defining a set of properties (also called \textit{axioms}) that a value must adhere to.
For example, the requirement that the values add up to the total produced worth $v(N)$
is also called the Efficiency axiom,
and the requirement that the value for a player that has a constant contribution
to each coalition should be equal to that contribution
is also called the Dummy axiom.

We will denote the set of games on a finite set of players $N$ as $\mathcal{G}(N)$.
We can view the set of games $\mathcal{G}(N)$ as a vector space
by defining addition and scalar multiplication as follows:
\begin{align*}
	(v + w)(S) & = v(S) + w(S) \\
	(cv)(S)    & = cv(S)
\end{align*}
for any $S \subseteq N, v,w \in \mathcal{G}(N), c \in \mathbb{R}$.
The vector space of games $\mathcal{G}(N)$ is isomorphic to the euclidean space $\mathbb{R}^{2^{n}-1}$:
any game $v \in \mathcal{G}(N)$ can be identified with a vector $\mathbf{v} \in \mathbb{R}^{2^{n}-1}$
where each entry corresponds to the worth $v(S)$ of a nonempty coalition $S \subseteq N$
(remember that $v(\emptyset) = 0$ by definition).
It is easy to see that the sum and scalar product in this euclidean space
indeed preserve the same structure in the space of games.

Consider a game $v \in \mathcal{G}(N)$,
and let $\pi \in \Pi(N), \pi: N \rightarrow N$ be any permutation of $N$.
For any $S \subseteq N$,
define $\pi S := \{\pi(i):i \in S\}$.
The \textit{permuted game} $\pi v \in \mathcal{G}(N)$ with respect to $\pi$ is defined as follows:
\begin{equation*}
	(\pi v)(\pi S) = v(S), \forall S \subseteq N
\end{equation*}
The permuted game $\pi v$ can be interpreted as a game obtained by ``relabeling'' the players in $N$ using $\pi$.
As an example, consider the game $v_{\text{ex}}$ introduced in the beginning of this section.
Consider also the permutation $\pi := (2,1,3)$.
This permutation swaps the first two players.
We then have:
\begin{align*}
	\pi(\{ 1 \})   & = \{ 2 \}   \\
	\pi(\{ 2 \})   & = \{ 1 \}   \\
	\pi(\{ 1,3 \}) & = \{ 2,3 \} \\
	\pi(\{ 2,3 \}) & = \{ 1,3 \} \\
\end{align*}
and $\pi(S) = S$ for all other coalitions $S$.
The corresponding permuted game $\pi v_{\text{ex}}$ is then:
\begin{align*}
	(\pi v_{\text{ex}})(\emptyset)                        & = 0 & (\pi v_{\text{ex}})(N)                                    & = 8 \\
	(\pi v_{\text{ex}})(\{ 1 \}) = v_{\text{ex}}(\{ 2 \}) & = 2 & (\pi v_{\text{ex}})(\{ 1,2 \})                            & = 3 \\
	(\pi v_{\text{ex}})(\{ 2 \}) = v_{\text{ex}}(\{ 1 \}) & = 1 & (\pi v_{\text{ex}})(\{ 2,3 \}) = v_{\text{ex}}(\{ 1,3 \}) & = 5 \\
	(\pi v_{\text{ex}})(\{ 3 \})                          & = 4 & (\pi v_{\text{ex}})(\{ 1,3 \}) = v_{\text{ex}}(\{ 2,3 \}) & = 7
\end{align*}
Indeed, it is easy to see that this game is in a sense equivalent to $v_{\text{ex}}$,
as the players $1$ and $2$ have simply swapped their labels.

In a given game $v$,
we say that a coalition $S \subseteq N$ is a \textit{dummy coalition} if:
\begin{equation*}
	\forall T \subseteq N \setminus S: v(T \cup S) = v(T) + v(S)
\end{equation*}
\textit{i.e.}~the contribution of the coalition $S$ to any other coalition is a constant.
A dummy coalition $S$ for which $v(S) = 0$ is also called a \textit{null coalition}.
If $S$ is a singleton, then we also speak of a dummy or null \textit{player}, respectively.
In the game $v_{\text{ex}}$,
player $1$ is an example of a dummy player, but not a null player.
Finally, a \textit{partnership} $P \neq \emptyset$ in $v$ is defined as a coalition $P \subseteq N$
such that:
\begin{equation*}
	\forall T \subseteq N \setminus P, S \subset P: v(T \cup S) = v(T)
\end{equation*}
\textit{i.e.}~if some but not all of the members of a partnership $P$
are present in the coalition,
then these members leave the worth of the coalition unchanged.
This also implies that $v(S) = 0$ for any strict subset $S \subset P$.
A \textit{dummy partnership} is a partnership that is also dummy.
A partnership can be viewed as equivalent to a single player.

Let $T \subseteq N, T \neq \emptyset$ be a non-empty coalition for a game $v$.
The \textit{reduced game} with respect to $T$
is denoted as $v_{[T]} \in \mathcal{G}((N \setminus T) \cup [T])$,
where $[T]$ denotes a single new player not in $N$,
and is defined as
\begin{align*}
	v_{[T]}(S)          & := v(S)        \\
	v_{[T]}(S \cup [T]) & := v(S \cup T)
\end{align*}
for any $S \subseteq N \setminus T$.
The reduced game with respect to $T$ can be interpreted
as the game obtained by ``merging'' the coalition $T$ into a single player $[T]$.
Note that this can be linked to the concept of a partnership.
Indeed, if $P$ is a partnership, then $v$ and the reduced game $v_{[P]}$
can be considered equivalent.

As an example, assume $T = \{ 2,3 \}$ in the game $v_{\text{ex}}$.
We can then define the new game $v_{\text{ex},[T]}$ with the players
$(N \setminus T) \cup [T] = \{ 1, [T] \}$:
\begin{align*}
	v_{\text{ex},[T]}(\emptyset)    & = 0 \\
	v_{\text{ex},[T]}(\{ 1 \})      & = 1 \\
	v_{\text{ex},[T]}(\{ [T] \})    & = 7 \\
	v_{\text{ex},[T]}(\{ 1, [T] \}) & = 8
\end{align*}
This game can indeed be interpreted as a ``merging''
of the players $2$ and $3$ into a single player $[T]$.
Note that player $1$ is still a dummy player in this reduced game.

\subsection{Simple games}

A \textit{simple game} is a game for which the characteristic function takes on only the values in $\{ 0,1 \}$.
Coalitions with a worth of $1$ and $0$ are also called \textit{winning} and \textit{losing coalitions}, respectively.
Such a game can alternatively be expressed by simply listing the winning coalitions.
A common additional assumption is that any coalition containing a winning coalition is also winning,
or equivalently, that any subset of a losing coalition is also losing.
In that case, the game can be represented even more tersely by listing only the minimal winning coalitions,
\textit{i.e.}~the winning coalitions for which each proper subset is losing.
A common application of simple games is in \textit{voting games},
where for example we want to model a bicameral legislature.
In such a situation, a coalition is winning if it contains a majority in both chambers.
In a voting game, the usual object of interest is the \textit{power} any one member of the group has
in influencing the outcome of the vote.
This power can be quantified using a \textit{power index} \citep{penrose1946},
which is a real vector where each entry reflects the voting power of a given member.
Although this problem might seem different enough from the quantification of value of a given player to the game,
power indices are actually very similar to values.
In fact, the Banzhaf and Shapley-Shubik power indices,
which are two of the most well-known power indices,
are simply defined as a value (the Banzhaf and Shapley value, respectively) for the voting game in its characteristic form.
In other words, determining the power of a single member in a simple game
can be viewed as quantifying the ``value'' of that player
to a game where the worth of any coalition is either 0 or 1,
and the worth of the total set of players is 1.

A specific type of simple game is the \textit{unanimity game} $v_{R}, R \subseteq N$.
The unanimity game for a set $R$ is defined as:
\begin{equation*}
	v_{R}(S) = \begin{cases}
		1 & \mbox{if } R \subseteq S \\
		0 & \mbox{otherwise.}
	\end{cases}
\end{equation*}
This is a simple game where a coalition is winning if and only if it contains $R$.
The main reason why unanimity games are interesting is the fact that they form a basis for $\mathcal{G}(N)$.
In the vector representation of games discussed above,
each unanimity game $v_{R}$ corresponds to a vector
containing 1 for every subset containing $R$,
and 0 for every other subset.
It is easy to see that there are $2^{n}-1$ such vectors,
and that they are linearly independent.
Therefore, they constitute a basis for $\mathbb{R}^{2^{n}-1}$,
and the corresponding unanimity games analogously constitute a basis for the vector space $\mathcal{G}(N)$.
This property will be very useful in defining values for games:
if we assume linearity as one of the axioms for a value, then defining this value on the set of unanimity games
is sufficient to define it on the entire space of games.

\subsection{The Value Problem}
\label{sec:game-theory-value-problem}
As mentioned earlier, one of the main objects of study in TU-games is the quantification of value of the members of $N$
to the game.
Such a value can be represented as a vector $\phi(v) \in \mathbb{R}^{n}$.
Often, the value will be represented as a separate function $\phi_{i}$ for each player:
$\phi(v) = (\phi_{1}(v), \dots, \phi_{n}(v))$.
In order to define a value that is fair, we must first define what it means to be ``fair.''
To this end, several \textit{axioms} have been proposed as ``reasonable'' properties that a given value should adhere to
in order to be called ``fair.''
In the following paragraphs,
we will first introduce some of the most important axioms in the context of this work.
Afterwards, we will construct a taxonomy of existing values for games
based on the axioms they do or do not adhere to.

\subsubsection{Axioms for Values}
Before we give an overview of the most important axioms,
we first introduce some necessary notation.
Let $\Pi(N)$ denote the set of permutations on the set $N$,
\textit{i.e.}~$\Pi(N) = \{ \pi: N \rightarrow N \mid \pi \text{ is a bijection} \}$.
If $\pi \in \Pi(N)$ and $i,j \in N$,
then the \textit{precedence} relation in $\pi$
will be denoted as $\prec_{\pi}, \preceq_{\pi}$
for strict and non-strict precedence, respectively.
For example, if $N = \{ 1,2,3 \}$ and $\pi(N) = ( 2,3,1 )$,
then $2 \prec_{\pi} 1$.
We will now introduce the axioms that will be relevant to the definition of values.

\begin{itemize}
	\item \textbf{Linearity:} Given games $v,w \in \mathcal{G}(N)$ and constants $\alpha,\beta \in \mathbb{R}$.
	      Using the vector space definition for games,
	      we can define the linear combination of games $\alpha v + \beta w$:
	      \begin{equation*}
		      (\alpha v + \beta w)(S) = \alpha v(S) + \beta w(S), \forall S \subseteq N
	      \end{equation*}
	      A value $\phi_i$ adheres to the Linearity axiom if it is a linear function on $\mathcal{G}(N)$:
	      \begin{equation*}
		      \forall \alpha, \beta \in \mathbb{R}, v, w \in \mathcal{G}(N):
		      \phi_i(\alpha v + \beta w) = \alpha \phi_i(v) + \beta \phi_i(w)
	      \end{equation*}
	      Intuitively, this axiom states that if the outcome of a given game
	      is defined as a linear combination of the outcomes of two or more ``sub-games,''
	      then the value of any given player should be the same linear combination
	      of values for the sub-games.
	\item \textbf{Null:} A value adheres to the Null axiom if,
	      for every game $v \in \mathcal{G}(N)$ with null player $i$:
	      \begin{equation*}
		      \phi_{i}(v) = 0
	      \end{equation*}
	      Intuitively, this axiom states that the value should reflect the fact that
	      a null player provides no contributions or value to the game in any context.
	\item \textbf{Dummy:} A value adheres to the Dummy axiom if, for every game $v \in \mathcal{G}(N)$ with dummy player $i$:
	      \begin{equation*}
		      \phi_i(v) = v(i)
	      \end{equation*}
	      The intuitive interpretation of this axiom is that,
	      if a player provides the exact same value independently of which other players are present in the coalition,
	      then the value of that player to the game in general should be equal to that value.
	      Note that this is a stronger version of the Null axiom:
	      The Null axiom simply states that the Dummy axiom must hold for all dummy players
	      $i$ with $v(i) = 0$.
	      In literature, the term Dummy is often used for both of these axioms interchangeably \citep{weber1988,sundararajan2020}.
	      To prevent confusion, we use the term \textit{Null} for the weaker axiom.
	\item \textbf{Monotonicity:} A game $v \in \mathcal{G}(N)$ is called \textit{monotonic}
	      if $\forall S \subseteq T \subseteq N: v(S) \leq v(T)$.
	      A value $\phi_{i}$ adheres to the Monotonicity axiom if,
	      for any monotonic game $v \in \mathcal{G}(N)$,
	      $\phi_{i}(v) \geq 0$.
	\item \textbf{Efficiency:} A value $\phi$ adheres to the Efficiency axiom if:
	      \begin{equation*}
		      \forall v \in \mathcal{G}(N): \sum_{i \in N} \phi_i(v) = v(N)
	      \end{equation*}
	      Intuitively, this means that the total utility $v(N)$
	      that is obtained through the cooperation is split up
	      entirely and exactly among the players.
	      This allows the value to be interpreted as a \textit{distribution} of the total utility $v(N)$:
	      in this case, each player receives a part of the total utility
	      that is equal to the value of  the player to the game.
	\item \textbf{2-Efficiency:} This is an alternative axiom to Efficiency.
	      Consider a game $v \in \mathcal{G}(N)$
	      and $i,j \in N$.
	      Consider also the reduced game with respect to $\{ i,j \}$:
	      \begin{align*}
		      v_{[ij]}(S)           & := v(S)                \\
		      v_{[ij]}(S \cup [ij]) & := v(S \cup \{ i,j \})
	      \end{align*}
	      for any $S \subseteq N \setminus \{ i,j \}$.
	      A value $\phi$ then adheres to the 2-Efficiency axiom if:
	      \begin{equation*}
		      \phi_{[ij]}(v_{[ij]}) = \phi_{i}(v) + \phi_{j}(v)
	      \end{equation*}
	      In other words, the 2-Efficiency axiom states that if two players are merged,
	      then they should have the same value as they did before merging.
	\item \textbf{Anonymity:} A value adheres to the Anonymity axiom if:
	      \begin{equation*}
		      \forall v \in \mathcal{G}(N), \pi \in \Pi(N), i \in N: \phi_i(v) = \phi_{\pi(i)}(\pi v)
	      \end{equation*}
	      Intuitively, the Anonymity axiom states that
	      relabeling the players of the game has no influence on the value.
	\item \textbf{Symmetry:} Two players $i,j \in N$ are called \textit{symmetric}
	      in a game $v \in \mathcal{G}(N)$ if,
	      for any subset $S \subseteq N \setminus \{i,j\}$:
	      \begin{equation*}
		      v(S \cup i) = v(S \cup j)
	      \end{equation*}
	      A value adheres to the Symmetry axiom if,
	      for any game $v \in \mathcal{G}(N)$ with symmetric variables $i,j \in N$:
	      \begin{equation*}
		      \phi_{i}(v) = \phi_{j}(v)
	      \end{equation*}
	      Note that the Symmetry axiom is weaker than the Anonymity axiom.
	      This can easily be seen by considering the permutation that simply swaps elements $i$ and $j$:
	      \begin{equation*}
		      \sigma_{ij}(k) = \begin{cases}
			      j & \mbox{if } k = i  \\
			      i & \mbox{if } k = j  \\
			      k & \mbox{otherwise.}
		      \end{cases}
	      \end{equation*}
	      If $i$ and $j$ are symmetric in $v$,
	      then $\sigma_{ij}v = v$
	      and the required result follows immediately from Anonymity.
	      Note also that the term Symmetry is often used in literature to denote the stronger Anonymity axiom \citep{weber1988}.
	      In this dissertation, I will use \textit{Symmetry} to denote the weaker version of the axiom.
\end{itemize}

\subsubsection{Taxonomy of values}
Now that we have an overview of the relevant axioms,
we can use them to construct a taxonomy of values
based on the axioms they adhere to.
First, we define the \textit{marginal contribution}
of a player $i$ to a coalition $S$,
which can be interpreted as the additional utility that $i$
contributes if $i$ decides to join the coalition $S$:

\begin{definition}[Marginal Contribution]
	Given a game $v \in \mathcal{G}(N)$, $i \in N$,
	and $S \subseteq N \setminus i$.
	The \textbf{marginal contribution} of $i$ to $S$ is defined as:
	\begin{equation*}
		\Delta_{i}v(S) = v(S \cup i) - v(S)
	\end{equation*}
\end{definition}

We will now construct the taxonomy of values.
\begin{figure}
	\centering
	\includegraphics[width=\textwidth]{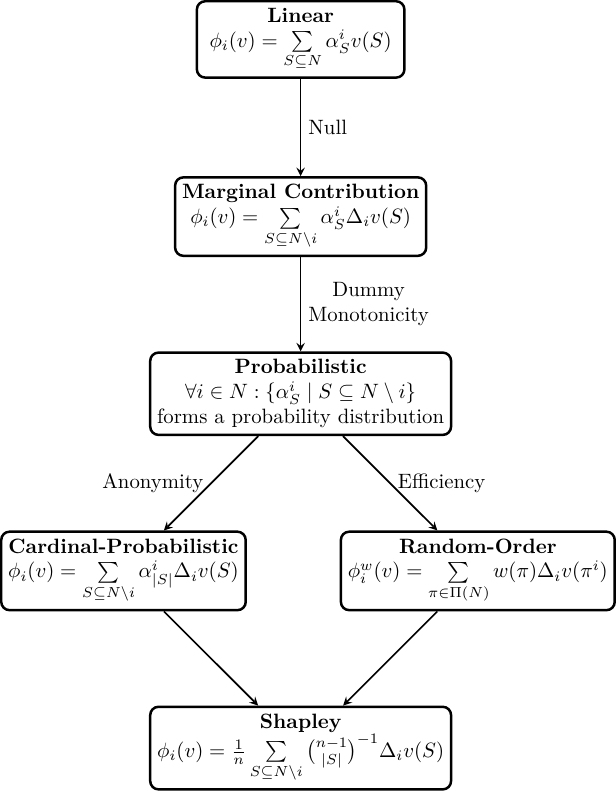}
	\caption{Summary of the different values and their corresponding axioms.}
	\label{fig:game-theory-taxonomy-values}
\end{figure}
This taxonomy is largely based on \citet{weber1988},
and we refer the reader to this publication for proofs and further details.
An overview of the taxonomy is given in \Cref{fig:game-theory-taxonomy-values}.
The most general kind of value that we will consider is a value that adheres
only to the Linearity axiom.
It can be shown \citep{weber1988} that if a value $\phi_{i}$ adheres to the Linearity axiom,
then for each player $i$
there exists a set of constants $\{ \alpha^{i}_{S} \mid S \subseteq N \}$ such that,
for any $v \in \mathcal{G}(N)$:
\begin{equation*}
	\phi_{i}(v) = \sum_{S \subseteq N}\alpha^{i}_{S}v(S)
\end{equation*}
\textit{i.e.}~the value is a linear combination of the characteristic function $v$
for all subsets $S \subseteq N$.
Furthermore, it can also be shown \citep{grabisch1999} that if a value additionally adheres to the Null axiom,
then for each player $i$
there exists a set of constants $\{ \alpha^{i}_{S} \mid S \subseteq N \}$ such that:
\begin{equation*}
	\phi_{i}(v) = \sum_{S \subseteq N \setminus i}\alpha_{S}^{i}\Delta_{i}v(S)
\end{equation*}
\textit{i.e.}~the value is a linear combination of \textit{marginal contributions}
of the corresponding player $i$ to all subsets $S \subseteq N \setminus i$.
For this reason, we will use the term \textit{marginal contribution (MC) value} for any value that adheres to the Linearity and Null axioms.
All values that we will cover in the following are MC values.

If a given MC value satisfies the Dummy and Monotonicity axioms,
then it can be shown that for each player $i$ the constants
$\alpha_{S}^{i}$ form a probability distribution
over all the sets not containing $i$ \citep{weber1988}:
\begin{align*}
	\forall i \in N, S \subseteq N \setminus i:\alpha_{S}^{i} & \geq 0 \\
	\sum_{S \subseteq N \setminus i} \alpha_{S}^{i}           & = 1
	,
\end{align*}
Such a value is also called a \textit{probabilistic value} \citep{weber1988},
as it can be viewed as a weighted average of marginal contributions
of player $i$ to all subsets not containing $i$.
If the probability distribution formed by the constants $\alpha_{S}^{i}$
describes the probability of player $i$ joining any given coalition $S$,
then the corresponding probabilistic value is the expected value
of player $i$'s marginal contribution.
If a probabilistic value additionally satisfies the Anonymity axiom,
then it is also called a \textit{cardinal-probabilistic value} or \textit{semivalue} \citep{dubey1981}.
In such a value, the constants $\alpha_{S}^{i}$ only depend on the cardinality of the set $S$:
$\forall i \in N, S,T \subseteq N \setminus i: |S| = |T| \implies \alpha_{S}^{i} = \alpha_{T}^{i}$.
An example of a semivalue is the Banzhaf value \citep{nowak1997}:
\begin{equation*}
	\phi_{i}^{\text{B}}(v) = \frac{1}{2^{n-1}} \sum_{S \subseteq N \setminus i} (v(S \cup i) - v(S))
\end{equation*}
This value has a simple interpretation:
it is the average marginal contribution of player $i$
over all coalitions not containing $i$,
\textit{i.e.}~it is the probabilistic value where the distribution over subsets
is the uniform distribution.
The Banzhaf value is the unique semivalue that also satisfies the 2-Efficiency axiom \citep{grabisch1999}.
In a voting game context, this value is also called the \textit{Banzhaf power index} \citep{penrose1946,banzhaf1964},
in which case it can be interpreted as the probability that $i$ has the ``swing vote:''
the vote that makes a coalition go from losing to winning.

A \textit{random-order value} or \textit{quasivalue} \citep{weber1988} is a probabilistic value
where the weighted average of marginal contributions
is computed over \textit{permutations} rather than subsets.
Let $w: \Pi(N) \rightarrow [0,1]$ form a probability distribution
over all permutations of $N$,
\textit{i.e.}~$\sum_{\pi \in \Pi(N)}w(\pi) = 1$.
Note that this probability distribution does not depend on any specific member of $N$,
as opposed to the subset probabilities in a probabilistic value.
The random-order value according to $w$ is then defined as:
\begin{equation*}
	\phi_i^w(v) = \sum_{\pi \in \Pi(N)}w(\pi)[v(\{j: j \preceq_\pi i\}) - v(\{j:j\prec_\pi i\})]
\end{equation*}
It can easily be shown that any random-order value is a probabilistic value
as the subset probabilities can be constructed by summing up the appropriate
permutation probabilities:
\begin{align*}
	\phi_i^{w}(v) & = \sum_{S \subseteq N \setminus i}p_S^i[v(S \cup i) - v(S)] \\
	p_S^i         & = \sum_{\pi \in \Pi_S^i} w(\pi)                             \\
	\Pi_S^i       & = \{\pi \in \Pi(N): \{ j: j \prec_{\pi} i \} = S\}
\end{align*}
It is easy to see that for any $i \in N$,
the set $\{ \Pi_{S}^{i} \mid S \subseteq N \setminus i \}$
forms a partition of the set of all permutations $\Pi(N)$,
which immediately implies that the probability distribution over subsets
formed by $\{ p_{S}^{i} \mid S \subseteq N \setminus i \}$
is also a valid probability distribution.
We can also see that any random-order value adheres to the Efficiency axiom:
\begin{align*}
	\sum_{i \in N}\phi_{i}^{w}(v) & = \sum_{i \in N}\sum_{\pi \in \Pi(N)}w(\pi)[v(\{j: j \preceq_\pi i\}) - v(\{j:j\prec_\pi i\})] \\
	                              & = \sum_{\pi \in \Pi(N)}w(\pi)\sum_{i \in N}[v(\{j: j \preceq_\pi i\}) - v(\{j:j\prec_\pi i\})] \\
	                              & = \sum_{\pi \in \Pi(N)}w(\pi)v(N)                                                              \\
	                              & = v(N)                                                                                         \\
\end{align*}
Furthermore, \citet{weber1988} shows that a probabilistic value being a random-order value is not
only sufficient for Efficiency, but also necessary:
a probabilistic value satisfies the Efficiency axiom if and only if it is a random-order value.

Note that both random-order and cardinal-probabilistic values
are specific kinds of probabilistic values,
but a random-order value is not necessarily a cardinal-probabilistic value,
and vice versa.
In fact, if we want to construct a value that is both random-order and cardinal-probabilistic,
\textit{i.e.}~a value that satisfies the Linearity, Dummy, Anonymity and Efficiency
(note that these axioms imply the Monotonicity axiom),
then this value is uniquely defined.
The resulting value is the Shapley value \citep{shapley1953}:
\begin{align*}
	\phi_{i}(v) & = \frac{1}{n}\sum_{S \subseteq N \setminus i} \binom{n-1}{|S|}^{-1} [v(S \cup i) - v(S)]          \\
	            & = \frac{1}{n!}\sum_{\pi \in \Pi(N)} [v(\{ j: j \preceq_{\pi} i \}) - v(\{ j: j \prec_{\pi} i \})]
\end{align*}
It is easy to see that this value is indeed both a random-order value and a semivalue,
and therefore satisfies the desired axioms.

We can prove the uniqueness of the Shapley value by defining it only for the set of unanimity games,
which immediately fixes the definition for the complete set of games on $N$,
as the unanimity games form a basis for the space of games.
The proof of the following theorem is based on \citet{dubey1975}.

\begin{theorem}
	For any finite set of players $N$ with $|N| = n$,
	the Shapley value is the unique value $\phi: \mathcal{G}(N) \rightarrow \mathbb{R}^{n}$
	such that $\phi$ satisfies the Dummy, Anonymity, Efficiency and Linearity axioms.
\end{theorem}
\begin{proof}
	Assume $v_{R}$ is the unanimity game for subset $R \subseteq N$:
	\begin{equation*}
		v_{R}(S) = \begin{cases}
			1 & \mbox{if } R \subseteq S \\
			0 & \mbox{otherwise.}
		\end{cases}
	\end{equation*}
	Assume also that $\phi_{i}$ is a value that adheres to the
	Dummy, Anonymity, Efficiency and Linearity axioms.
	As we know that the Shapley value adheres to these axioms,
	we know that at least one such value exists.
	It is clear that any player $i \notin R$ is a null player in $v_{R}$.
	Therefore:
	\begin{equation}
		\label{eqn:game-theory-shapley-unique-null}
		\forall i \notin R: \phi_{i}(v_{R}) = 0
	\end{equation}
	Using Efficiency, we therefore get:
	\begin{equation}
		\label{eqn:game-theory-shapley-unique-efficiency}
		\sum_{i \in N} \phi_{i}(v_{R}) = \sum_{i \in R} \phi_{i}(v_{R}) = 1
	\end{equation}
	If $|R| = 1$, this immediately implies that $\phi_{i}(v_{R}) = 1$, and we are done.
	Note that $|R| = 0$ is not an option, as this would imply that $v_{R}(\emptyset) = 1$,
	which is not a valid game.
	Assume now that $|R| > 1$.
	Choose any two players $i, j \in R$.
	We can see that $i$ and $j$ are symmetric:
	if $S \subseteq N \setminus \{ i,j \}$,
	then $v_{R}(S) = v_{R}(S \cup i) = v_{R}(S \cup j) = 0$.
	Therefore, $\phi_{i}(v_{R}) = \phi_{j}(v_{R})$ for any two players $i,j \in R$.
	Combining this with \Cref{eqn:game-theory-shapley-unique-efficiency,eqn:game-theory-shapley-unique-null}
	we see that the only possible definition for $\phi_{i}(v_{R})$ is:
	\begin{equation*}
		\phi_{i}(v_{R}) = \begin{cases}
			\frac{1}{|R|} & \mbox{ if } i \in R \\
			0             & \mbox{ otherwise.}
		\end{cases}
	\end{equation*}
	Assume now that $v \in \mathcal{G}(N)$.
	As the unanimity games form a basis for $\mathcal{G}(N)$,
	we can write $v$ as:
	\begin{equation*}
		v(S) = \sum_{R \subseteq N} a_{R}v_{R}(S)
	\end{equation*}
	for some set of constants $\{ a_{R} \mid R \subseteq N \}$.
	Using Linearity, we therefore get:
	\begin{equation*}
		\phi_{i}(v) = \sum_{R \subseteq N} a_{R} \phi_{i}(v_{R})
	\end{equation*}
	which completely defines the value, implying that it is indeed unique.
\end{proof}

A common intuitive interpretation of the Shapley value is the following.
Assume the players enter the coalition in a random order.
If a player $i$ is preceded by a set of players $S$,
then the contribution that $i$ makes when entering the coalition
is the marginal contribution of $i$ to $S$.
The Shapley value for any given player is therefore equal to their
expected contribution if the order in which the players enter the coalition
is chosen uniformly at random.

\subsection{Interaction Indices}
\label{sec:game-theory-interaction-indices}
The value problem described in \Cref{sec:game-theory-value-problem}
focuses on the value of individual players
to a given cooperative game.
However, the contribution of one player is not necessarily independent of the other players,
\textit{i.e.}~the worth $v(S)$ of a set $S = \{ i,j \}$ need not be equal
to the sum of the worths $v(i) + v(j)$ of the individual players $i$ and $j$.
The difference between $v(\{ i,j \})$ and $v(i) + v(j)$
can be viewed as the added value of cooperation,
or the \textit{interaction effect},
between $i$ and $j$.
This leads to a generalization of the concept of a value to subsets of players $S \subseteq N$ \citep{grabisch1999}.
In this case, the value $\phi_S: 2^{N} \rightarrow \mathbb{R}$ is called an \textit{interaction index}.
A value can be viewed simply as an interaction index where
\begin{equation*}
	\forall v \in \mathcal{G}(N), S \subseteq N: | S | > 1 \implies \phi_S(v) = 0
\end{equation*}
The \textit{order} of an interaction index $\phi$
is defined as the minimal $k \in \mathbb{N}$ such that $\forall v \in \mathcal{G}(N), S \subseteq N: | S | > k \implies \phi_S(v) = 0$.
A value is then equivalent to an interaction index of order $1$.
In the following paragraphs, we will first introduce the concept of
the \textit{discrete derivative},
which will be of central importance in the definition of interaction indices.
After this, I will give an overview of the generalized versions
of axioms for values to interaction indices.
Finally, we will use these axioms,
along with a selection of new axioms defined specifically for interaction indices,
to construct a taxonomy of interaction indices
based on the axioms they do or do not adhere to.

\subsubsection{Discrete Derivatives}
Interaction indices are designed to capture
the non-additive interaction effects that are caused by the set $S$ as a whole.
We will quantify these interaction effects
by generalizing the marginal contribution $\Delta_{i}v(S)$ of a player $i$
to coalitions $S \subseteq N$: $\Delta_{S}v(T), \forall S, T \subseteq N$.
This generalization will be called the \textit{discrete derivative}.
To see how these concepts are connected,
assume $i,j \in N$ are two players for a given game $v \in \mathcal{G}(N)$.
We distinguish three possibilities (the following explanation is based on \citet{fujimoto2006}):
\begin{itemize}
	\item $v(i) + v(j) < v(\{ i,j \})$.
	      In this case, both $i$ and $j$ would be interested in forming a coalition,
	      as together they can achieve more than they would if they operate separately.
	      This difference between the sum of individual utilities and the utility of $\{ i,j \}$ as a whole
	      can be viewed as a \textit{positive interaction} between $i$ and $j$.
	\item $v(i) + v(j) > v(\{ i,j \})$.
	      This is the reverse of the previous case: now $i$ and $j$ would not be interested to form a coalition,
	      as they can achieve more on their own.
	      The difference can now be viewed as a \textit{negative interaction} between the two players.
	\item $v(i) + v(j) = v(\{ i,j \})$.
	      In this case, whether $i$ and $j$ decide to collaborate makes no difference.
	      The two players behave \textit{additively}, \textit{i.e.}~there is no interaction between them.
\end{itemize}
These possibilities show that the quantity of interest in defining the interaction
between $i$ and $j$ is the difference between their worth as a coalition
and the sum of their worths as singletons:
\begin{equation*}
	\delta := v(\{ i,j \}) - v(\{ i \}) - v(\{ j \})
\end{equation*}
If $\delta > 0$, then $i$ and $j$ interact positively, and vice versa.
If $\delta = 0$, then $i$ and $j$ behave additively, \textit{i.e.}~they do not interact.

Of course, this does not tell the complete story of interaction between players.
First, we are not only interested in the interaction between two players $i,j$,
but more generally in interactions between arbitrary subsets of players $S \subseteq N$.
Furthermore, whereas we considered the interaction between $i$ and $j$ in isolation,
we should also be interested in the interaction between players when
a set of other players is already present.
Perhaps $i$ and $j$ behave additively in isolation,
but have a positive interaction when player $k$ is present:
\begin{equation*}
	(v(\{ i,k \}) - v(\{ k \})) + (v(\{ j,k \}) - v(\{ k \})) < v(\{ i,j,k \}) - v(\{ k \})
\end{equation*}
Notice that in this example, we subtract $v(\{ k \})$ from each of the terms.
This is because we view the interaction effect as the difference in \textit{contributions}
when $k$ is present: on the left hand side we have the contributions of $i$ and $j$
separately in the presence of $k$, and on the right hand side we have
the joint contribution of $\{ i,j \}$ in the presence of $k$.

We can generalize this idea of interaction \textit{in the presence of} a coalition
$T \subseteq N \setminus \{ i,j \}$ as:
\begin{equation*}
	\Delta_{ij}v(T) := v(T \cup \{ i,j \}) - v(T \cup i) - v(T \cup j) + v(T)
\end{equation*}
Where we have simply moved all of the terms in the inequality to the same side.
The earlier example of the interaction between $i$ and $j$ in isolation
then corresponds to $\Delta_{ij}v(\emptyset)$.
We call this the \textit{discrete derivative of $\{ i,j \}$ at $T$}.
It is easy to see that the discrete derivative of $\{ i,j \}$ at $T$
is a difference in marginal contributions:
\begin{align*}
	\Delta_{ij}v(T) & = \Delta_{i}v(T \cup j) - \Delta_{i}v(T) \\
	                & = \Delta_{j}v(T \cup i) - \Delta_{j}v(T)
\end{align*}
Intuitively, the discrete derivative can therefore be viewed as the
difference between the marginal contribution of $i$ to $T$
when $j$ is present or absent, and vice versa.
If this quantity is above resp. below 0,
then it seems natural to consider that $i$ and $j$ interact positively resp. negatively,
as the presence of one player increases resp. decreases the contribution of the other player
to $T$.

To generalize this reasoning to simultaneous interactions between coalitions $S$
with $|S| > 2$,
we can proceed analogously.
It is easy to see that, for any $\{ i,j,k \} \subseteq N$ and $T \subseteq N \setminus \{ i,j,k \}$:
\begin{align*}
	\Delta_{ijk}v(T) & = \Delta_{jk}v(T \cup i) - \Delta_{jk}v(T) \\
	                 & = \Delta_{ik}v(T \cup j) - \Delta_{ik}v(T) \\
	                 & = \Delta_{ij}v(T \cup k) - \Delta_{ij}v(T)
\end{align*}
If this quantity is larger than 0,
then adding the player $i$ increases the interaction between $j$ and $k$ at $T$,
and analogously for the other players.
This effect can be viewed as the \textit{simultaneous} interaction
between $i$, $j$ and $k$ in the presence of $T$.
We can now define the discrete derivative more generally
(for more details, see \citet{fujimoto2006}).
\begin{definition}[Discrete Derivative]
	For given coalitions $S,T \subseteq N$, the $S$\textit{-derivative} of $v$ at $T$,
	denoted as $\Delta_{S}v(T)$,
	is defined recursively as:
	\begin{align*}
                \Delta_{\emptyset}v(T) &:= v(T)\\
		\Delta_{i}v(T) &:= v(T \cup i) - v(T \setminus i), \forall i \in N         \\
		\Delta_{S}v(T) &:= \Delta_{i}[\Delta_{S \setminus i}v(T)], \forall i \in S
	\end{align*}
\end{definition}
Note that the operator $\Delta_{i}$ in the recursive rule for $\Delta_{S}$
acts on each of the terms in $\Delta_{S \setminus i}v(T)$ separately.
For example, if $|S| = 2$:
\begin{align*}
	\Delta_{ij}v(T) & = \Delta_{i}[\Delta_{j}v(T)]                             \\
	                & = \Delta_{i}[v(T \cup j) - v(T)]                         \\
	                & = \Delta_{i}v(T \cup j) - \Delta_{i}v(T)                 \\
	                & = v(T \cup \{ i,j \}) - v(T \cup j) - v(T \cup i) + v(T)
\end{align*}
Which is indeed the same quantity we had derived before.
It is easy to show by induction that, for any $S,T \subseteq N$:
\begin{equation*}
	\Delta_{S}v(T) = \Delta_{S}v(T \setminus S) = \sum_{L \subseteq S}(-1)^{|S|-|L|}v((T \setminus S) \cup L)
\end{equation*}
For a cooperative game $v$,
the discrete derivative at $\emptyset$ is also called the
\textit{Harsanyi dividend \citep{harsanyi1963}:}
\begin{align*}
	d(v,S) & = \Delta_{S}v(\emptyset)                 \\
	       & = \sum_{T \subseteq S}(-1)^{|S|-|T|}v(T) \\
\end{align*}
It can be shown that any game $v \in \mathcal{G}(N)$ can be uniquely expressed as:
\begin{equation*}
	v(S) = \sum_{T \subseteq S}d(v,T)
\end{equation*}
In a combinatorics context, the set function $d(v,\cdot)$
is also called the \textit{M\"obius transform} of $v$ \citep{rota1964}.

\subsubsection{Generalized Axioms for Interaction Indices}
\begin{figure}
	\centering
	\includegraphics[width=\textwidth]{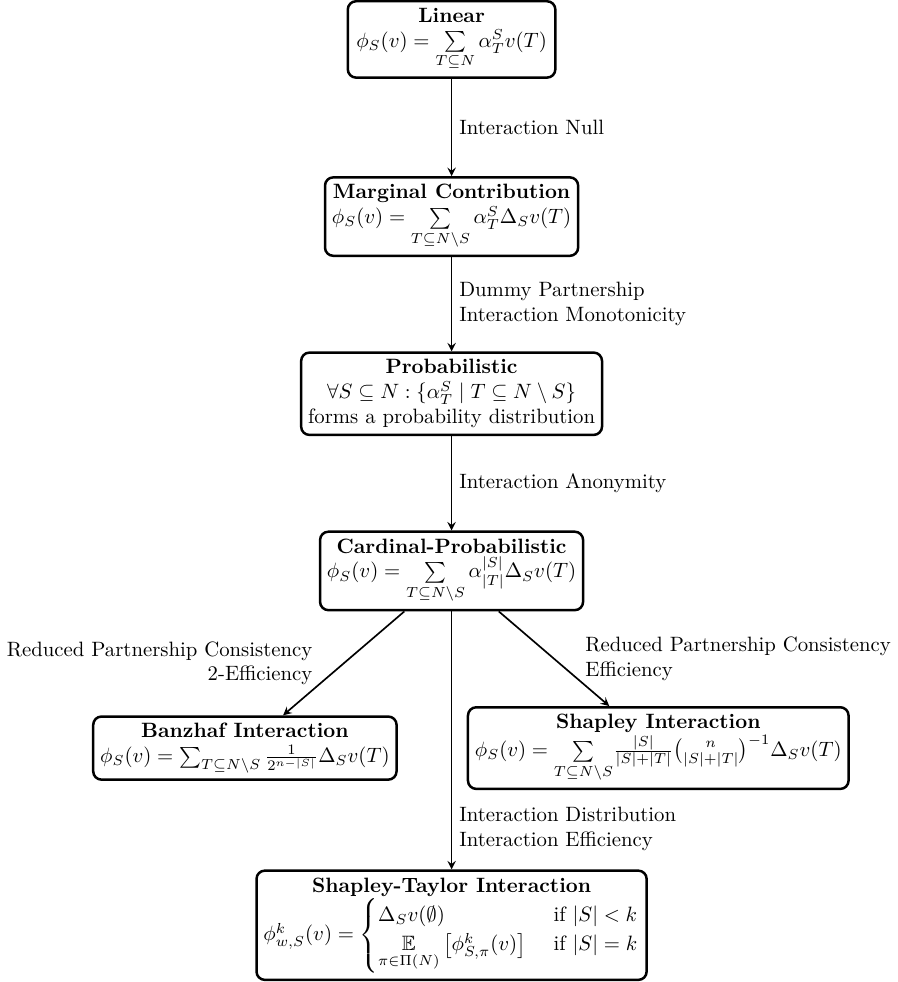}
	\caption{summary of the different interaction indices and their corresponding axioms.}
	\label{fig:game-theory-taxonomy-indices}
\end{figure}

The axioms from \Cref{sec:game-theory-value-problem} can be generalized to interaction indices,
which leads to a taxonomy of interaction indices that is very similar to that of values.
We will cover this taxonomy in the following paragraphs.
For proofs and further details, we refer the reader to \citet{grabisch1999} and \citet{fujimoto2006}.
A summary is given in \Cref{fig:game-theory-taxonomy-indices}.
We will now provide an overview
of the generalization of the axioms for values introduced in \Cref{sec:game-theory-value-problem}
to interaction indices.
\begin{itemize}
	\item \textbf{Interaction Null:} For any null player $i$ in $v$,
	      we have:
	      \begin{equation*}
		      \forall S \subseteq N \setminus i: \phi_{S \cup i}(v) = 0
	      \end{equation*}
	      \textit{i.e.}~the simultaneous interaction in a coalition
	      containing a null player must be zero.
	\item \textbf{Dummy Partnership:} If $P \neq \emptyset$ is a dummy partnership in $v$,
	      then:
	      \begin{align*}
		      \phi_{P}(v)                                                             & = v(P) \\
		      \forall S \subseteq N \setminus P, S \neq \emptyset: \phi_{S \cup P}(v) & = 0
	      \end{align*}
	      The first part of this axiom is a simple generalization of the Dummy axiom
	      to the more general dummy partnerships (remember that a dummy player is simply
	      a dummy partnership with a single member).
	      The second part states that there can be no simultaneous interaction
	      in a coalition that contains a dummy partnership.
	\item \textbf{Interaction Monotonicity:} It is easy to see that a game is monotonic if and only if
	      $\forall i \in N, S \subseteq N \setminus i: \Delta_{i}v(S) \geq 0$.
	      We can use this observation to generalize this definition to \textit{$k$-monotonicity} \citep{fujimoto2006}.
	      A game $v$ is $k$-monotonic if:
	      \begin{equation*}
		      \forall S \subseteq N, |S| \leq k, T \subseteq N \setminus S:\Delta_{S}v(T) \geq 0
	      \end{equation*}
	      An interaction index $\phi$ then adheres to the Interaction Monotonicity axiom if,
	      for any $k$-monotonic game $v$
	      and any coalition $S \subseteq N, |S| \leq k: \phi_{S}(v) \geq 0$.
	\item \textbf{Interaction Anonymity:} For a permutation $\pi \in \pi(N)$ of $N$ and subset $S \subseteq N$,
	      we denote $\pi(S) = \{ \pi(i) \mid i \in S \}$.
	      An interaction index then adheres to the Interaction Anonymity axiom if,
	      for any permutation $\pi \in \Pi(N)$, game $v \in \mathcal{G}(N)$
	      and subset $S \subseteq N$:
	      \begin{equation*}
		      \phi_{\pi(S)}(\pi v) = \phi_{S}(v)
	      \end{equation*}
	      This is again a straightforward generalization of the Anonymity axiom for values
	      to coalitions, in that the axiom states that relabeling the players
	      should have no influence on the outcome.
	\item \textbf{Interaction Efficiency:} An interaction index adheres to Interaction Efficiency if,
	      for any game $v$:
	      \begin{equation*}
		      \sum_{S \subseteq N} \phi_{S}(v) = v(N)
	      \end{equation*}
	      The only difference with the Efficiency axiom for values is that we now sum over subsets,
	      allowing us to view the interaction index as a way to distribute the total worth $v(N)$
	      over all coalitions, rather than players.
\end{itemize}
It is easy to see that these axioms are generalizations of their respective counterparts for values,
in the sense that any value, when viewed as an interaction index of order 1,
adheres to Null, Dummy, etc. if and only if it adheres to Interaction Null, Dummy Partnership, etc.

\subsubsection{Taxonomy of Interaction Indices}
Analogously to the value case, it can be shown that an interaction index $\phi_{S}$
satisfies Linearity if and only if for each $S \subseteq N$
there exists a set of constants $\{ \alpha_{T}^{S} \mid T \subseteq N \}$
such that:
\begin{equation*}
	\phi_{S}(v) = \sum_{T \subseteq N}\alpha_{T}^{S}v(T)
\end{equation*}
Similarly, it can be shown that an interaction index $\phi_{S}$
satisfies Linearity and Interaction Null if and only if
for each $S \subseteq N$ there exists a set of constants
$\{ \alpha_{T}^{S} \mid T \subseteq N \setminus S \}$ such that:
\begin{equation*}
	\phi_{S}(v) = \sum_{T \subseteq N \setminus S}\alpha_{T}^{S}\Delta_{S}v(T)
\end{equation*}
As this is a generalization of the MC value,
I will call an interaction index that satisfies
Interaction Null and Linearity a \textit{marginal contribution (MC) interaction index}.
Analogously, an interaction index satisfies Linearity, Dummy Partnership and $k$-Monotonicity
if and only if the set of constants $\{ \alpha_{T}^{S} \mid T \subseteq N \setminus S \}$
forms a valid probability distribution for each subset $S \subseteq N$.
Such an interaction index is called a \textit{probabilistic interaction index}.

It can again be shown that a probabilistic interaction index satisfies Interaction Anonymity
if and only if the constants $\alpha_{T}^{S}$ depend only on the cardinalities of $S$ and $T$,
\textit{i.e.}~for each $s \in \{ 1, \dots, n \}$ there exists a set of constants $\{ \alpha_{t}^{s} \mid t = 0, \dots, n-s \}$
such that for each subset $S \subseteq N$:
\begin{equation*}
	\phi_{S}(v) = \sum_{T \subseteq N \setminus S}\alpha_{|T|}^{|S|}\Delta_{S}v(T)
\end{equation*}
A probabilistic interaction index that adheres to Interaction Anonymity
is also called a \textit{cardinal-probabilistic interaction index}.

Although the generalization from values to interaction indices of the Linearity,
Null, Dummy and Anonymity axioms results in a completely analogous taxonomy
as in the context of values,
there are some differences when we attempt to generalize the Shapley and Banzhaf values.
Whereas in the value case we can uniquely identify the Shapley and Banzhaf values
as the semivalues that additionally adhere to Efficiency and 2-Efficiency, respectively,
this uniqueness is no longer valid in the interaction case.
To ensure uniqueness, a \textit{Recursive} axiom was introduced by \citet{grabisch1999}.
It can be shown that, if a cardinal-probabilistic interaction index adheres to this axiom,
then specifying Efficiency or 2-Efficiency uniquely defines an interaction index
that we call the Shapley and Banzhaf interaction index, respectively.
However, as this recursive axiom is much more technical than the other axioms,
it has been criticized for being specifically designed only to make the resulting interaction indices unique.
For this reason, alternative axiomatizations that avoid the recursive axiom have been proposed.
In the following, I will use the axiomatization as given by \citet{fujimoto2006}.
This axiomatization avoids the recursive axiom by introducing the following alternative:
\begin{itemize}
	\item \textbf{Reduced Partnership Consistency (RPC):} If $P$ is a partnership in a game $v$, then
	      \begin{equation*}
		      \phi_{P}(v) = \phi_{[P]}(v_{[P]})
	      \end{equation*}
	      This axiom states that the simultaneous interaction in a partnership,
	      which is a coalition that can be interpreted as
	      behaving like a single player,
	      should be equal to its marginal contribution when considered as a single player in the reduced game
	      $v_{[P]}$.
\end{itemize}
To see why this axiom is reasonable, it is sufficient to observe that if $P$ is a partnership, then:
\begin{equation*}
	\forall S \subseteq N \setminus P: \Delta_{P}(S) = v(P \cup S) - v(S)
\end{equation*}
\textit{i.e.}~the interaction within $P$ in the presence of $S$ is equal to its contribution to $S$.
With the RPC axiom, we can uniquely define the Shapley and Banzhaf interaction indices.
Assume an interaction index $\phi_{S}$ satisfies Linearity, Dummy Partnership, Interaction Anonymity,
and Reduced Partnership Consistency.
It can then be shown that:
\begin{itemize}
	\item $\phi_{S}$ satisfies Efficiency if and only if it is the Shapley interaction index:
	      \begin{equation*}
		      \phi_{S}(v) = \sum_{T \subseteq N \setminus S}\frac{|S|}{|S| + |T|}\binom{n}{|S|+|T|}^{-1}\Delta_{S}v(T)
	      \end{equation*}
	\item $\phi_{S}$ satisfies 2-Efficiency if and only if it is the Banzhaf interaction index:
	      \begin{equation*}
		      \phi_{S}(v) = \sum_{T \subseteq N \setminus S}\frac{1}{2^{n-|S|}}\Delta_{S}v(T)
	      \end{equation*}
\end{itemize}
Note that the Efficiency axiom here is the same axiom as defined for values,
\textit{i.e.}~the Shapley interaction indices for \textit{singletons} add up to the total utility $v(N)$.
This implies that the Shapley interaction index \textit{does not} adhere to Interaction Efficiency.
In fact, the Shapley interaction indices for singleton coalitions are simply identical to the Shapley values.
This makes sense in a game theoretic context,
as the main goal there is to fairly divide the utility $v(N)$ among the players $i \in N$.
The interaction index merely provides extra information on top of the Shapley value
about which coalitions seem to have stronger or weaker interactions,
and should therefore not be interpreted as parts of the total utility that are paid off to those coalitions.
In the game theoretic context, it would make no sense to distribute part of the utility to players $i$ and $j$ individually,
and then distribute another part of the utility to $\{ i,j \}$ as a coalition.

However, in other contexts, it might be preferable to have an interaction index
adhere to Interaction Efficiency rather than Efficiency.
An interaction index in a machine learning context is used to distribute the total behaviour of a function $f$
among its inputs (\textit{direct effects}) and groups of inputs (\textit{interaction effects}).
These effects are completely separate from each other,
so in this context it does make sense to attribute parts of the function's behaviour to a non-singleton coalition.
\citet{sundararajan2020a} address this issue by introducing the Shapley-Taylor interaction index.
The Shapley-Taylor interaction index is defined for a specific order $k,1 \leq k \leq n$,
and is a cardinal-probabilistic interaction index that additionally adheres to
Interaction Efficiency and a new Interaction Distribution axiom:
\begin{itemize}
	\item \textbf{Interaction Distribution:} an interaction index $\phi_{S}$
	      of order $k$
	      adheres to the Interaction Distribution axiom if,
	      for any unanimity game $v_{T}, T \subseteq N$:
	      \begin{equation*}
		      \forall S \subset T, |S| < k: \phi_{S}(v_{T}) = 0
	      \end{equation*}
	      The unanimity game $v_{T}$
	      can be viewed as a game that is completely determined by a single pure interaction between all members of $T$.
	      The Interaction Distribution axiom then states that, for such games,
	      the strict subsets of $T$ should receive no attribution except if they are of maximal size,
	      \textit{i.e.}~their size is equal to the order of explanation $k$.
\end{itemize}
Much like the original Shapley value, the Shapley-Taylor interaction index
can be defined as an expected value over permutations of players $\pi \in \Pi(N)$.
For a given permutation $\pi$, we first define:
\begin{equation*}
	\phi_{S,\pi}^{k}(v) = \begin{cases}
		\Delta_{S}v(\emptyset) & \mbox{ if } |S| < k \\
		\Delta_{S}v(\pi^{S})   & \mbox{ if } |S| = k \\
	\end{cases}
\end{equation*}
where $\pi^{S}$ is the set of players in $N$ that precede all of the elements in $S$:
\begin{equation*}
	\pi^{S} = \{ i \in N \mid \forall j \in S: i \prec_{\pi} j \}
\end{equation*}
The Shapley-Taylor interaction index is then defined as the expected value of this expression
over all permutations:
\begin{equation*}
	\phi_{S}^{k}(v) = \mathbb{E}_{\pi \in \Pi}[\phi_{S,\pi}^{k}]
\end{equation*}
As the value $\phi_{S,\pi}^{k}$ is constant for any subset $S$ with $|S| < k$,
we have:
\begin{equation*}
	\forall S \subset N, |S| < k: \phi_{S}^{k}(v) = \Delta_{S}v(\emptyset)
\end{equation*}
On the other hand, the authors show:
\begin{equation*}
	\forall S \subset N, |S| = k: \phi_{S}^{k}(v) = \frac{k}{n} \sum_{T \subseteq N \setminus S} \binom{n-1}{t}^{-1} \Delta_{S} v(T)
\end{equation*}
\citet{sundararajan2020a} then show that this interaction index indeed adheres to Interaction Efficiency.
This is proved by showing that $\phi_{S,\pi}^{k}$ adheres to Interaction Efficiency for any permutation $\pi$,
which immediately implies that the property also holds for the Shapley-Taylor interaction index
as this is simply an expected value of $\phi_{S,\pi}^{k}$ over all permutations.
Because the proof shows the desired property for any permutation $\pi$,
it immediately follows that we can generalize the Shapley-Taylor interaction index while retaining
Interaction Efficiency by simply taking the expected value over a different distribution of permutations
$w(\pi)$, much like the generalization of the Shapley value to random-order values.
However, as opposed to the case of random-order values,
it is unclear if this generalization is also a necessary condition,
\textit{i.e.}~if \textit{any} probabilistic interaction index that adheres to
Interaction Distribution and Interaction Efficiency is of this form.

\section{Removal-based Attribution}
\label{sec:removal-based-attribution}
In this section, we introduce a formal definition of removal-based attribution methods (RBAMs),
and show that many existing explanation methods can be viewed as RBAMS.
See \Cref{tbl:fdfi-shapley-methods,tbl:fdfi-non-shapley-methods} for examples of methods that can be viewed as RBAMs.
Assume $\mathcal{X} \subseteq \mathbb{R}^{d}$ is a (possibly infinite) hyperrectangle,
\textit{i.e.}~there exists a collection $\{ A_{i} \mid i = 1, \dots, d \}$
of non-empty intervals $A_{i}$
such that:
\begin{equation*}
	\mathcal{X} = \prod_{i = 1}^{d} A_{i}
\end{equation*}
Assume also that $\mathcal{F}$ is a linear space of functions $f: \mathcal{X} \rightarrow \mathbb{R}$.
An \textit{attribution method} is a function $m: \mathcal{F} \times 2^{[d]} \rightarrow \mathcal{F}$
mapping a function $f \in \mathcal{F}$ and feature subset $S \subseteq [d]$
to a new function $m(f,S) \in \mathcal{F}$.
This function can be interpreted as mapping each input point $\mathbf{x}$
to the ``importance'' (or attribution value)
of the features in $S$ to the function $f$ at that point $\mathbf{x}$.
We say that an attribution method is \textit{of order $k$} if
\begin{equation*}
	\forall f \in \mathcal{F}, S \subseteq [d]: |S| > k \implies m(f,S) = 0
\end{equation*}
If $k=1$, we call $m$ a \textit{feature attribution} method.
If $k>1$, we call $m$ an \textit{interaction attribution} method.
An example of a feature attribution method is SHAP \citep{lundberg2017},
whereas an example of an interaction attribution method is
the Shapley-Taylor interaction index \citep{sundararajan2020a}.

If for any function $f \in \mathcal{F}$
and any subset $S \subseteq [d]$,
the function $m(f,S) \in \mathcal{F}$ is a constant function,
then we call $m$ a \textit{global attribution method.}
If $m$ is not global, then we call it a
\textit{local attribution method.}
Intuitively, this means that the global attribution of $S$ to $f$
does not depend on any specific input point $\mathbf{x}$.
We call a global attribution method $m$ of order $k$
a global feature attribution or global interaction attribution method
if $k=1$ or $k>1$, respectively.
An example of a global attribution method is the PFI method \citep{breiman2001a}.

Let $\{ P_{S}: \mathcal{F} \rightarrow \mathcal{F} \mid S \subseteq [d] \}$
be a set of operators on the linear function space $\mathcal{F}$
with the following properties:
\begin{itemize}
	\item $\forall f \in \mathcal{F}, S \subseteq [d]: P_{S}(f) \text{ is independent of } X_{S}$
	\item $P_{\emptyset} = I$, \textit{i.e.}~$\forall f \in \mathcal{F}: P_{\emptyset}(f) = f$
\end{itemize}
We call these operators the \textit{removal operators.}
The first condition states that removing a subset of variables $S$
results in a function that is independent of those variables.
The second condition can be interpreted as stating that removing
none of the variables is equivalent to leaving the function unchanged.
Using these removal operators,
we can now define the specific class of attribution methods
that will be the object of study for the rest of this work:
\begin{definition}[Removal-Based Attribution Method]
	An attribution method $m: \mathcal{F} \times 2^{[d]} \rightarrow \mathcal{F}$
	is a \textbf{removal-based attribution method (RBAM)}
	if there exists a collection of constants $\{ \alpha_{T}^{S} \mid S,T \subseteq [d] \}$ (also called aggregation coefficients),
	a set of removal operators $\{ P_{T} \mid T \subseteq [d] \}$,
	and a mapping $\Phi: \mathcal{F} \rightarrow \mathcal{F}$
	such that for any $f \in \mathcal{F}, \mathbf{x} \in \mathcal{X}$:
	\begin{equation*}
		m(f,S)(\mathbf{x}) = \sum_{T \subseteq [d]} \alpha_{T}^{S}\Phi(P_{T}(f))(\mathbf{x})
	\end{equation*}
	\textit{i.e.}~the attribution for a subset of features $S$
	is a linear combination
	of $\Phi(P_{T}(f))(\mathbf{x})$,
	where $P_{T}(f)$ is the function $f$ where the features
	in $T$ have been removed.
	We call $\Phi(f)$ the explained behaviour of $f$,
	and the mapping $\Phi$ itself the behaviour mapping
	of $m$.
	If the behaviour mapping $\Phi$ is the identity mapping,
	then we call $m$ a simple RBAM.
	If for any $f \in \mathcal{F}$,
	$\Phi(f)$ is a constant function,
	then we call $\Phi$ a global behaviour mapping
	and $m$ a global RBAM.
	Otherwise, we call $\Phi$ a local behaviour mapping
	and $m$ a local RBAM.
\end{definition}

Note that the definition of removal-based attribution methods
reduces their specification to three simple choices:
\begin{itemize}
	\item \textbf{The choice of behaviour mapping $\Phi$.}
	      The behaviour mapping encodes which aspect of the model $f$ should be explained.
	      In many cases, this is simply the identity function,
	      which leads to an explanation of the model output.
	      However, it can also be some function of $f$, such as its dataset loss \citep{casalicchio2019},
	      the loss at a specific point \citep{lundberg2019a}, or its variance \citep{song2016}.
	\item \textbf{The choice of removal operators $P_{T}$.}
	      The operator $P_T(f)$ maps the function $f \in \mathcal{F}$ 
		  onto a subset of $\mathcal{F}$ containing those functions that are independent of all variables in $T$.
	      In other words, the subset operators define \emph{how features are removed} from the model.
	\item \textbf{The choice of aggregation coefficients $\alpha_{S}^{T}$.}
	      These coefficients define how the behaviour $\Phi(P_{T}(f))$ for different sets of removed features $T$
	      should be \emph{aggregated into a single number}
	      that represents the attribution to the subset of features $S$.
\end{itemize}

Note that these three choices correspond to the choices of the XBR framework \citep{covert2021}.
However, whereas XBR allows for different types of aggregation methods,
we narrow down the scope to those methods that produce a real-valued attribution score
for each feature or subset of features.
This will allow us to develop a more rigorous mathematical theory
about the behaviour of removal-based attribution methods.

A simple example of a removal-based attribution method is the occlusion method \citep{zeiler2014}.
Occlusion was originally proposed for image data, but can easily be applied to tabular data as well.
For a given baseline color
(\textit{e.g.}~the dataset mean, which for natural color images is usually a shade of gray),
and window size $s$, an $s \times s$ square patch is removed from the image by replacing it
with the baseline color.
This square patch then systematically moves across the image,
while the change in output is monitored.
The attribution value for a given pixel is then the average change in output
when that pixel is included in the window.
A simple modification of this method for tabular data
can be devised by replacing each feature of an input vector
with its average value over the dataset and recording the change in output.
To show that occlusion is a RBAM, we need to show that there is a set of removal operators $P_{S}$
such that the occlusion method is a linear combination of functions $P_{S}(f)$.
It is easy to see that the behaviour mapping for occlusion is the identity mapping: $\Phi(f) = f$.
The removal operators can be defined by replacing the features in $S$ with a constant:
\begin{equation*}
	P_{S}(f)(\mathbf{x}) = f(\mathbf{c}_{S},\mathbf{x}_{\overline{S}})
\end{equation*}
for some constant vector $\mathbf{c} = (c, \dots, c)$.
It is easy to see that these operators fulfill the requirements of a removal operator.
We now need to show that the occlusion value for a given feature $i$ is a linear combination of these functions.
For any subset $T \subseteq [d]$ that corresponds to the $s \times s$ square patch at a certain location,
the difference in output is $P_{\emptyset}(f) - P_{T}(f)$.
Denote $\mathcal{T} := \{ T \subseteq [d] \mid T \text{ corresponds to an $s \times s$ patch at some location in the image} \}$.
We then have:
\begin{equation*}
	\alpha_{T}^{i} = \begin{cases}
		1              & \mbox{ if } T = \emptyset                          \\
		-\frac{1}{n_{i}} & \mbox{ if } i \in T \mbox{ and } T \in \mathcal{T} \\
		0               & \mbox{ otherwise.}
	\end{cases}
\end{equation*}
where $n_{i}$ is the total number of patch locations that includes $i$.
When applying occlusion to tabular data, the ``patches'' are simply singleton sets $T = \{ i \}$,
and $n_{i} = 1, \forall i \in [d]$.
Note that occlusion is a feature attribution method, so $\alpha_{T}^{S} = 0, \forall |S| > 1$.

Another simple example is the \textit{leave-one-covariate-out} (LOCO) method for feature importance
in tabular data \citep{kohavi1997}.
This method works as follows:
denote the original dataset as $X$.
For any subset of features $S \subseteq [d]$,
we denote $X_{\overline{S}}$ as the dataset
consisting of all columns in the original dataset $X$
except for the columns $i \in S$.
For each feature $i \in [d]$,
LOCO trains a model on $X_{\overline{i}}$ and the performance of this model is recorded.
The LOCO feature importance value of a feature $i$ is then the difference in performance
when $i$ is included vs.\ when it is not.
Note that this method is designed for a very different purpose than occlusion:
Whereas occlusion performs \textit{attribution} for a single model, LOCO measures \textit{feature importance},
which should be interpreted as the \textit{usefulness} of a specific feature
for training a given type of model.
Consequently, LOCO is typically used as a feature selection method
before training the final model,
whereas occlusion is used as an explanation method for an already-trained model.
Despite this difference in interpretation, it is easy to see that LOCO is also a RBAM.
LOCO differs from occlusion in its removal operators and the behaviour mapping.
Denote the set of matrices over $\mathbb{R}$ for any finite number of rows and columns as
$\mathcal{M} := \bigcup_{m,n \in \mathbb{N}} \mathbb{R}^{m \times n}$.
We can then denote the learning algorithm as a function $\Psi: \mathcal{M} \rightarrow \mathcal{F}$.
Using this notation, we can define the removal operators as:
\begin{equation*}
	P_{S}(f) = \Psi(X_{\overline{S}})
\end{equation*}
\textit{i.e.}~the features in $S$ are removed by training a new model
on the dataset consisting of all columns not in $S$.
It is easy to see that these operators satisfy the requirements
for being valid removal operators.
The behaviour mapping $\Phi$ is in this case not the identity mapping,
but some mapping that quantifies the performance of a model $f$.
This is typically the negative average loss of $f$:
\begin{equation*}
	\Phi(f) = -\mathbb{E}_{X}[l(f(X), y(X))]
\end{equation*}
where $l$ is a loss function like the cross-entropy or \textit{mean squared error} (MSE) loss,
and $y(\mathbf{x})$ is the ground-truth label of the datapoint $\mathbf{x}$.
Note that this behaviour mapping maps any function $f$ to a constant,
implying that this is a global removal-based attribution method.
This quantity can be estimated by computing the empirical loss of $f$
on a held-out test set.
The coefficients $\alpha_{T}^{S}$ are identical to those of the occlusion method.
Note that, in contrast to occlusion, the explained behaviour $\Phi(f)$
for LOCO is the expected loss (or \textit{risk}) of $f$
if it is trained on the full dataset.
This reflects two fundamental differences of the LOCO method w.r.t.\ occlusion:
\begin{itemize}
	\item LOCO provides an explanation of the \textit{performance} of a model,
	      whereas occlusion explains the output.
	      This is reflected by the difference in behaviour mapping.
	      A feature might have a large impact on the model output when it is included or removed,
	      without having a large impact on its performance.
	      Such a feature would have a large occlusion value but a small LOCO value.
	\item LOCO provides information about the learning algorithm that
	      produces the model, whereas occlusion explains a specific trained model.
	      This is reflected by the difference in removal operators.
	      For example, if LOCO attributions are computed for a random forest model,
	      then a feature with a large LOCO value can be interpreted as having a large impact
	      on the performance of random forests in general,
	      rather than any one specific random forest model.
\end{itemize}

We can identify a set of sufficient conditions on $m$
and the removal operators $P_{S}$
such that the resulting attribution method is a simple RBAM.
These sufficient conditions can be used to determine if a given method is a RBAM
without needing to derive the exact coefficients $\alpha_{S}^{T}$.

\begin{restatable}{proposition}{proprbamsufficientconditions}
	\label{prop:fdfi-rbam-sufficient-conditions}
	Assume $m: \mathcal{F} \times 2^{[d]} \rightarrow \mathcal{F}$ is an attribution method.
	If there exists a set of removal operators $\{ P_{S} \mid S \subseteq [d] \}$
	such that:
	\begin{itemize}
		\item For each $S \subseteq [d], \mathbf{x} \in \mathcal{X}$,
		      the attribution $m(f,S)(\mathbf{x})$ is completely determined by the values
		      $\{ P_{S}(f)(\mathbf{x}) \mid S \subseteq [d] \}$,
		      \textit{i.e.}~there exists a vector function $m_{S}: \mathbb{R}^{2^{d}} \rightarrow \mathbb{R}$
		      such that $\forall \mathbf{x} \in \mathcal{X}: m(f,S)(\mathbf{x}) = m_{S}((P_{S}(f)(\mathbf{x}) \mid S \subseteq [d]))$
		\item $m(\cdot, S)$ is a linear function,
		      \textit{i.e.}~for any $f,g \in \mathcal{F}, S \subseteq [d], \alpha \in \mathbb{R}$:
		      \begin{align*}
			      m(\alpha f, S) & = \alpha m(f,S)   \\
			      m(f+g,S)       & = m(f,S) + m(g,S)
		      \end{align*}
		\item The removal operators are linear operators,
		      \textit{i.e.}~for any $f,g \in \mathcal{F}, S \subseteq [d], \alpha \in \mathbb{R}$:
		      \begin{align*}
			      P_{S}(\alpha f) & = \alpha P_{S}(f)      \\
			      P_{S}(f + g)    & = P_{S}(f) +  P_{S}(g)
		      \end{align*}
	\end{itemize}
	then $m$ is a simple removal-based attribution method.
\end{restatable}
\Cref{prop:fdfi-rbam-sufficient-conditions} makes it easier to prove that certain attribution methods are valid RBAMs,
as it suffices to show that both the feature removal operations and attribution method itself are linear,
and that the method is fully defined by the value of $f$ after certain subsets of features have been removed.
For example, these sufficient conditions make it possible to prove under certain conditions that LIME \citep{ribeiro2016}
is a RBAM without having to derive the exact expression for the coefficients $\alpha_{T}^{i}$
(note that LIME is also a feature attribution method, implying $\alpha_{T}^{S} = 0, \forall |S| > 1$).
LIME works by removing features in a similar way as occlusion, \textit{i.e.}~by replacing them with some baseline value.
For tabular data, this implies that the removal operators $P_{S}$ are identical to those of occlusion.
Next, LIME constructs a synthetic dataset by removing subsets of features from the query point $\mathbf{x}$.
The labels of this synthetic dataset are the outputs of the model $f$ for each of the synthetic input samples.
A linear model is then trained on this synthetic dataset.
This linear model can be viewed as a local approximation of the black box model $f$ around the query point $\mathbf{x}$.
The coefficients of this linear model are the attribution values of the corresponding features.
It is easy to see that the output of LIME is entirely determined by the values $\{ P_{S}(f)(\mathbf{x}) \mid S \subseteq [d] \}$.
Furthermore, the removal operators are obviously linear.
We can also see that $m(\cdot, i)$ is a linear function:
if $f = \alpha f_{1} + \beta f_{2}$ for some functions $f_{1},f_{2} \in \mathcal{F}$ and constants $\alpha,\beta \in \mathbb{R}$,
then the labels of the synthetic dataset will also be linear combinations
of the labels for $f_{1}$ and $f_{2}$.
Therefore, the coefficients of the linear model that is trained on this dataset will simply
follow the same linear combination.
This shows that the three conditions are fulfilled, implying that LIME on tabular data is indeed a RBAM.
Note that when the method is applied to image data,
the image is first segmented using an image segmentation algorithm,
after which the segments are treated as individual ``interpretable features.''
However, the number and locations of the segments can vary image per image.
This implies that the number of ``features'' does not remain constant across different instances.
Therefore, the derivation above is no longer valid, and LIME in this case is no longer guaranteed to be a RBAM.

\subsection{Internal Consistency}
In this section, we introduce two forms of internal consistency
for any removal-based method,
stating that any variable that has no influence on the explained behaviour when removed
should receive no attribution.
To express this idea formally,
we first define a \textit{locally independent variable} of a removal-based attribution method:

\begin{definition}[Locally independent variable]
	Assume $m$ is a removal-based attribution method
	with removal operators $\{ P_{T} \mid T \subseteq [d] \}$
	and behaviour mapping $\Phi: \mathcal{F} \rightarrow \mathcal{F}$.
	Let $f \in \mathcal{F}, \mathbf{x} \in \mathcal{X}$.

	A variable $X_{i}, i \in [d]$ is \textbf{locally independent}
	with respect to $m(f, \cdot)$ at $\mathbf{x}$
	if adding or removing the variable has no influence on any of the values
	$\Phi(P_{T}(f))(\mathbf{x})$:
	\begin{equation*}
		\forall T \subseteq [d] \setminus i: \Phi(P_{T \cup i}(f))(\mathbf{x}) = \Phi(P_{T}(f))(\mathbf{x})
	\end{equation*}
\end{definition}
Note that if $m$ is a global RBAM, then any variable that is locally independent
at any given point $\mathbf{x}$ is immediately locally independent
at all points $\mathbf{x}$,
as the global RBAM is a constant function.

As an illustrative example of a locally independent variable,
consider the function $f(x_{1},x_{2}) = \max(x_{1},x_{2})$.
Assume that $m$ is a simple RBAM,
\textit{i.e.}~$\Phi$ is the identity mapping,
and the removal operators $P_{S}$ remove features by replacing them with $0$.
This definition of the removal operators
corresponds to an implicit assumption that a value of $0$ symbolizes the \textit{absence}
of the corresponding feature from the model.
Let $\mathbf{x} = (0, 2)$.
Then $f(\mathbf{x}) = P_{\emptyset}(f)(\mathbf{x}) = P_{1}(f)(\mathbf{x}) = 2$,
and $P_{2}(f)(\mathbf{x}) = P_{1,2}(f)(\mathbf{x}) = 0$.
In other words, adding or removing the feature $X_{1}$
has no influence on the output of $f$ at $\mathbf{x}$,
\textit{regardless of whether feature $X_{2}$ is present or not.}
This is not surprising, as $x_{1}=0$,
\textit{i.e.}~it is already considered ``absent'' in $\mathbf{x}$.
In this case, $X_{1}$ is a locally independent variable of $f$.

Now let $\mathbf{x'} = (1,2)$.
We then still have $P_{\emptyset}(f)(\mathbf{x'}) = P_{1}(f)(\mathbf{x'}) = 2$,
\textit{i.e.}~removing $x'_{1}$ from the original $\mathbf{x'}$ has no influence on $f$
since $x'_{2}$ is the maximum of the two values.
However, we now have $P_{2}(f)(\mathbf{x'}) = 1$,
whereas $P_{1,2}(f)(\mathbf{x'}) = 0$.
In other words, adding $X'_{1}=1$ when $X'_{2}$ is \textit{absent}
does have an influence on $f$.
In this case, $X_{1}$ is not a locally independent variable of $f$.

Using this definition of a locally independent variable,
we can now define two forms of internal consistency
for removal-based attribution methods:
feature-level and interaction-level consistency.
Feature-level consistency states that a locally independent variable
should receive zero attribution.
Interaction-level consistency additionally requires that
any interaction containing a locally independent variable
also receives zero attribution.
These two definitions reflect the idea that a variable
that has no influence on the behaviour if removed
cannot have any direct influence or interaction effect
with any other variable.
\begin{definition}[Internal Consistency]
	A removal-based attribution method $m$
	is \textbf{feature-level consistent} if,
	for any $X_{i}, i \in [d], \mathbf{x} \in \mathcal{X}$, $f \in \mathcal{F}$
	such that $X_{i}$ is locally independent with respect to $m(f, \cdot)$ at $\mathbf{x}$:
	\begin{equation*}
		m(f,i)(\mathbf{x}) = 0
	\end{equation*}
	$m$ is \textbf{interaction-level consistent} if we additionally have:
	\begin{equation*}
		m(f,S \cup i)(\mathbf{x}) = 0, \forall S \subseteq [d]
	\end{equation*}
\end{definition}
It is easy to see that interaction-level consistency implies feature-level consistency.
If a removal-based attribution method is feature-level consistent,
then any locally independent variable of that method
receives zero attribution.
If the attribution method is also interaction-level consistent,
then any interaction effect that contains a locally independent variable
also receives zero attribution.
Whereas feature-level consistency can be viewed as a minimal requirement
of consistency for \textit{any} removal-based attribution method,
interaction-level consistency is only required if the attribution method
is supposed to quantify the \textit{pure} interaction effect between variables.
If, however, the method is designed to quantify the \textit{total} effect
of subsets of variables, it can break this requirement.
An example of such a method is the \textit{group Shapley value}
or \textit{generalized Shapley value} \citep{marichal2007},
which computes a variant of the Shapley value for a subset of players $S$
by treating the subset as a single, unified player.
The result is a quantification of the \textit{total} effect
of the subset $S$.
If $S$ contains a (locally) neutral variable along with one or more
non-neutral variables, then this total effect can still be nonzero.
This method therefore is not interaction-level consistent,
which simply reflects that the group Shapley value for a subset $S$
should not be interpreted as a pure interaction effect \citep{janizek2021}.

Note that the local independence of a variable depends on the specific removal-based attribution method:
a variable that is locally independent with respect to one method
might not be locally independent with respect to another.
In our previous example, if features were removed by replacing them with a value of $1$
instead of $0$, then $X_{1}$ would not have been locally independent at $\mathbf{x} = (0,2)$.
A locally independent variable of a method $m$ can therefore be interpreted
as a variable that has no influence on the behaviour
$\Phi(f)$ at $\mathbf{x}$ \textit{if features are removed according to $m$}.
This explains why feature-level and interaction-level consistency are forms
of \textit{internal} consistency of removal-based attribution methods.

\subsection{Examples}
\label{sec:fdfi-rbam-examples}

We have seen before how occlusion \citep{zeiler2014},
LOCO \citep{kohavi1997} and tabular LIME \citep{ribeiro2016}
are examples of removal-based attribution methods
with different definitions for the removal operators and behaviour mapping.
In the following paragraphs, we will give an overview of commonly used
removal operators and behaviour mappings.
Next, we will provide more examples of removal-based attribution methods
and their corresponding removal operators, behaviour mappings and aggregation coefficients.
As our definition of RBAMs is based on the XBR framework,
some of the notation and terminology in the following overview is based on \citep{covert2021}.

The simplest behaviour mapping that we will consider is the \textit{local output mapping}.
This is simply the identity mapping: $\Phi(f) = f$.
Explanation methods that use the local output behaviour mapping
provide an explanation of the behaviour of the model $f$ for a specific instance.
We will also call these methods \textit{simple RBAMs.}
Another local behaviour mapping is the \textit{local loss mapping}.
For a given loss function $l$, this corresponds to the mapping:
$\Phi(f)(\mathbf{x}) = -l(f(\mathbf{x}),y)$
where $y$ is the true label of $\mathbf{x}$.
Explanations that use this behaviour mapping
should be interpreted as targeting the performance of the model $f$,
rather than the behaviour itself.
The local loss mapping can easily be transformed into a global variant
by taking the average loss across the dataset,
resulting in the \textit{dataset loss mapping}:
$\Phi(f) = -\mathbb{E}_{X,Y}[l(f(X),Y)]$.
Finally, another global behaviour mapping is the \textit{variance mapping:}
$\Phi(f) = \text{Var}_{X}[f(X)]$.
In certain cases, the variance of the function $P_{T}(f)$
can be viewed as the mean squared error loss between $P_{T}(f)$
and $f$ itself.
This can be interpreted as the degree to which
the function behaviour
can be explained using only
the features in $\overline{S}$.

We have already encountered the \textit{single baseline}
and the \textit{retraining} removal operators:
$P_{T}(f)(\mathbf{x}) = f(\mathbf{c}_{T},\mathbf{x}_{\overline{T}})$
and
$P_{T}(f) = \Psi(X_{\overline{T}})$,
respectively,
where $X_{T}$ is a matrix of training data containing only features in $T$
and $\Psi$ is the learning algorithm that produces a model $f$
given this matrix $X_{T}$.
The other removal operators that we will consider in this work
are all based on marginalization of features using different distributions:
\begin{itemize}
	\item Marginal: $P_{T}(f)(\mathbf{x}) = \mathbb{E}_{X_{T}}[f(X_{T},\mathbf{x}_{\overline{T}})]$
	\item Product of Marginals (PM): $P_{T}(f)(\mathbf{X}) = \mathbb{E}_{\prod_{i \in [d]}\Pr[X_{i}]}[f(X_{T},\mathbf{X}_{\overline{T}})]$ where $[d]$ is the set of features and $\Pr[X_{i}]$ is the marginal distribution of feature $i$.
	\item Uniform: $P_{T}(f)(\mathbf{x}) = \mathbb{E}_{\mathcal{U}[X]}[f(X_{T},\mathbf{x}_{\overline{T}})]$ where $\mathcal{U}[X]$ is the uniform distribution over the domain of the input variables $X$.
	\item Conditional: $P_{T}(f)(\mathbf{x}) = \mathbb{E}[f(X) \mid X_{\overline{T}} = \mathbf{x}_{\overline{T}}]$
	\item Interventional: $P_{T}(f)(\mathbf{x}) = \mathbb{E}[f(X) \mid do(X_{\overline{T}} = \mathbf{x}_{\overline{T}})]$, where $do(X=\mathbf{x})$ is the $do$-operator \citep{pearl2012} using an assumed causal graph $G$.
	\item Joint Baseline Distribution (JBD): $P_{T}(f)(\mathbf{x}) = f(\mathbf{z}_{T},\mathbf{x}_{\overline{T}})\Pr[\mathbf{z}_{T},\mathbf{x}_{\overline{T}}]$
	\item Random Joint Baseline Distribution (RJBD): $P_{T}(f)(\mathbf{x}) = \mathbb{E}_{X}[f(X_{T},\mathbf{x}_{\overline{T}})\Pr[X_{T},\mathbf{x}_{\overline{T}}]]$
	\item Tree Distribution (TD): this is an approximation of the conditional distribution
		using the structure of a set of learned decision trees.
		See \citet{lundberg2019} for more details.
\end{itemize}

A large number of explanation methods are based on the Shapley value.
These methods are defined using a characteristic function $v(S), S \subseteq [d]$
which computes some desired behaviour of $f$ when only the features in $S$ are known.
This can typically be expressed as a removal operator $P_{\overline{S}}$ followed by a behaviour mapping $\Phi$:
\begin{equation*}
v(S) = \Phi(P_{\overline{S}}(f))(\mathbf{x})
\end{equation*}
The Shapley value for a given characteristic function $v$ is then defined as:
\begin{equation*}
	\phi_{i}(v) = \frac{1}{d}\sum_{S \subseteq [d] \setminus i}\binom{d-1}{|S|}^{-1}(v(S \cup i) - v(S))
\end{equation*}
It is easy to see that this is a linear combination of $v(S)$ for different subsets $S$.
Therefore, any Shapley-based method is also a RBAM,
provided that the characteristic function $v$ can indeed be expressed as
the composition of a removal operator and a behaviour mapping.
As all Shapley-based RBAMs use the same aggregation coefficients,
we will call these the \textit{Shapley aggregation coefficients}.
Examples of Shapley-based RBAMs using different removal operators and behaviour mappings
are given in \Cref{tbl:fdfi-shapley-methods}.

Although the Shapley aggregation coefficients are arguably the most popular option at the time of writing,
they are not the only one.
We have already encountered an alternative definition when discussing occlusion.
Another alternative definition for $\alpha_{T}^{i}$ is given by:
\begin{equation*}
\alpha_{T}^{i} = \begin{cases}
	1 & \mbox{if } T = \emptyset \\
	-1 & \mbox{if } T = \{ i \}\\
	0 & \mbox{otherwise.}
\end{cases}
\end{equation*}
This corresponds to removing a single feature from the input
and measuring the difference in behaviour.
These aggregation coefficients are used by the LOCO method \citep{kohavi1997},
as well as PFI \citep{breiman2001a} and conditional PFI \citep{strobl2008}.
Note that this definition is identical to the definition for occlusion
with a patch size of 1.
Analogously, we can define aggregation coefficients
that correspond to \textit{including} a single feature and
measuring the difference in behaviour
when none of the features are available:
\begin{equation*}
\alpha_{T}^{i} = \begin{cases}
	-1 & \mbox{if } T = [d] \\
	1 & \mbox{if } T = [d] \setminus \{ i \}\\
	0 & \mbox{otherwise.}
\end{cases}
\end{equation*}
This approach is used by the \textit{univariate predictors} method \citep{guyon2003}.
An overview of examples 
of methods that do not use Shapley aggregation coefficients
is given in \Cref{tbl:fdfi-non-shapley-methods}.

Interaction attribution methods
are defined by the fact that they attribute non-zero values to feature subsets
with more than one element.
This implies that their defining characteristics are captured
entirely within the choice of aggregation coefficients.
For this reason, they are typically introduced
without specifying a choice for removal operators or behaviour mapping.
Examples of interaction attribution methods include the Shapley interaction index (see \Cref{sec:game-theory-interaction-indices}),
Banzhaf interaction index \citep{patel2021},
Shapley-Taylor interaction index \citep{sundararajan2020a},
Faith-SHAP \citep{tsai2023},
Faith-BANZHAF \citep{tsai2023},
n-Shapley Values \citep{bordt2023},
and q-interaction SHAP \citep{hiabu2023}.

\subsection{Functional axioms}
\label{sec:fdfi-functional-axioms}
Recently, it has been shown that certain implementations of Shapley values can have unintuitive properties,
such as assigning non-zero attributions to features that are not referenced by the model \citep{chen2020}.
This is surprising, as it seems to be in conflict with the Null axiom of the Shapley value \citep{shapley1953},
which states that a player who has no influence on the cooperative game should receive no payoff.
However, if we inspect the details of the attribution method more carefully,
we can see that an independent variable of the function is not guaranteed to be a null player
in the cooperative game for which the Shapley values form the attributions.
This shows that the game-theoretic axioms defined for Shapley values
are not sufficient for describing the behaviour of attribution methods,
as they do not necessarily line up
with the intuition practitioners might have when applying these methods
to their machine learning models.

To tackle this problem,
we introduce a set of \textit{functional axioms}.
These axioms are defined in terms of the actual model or behaviour being explained,
rather than a cooperative game that is derived from it.
This makes them more suitable as descriptions of a method's behaviour,
as they can directly be interpreted as guarantees about the explanations
provided by the method,
whereas to correctly interpret the implications of game-theoretic axioms
requires reasoning about the axioms themselves as well as the translation of the machine learning
setting into a cooperative game simultaneously.

Note that these functional axioms should not be interpreted as \textit{necessary}
requirements for any attribution method to be considered valuable.
For example, the ``unintuitive'' behaviour concerning the Null axiom mentioned above
has led to discussions among researchers about whether this behaviour is actually desirable or not \citep{chen2020}.
One can argue that this depends on the specific use case and the expectations that
are made for the explanation.
For example, if a given feature $X_{i}$ is not referenced by the model,
but another feature $X_{j}$ that is heavy correlated to $X_{i}$ is referenced,
then in some contexts it can be preferable to attribute a nonzero value
to $X_{i}$ as it can be viewed as having an \textit{indirect effect}
on the model through its correlation with $X_{j}$.
Therefore, this debate is also called the
\textit{indirect influence debate}.
Rather than expressing necessary conditions for the validity of explanation methods,
therefore, the functional axioms should be viewed more as
\textit{descriptions of guaranteed behaviour} of a given explanation method.
This can make it easier for practitioners to decide if a given method is suitable to them,
for example by clarifying the choice between a method that does or does not
account for indirect influence.

\subsubsection{Functional null}
\label{sec:fdfi-functional-null}
The game-theoretic Null axiom states that a \textit{null player},
\textit{i.e.}~a player who has no influence on the outcome of the game,
should receive a payoff of zero.
As the ``players'' in the machine learning context are the variables of the model $f$,
the intuitive analogue of a null player
is a variable that is independent of the model $f$.
However, an independent variable of $f$ is not guaranteed to be a null player.
Consider the function $f(x_{1},x_{2}) = x_{1}$.
$X_{2}$ is clearly an independent variable of $f$.
Assume that $X_{1},X_{2}$ are jointly normally distributed with mean zero
and some non-zero covariance $\sigma$,
and $m(f,i)$ is the conditional Shapley method:
\begin{align*}
	\begin{pmatrix}X_{1} \\
		X_{2}
	\end{pmatrix} & \sim \mathcal{N}
	\begin{bmatrix}
		\begin{pmatrix}
			0 \\
			0
		\end{pmatrix}\!\!,
		\begin{pmatrix}
			1      & \sigma \\
			\sigma & 1      \\
		\end{pmatrix}
	\end{bmatrix}                                                                 \\
	m(f,1)(\mathbf{x})      & = \frac{1}{2}[v(\{ 1,2 \}) - v(\{ 2 \}) + v(\{ 1 \}) - v(\emptyset)] \\
	m(f,2)(\mathbf{x})      & = \frac{1}{2}[v(\{ 1,2 \}) - v(\{ 1 \}) + v(\{ 2 \}) - v(\emptyset)] \\
	v(S)                    & = \mathbb{E}[f(X) \mid X_{S} = \mathbf{x}_{S}]
\end{align*}
As $m$ is the Shapley value for the cooperative game $v$,
it adheres to the game-theoretic Null axiom by definition.
However, if $\mathbf{x} = (0,1)$, we then have:
\begin{align*}
	m(f, 2)(0,1) & = \frac{1}{2}[f(0,1) - \mathbb{E}[f(X_{1},X_{2}) \mid X_{2} = 1] + \mathbb{E}[f(X_{1},X_{2}) \mid X_{1} = 0] - \mathbb{E}[f]] \\
	             & = \frac{1}{2}[0 - \sigma + 0 - 0]                                                                                             \\
	             & = -\sigma / 2
\end{align*}
since $\mathbb{E}[X_{1} \mid X_{2} = x_{2}] = \mu_{1} + \frac{\sigma_{12}}{\sigma_{2}}(x_{2} - \mu_{2}) = \sigma x_{2}$.
Therefore, if the correlation $\sigma$ between the two variables is nonzero,
then the attribution to $X_{2}$ will also be nonzero at $\mathbf{x} = (0,1)$,
despite $X_{2}$ being an independent variable of $\Phi(f)$
and $m$ adhering to the game-theoretic Null axiom.
As this example shows that a method that adheres to the game-theoretic
Null axiom can still produce
nonzero attributions for independent variables of $f$,
we introduce the \textit{functional null axiom:}
\begin{definition}[Functional Null]
	A removal-based attribution method adheres to the functional null axiom if
	for any function $f$ with independent variable $X_{i}$ and subset $S \subseteq [d]$:
	\begin{equation*}
		m(f,S \cup i) = 0
	\end{equation*}
\end{definition}
Intuitively, this means that if $X_i$ is an independent variable of the model $f$,
then the attribution for any set containing $X_i$ is zero everywhere.
The example given above shows that the conditional Shapley method does not adhere to the
Functional Null axiom, despite adhering to the game-theoretic Null axiom.
\subsubsection{Functional dummy}
In the context of cooperative game theory,
a Dummy player is a player $i \in N$ such that:
\begin{equation*}
	\forall S \subseteq N \setminus i: v(S \cup i) = v(S) + v(i)
\end{equation*}
This is a player that may have a non-zero contribution to the outcome,
but does not interact with any other players.
The Dummy Partnership axiom for interaction indices
states that any interaction containing such a Dummy player
should receive zero attribution.
The functional equivalent of a Dummy player is an additive variable:
an additive variable $X_{i}$ of $f$ is precisely a variable
that does not interact with any other variable in $f$.
One might then expect that the Dummy Partnership axiom then implies
that any interaction effect containing an additive variable of $f$
should also be zero.
However, similarly to Null players and independent features,
an additive feature is not guaranteed to be a Dummy player in the cooperative game.

Let $f(x_{1},x_{2}) = x_{1} + x_{2}$,
where $X_{1}$ and $X_{2}$ are distributed in the same way as in the previous example.
Assume $m(f,S)$ is now the Shapley interaction index using the conditional distribution,
\textit{i.e.}~the cooperative game for a given point $\mathbf{x}$
is defined as:
\begin{align*}
	v(\{ 1,2 \}) & = f(x_{1},x_{2}) = x_{1} + x_{2}                                   \\
	v(\{ 1 \})   & = \mathbb{E}[f(X_{1},X_{2}) \mid X_{1} = x_{1}] = x_{1} - \sigma/2 \\
	v(\{ 2 \})   & = \mathbb{E}[f(X_{1},X_{2}) \mid X_{2} = x_{2}] = x_{2} - \sigma/2 \\
	v(\emptyset) & = \mathbb{E}[f(X_{1},X_{2})] = 0
\end{align*}
Computing the interaction effect for $S = \{ 1,2 \}$ using the Shapley interaction index,
we get:
\begin{align*}
	\phi_{S}(v) & = \sum_{T \subseteq N \setminus S}\frac{|S|}{|S| + |T|}\binom{n}{|S|+|T|}^{-1}\Delta_{S}v(T) \\
	            & = \frac{2}{2} \binom{2}{2}^{-1} \Delta_{S}v(\emptyset)                                       \\
	            & = \Delta_{S}v(\emptyset)                                                                     \\
	            & = v(\{ 1,2 \}) - v(\{ 1 \}) - v(\{ 2 \}) + v(\emptyset)                                      \\
	            & = x_{1} + x_{2} - x_{1} + \sigma/2 - x_{2} + \sigma/2 + 0                                    \\
	            & = \sigma
\end{align*}
Therefore, when $\sigma$ is nonzero, the interaction effect according to the
Shapley interaction index
between these two additive variables will also be nonzero.
Analogously to the Functional Null axiom,
we therefore also introduce the Functional Dummy axiom:
\begin{definition}[Functional Dummy]
	A removal-based attribution method adheres to the functional dummy axiom if
	for any function $f$
	with additive variable $X_i$ and any subset $S \subseteq [d]$ with $S \neq \emptyset$:
	\begin{equation*}
		m(f, S \cup i) = 0
	\end{equation*}
	and $m(f,i)$ depends only on $X_{i}$.
\end{definition}
Intuitively, this means that if $f$ is additive in $X_i$,
the attribution function for $X_i$ is independent of all other variables,
and any higher-order attributions containing $X_i$ are zero.

Note that the functional dummy axiom does not imply the functional null axiom,
in contrast to their game-theoretic counterparts.
Consider the following counterexample.
Assume an attribution method $m$ is defined as:
\begin{equation*}
	m(f,S) = \begin{cases}
		1 & \mbox{if } |S| = 1 \\
		0 & \mbox{otherwise}
	\end{cases}
\end{equation*}
Then $m$ is a trivial univariate removal-based attribution method
that simply assigns an attribution value of 1 to each variable,
independent of the function or the input point.
This is a valid RBAM:
if we define the behaviour mapping as being the identity mapping,
and we choose removal operators such that $P_{[d]}(f)(\mathbf{x}) = 1, \forall \mathbf{x} \in \mathcal{X}$,
then it is easy to see that this method indeed fits the definition of a RBAM.
This method adheres to the functional dummy axiom:
the condition holds trivially for any variable,
additive or not.
However, this method does not adhere to the functional null axiom:
if $X_{i}$ is an independent variable of $f$, then the attribution for $X_i$ will not be 0.

\subsubsection{Functional symmetry}
Two players $i,j$ in a cooperative game $v$ are called \textit{symmetric}
if their marginal contributions to any coalition $S \subseteq N \setminus \{ i,j \}$
are identical:
\begin{equation*}
	\forall S \subseteq N \setminus \{ i,j \}: v(S \cup i) = v(S \cup j)
\end{equation*}
Such players can be viewed as completely interchangeable,
which is why the Symmetry axiom states that any two symmetric players
should receive the same attribution.
Symmetric players in a cooperative game can be linked to symmetric variables
in a function $f$:
it seems reasonable to impose that symmetric variables
that have the same value in $\mathbf{x}$
should also receive the same attribution at $\mathbf{x}$,
as their contributions are arguably identical.
However, analogously to the previous axioms,
the game-theoretic Symmetry axiom does not guarantee that symmetric variables
that have the same value at $\mathbf{x}$
are also symmetric players in the cooperative game.

As a simple example, consider the function
\begin{equation*}
	f(\mathbf{x}) = x_{1} + x_{2}
\end{equation*}
The variables $X_{1}$ and $X_{2}$ are symmetric in $f$,
\textit{i.e.}~swapping their values in any vector $\mathbf{x}$
has no influence on $f(\mathbf{x})$.
Now let $m$ be the Shapley value-based method
where the cooperative game $v$ is defined
using a baseline vector $\mathbf{b} = (b_{1}, b_{2})$:
\begin{align*}
	v(\emptyset) & = f(b_{1}, b_{2}) \\
	v(\{ 1 \})   & = f(x_{1}, b_{2}) \\
	v(\{ 2 \})   & = f(b_{1}, x_{2}) \\
	v(\{ 1,2 \}) & = f(x_{1}, x_{2}) \\
\end{align*}
As this method is a Shapley value, it satisfies the game-theoretic Symmetry axiom
by definition.
The attributions for $X_{1}$ and $X_{2}$ at any point $\mathbf{x}$
are then:
\begin{align*}
	m(f,1)(\mathbf{x}) & = \frac{1}{2}[f(x_{1},x_{2}) - f(b_{1},x_{2}) + f(x_{1}, b_{2}) - f(b_{1},b_{2})] \\
	m(f,2)(\mathbf{x}) & = \frac{1}{2}[f(x_{1},x_{2}) - f(x_{1},b_{2}) + f(b_{1}, x_{2}) - f(b_{1},b_{2})]
\end{align*}
As $f$ is symmetric, these two attribution values
will be identical for any point $\mathbf{x} = (x,x)$
if $b_{1} = b_{2}$.
However, if these baseline values are not identical,
then the attribution values for $X_{1}$ and $X_{2}$ are not guaranteed to be identical
even if the two variables have equal values in $\mathbf{x}$.
In that case, we have an attribution method that adheres to the game-theoretic
Symmetry axiom, but that does not produce identical attributions
for symmetric variables with identical values.
We therefore define the Functional Symmetry axiom:
\begin{definition}[Functional Symmetry]
	A removal-based attribution method $m$ adheres to the functional symmetry axiom if
	for any function $f \in \mathcal{F}$ with symmetric variables $X_i,X_j$
	and subset $S \subseteq [d] \setminus i,j$:
	\begin{equation*}
		\forall \mathbf{x} \in \mathcal{X}: x_i = x_j \implies m(f,S \cup i)(\mathbf{x}) = m(f, S \cup j)(\mathbf{x})
	\end{equation*}
\end{definition}
Indeed, the example given above adheres to the game-theoretic Symmetry axiom,
as it is a Shapley value,
but not to the Functional Symmetry axiom.

\subsubsection{Functional anonymity}
The final functional axiom we introduce is Functional Anonymity,
which is the functional equivalent of the game-theoretic Anonymity axiom.
This axiom states that a relabeling of the players in the cooperative game
should have no influence on the payoff vector.
We define the Functional Anonymity axiom as follows:
\begin{definition}[Functional Anonymity]
	A removal-based attribution method $m$ adheres to the functional anonymity axiom if
	for any permutation $\pi \in \Pi([d])$,
	function $f \in \mathcal{F}$,
	subset $S \subseteq [d]$
	and point $\mathbf{x} \in \mathcal{X}$:
	\begin{equation*}
		m(\pi f, S)(\pi \mathbf{x}) = m(f, \pi S)(\mathbf{x})
	\end{equation*}
\end{definition}
Intuitively this means that, if we reorder the variables in $\mathbf{x}$ according to $\pi$,
and then compute the attributions for $S$ for the function that simply undoes the reordering of $\pi$ and then computes $f$,
then we should obtain the same result
as if we simply reordered the variables in $S$
and then computed attributions for those variables for the original function $f$.
This can be viewed as a generalization of the Functional Symmetry axiom.
Indeed, if $m$ adheres to Functional Anonymity, then it is easy to prove that it also adheres
to Functional Symmetry.

We now give an illustrative example of the functional anonymity axiom.
Let the function $f: \mathbb{R}^3 \rightarrow \mathbb{R}$ be defined as follows:
\begin{equation*}
	f(\mathbf{x}) = x_1 + x_2^2 + x_3^3
\end{equation*}
Note that this function has no symmetric variables.
Consider the following definition for the attribution method $m$:
\begin{align*}
	m(f,1)(\mathbf{x}) & = f(x_1, b_2, b_3) \\
	m(f,2)(\mathbf{x}) & = f(b_1, x_2, b_3) \\
	m(f,3)(\mathbf{x}) & = f(b_1, b_2, x_3)
\end{align*}
This is a simple univariate ($k=1$) method that computes attributions for feature $i$
by removing all other features and using the resulting function output as the attribution score.
This can be viewed intuitively as the ``isolated influence'' of feature $i$.
Features are removed by replacing them with a certain baseline value $b_i$.

We now define the feature permutation $\pi = ( 2,3,1 )$.
Let $S = \{1\}$.
Then $\pi S = \{2\}$.
To avoid confusing notation,
we will denote the specific point
at which we want to compute attributions as
$\mathbf{z}$.
We then have:
\begin{align*}
	m(f,\pi S)(\mathbf{z})     & = m(f,2)(\mathbf{z})            \\
	                           & = f(b_1,z_2,b_3)                \\
	                           & = b_1 + z_2^2 + b_3^3           \\
	m(\pi f,S)(\pi \mathbf{z}) & = m(\pi f,1)(\pi\mathbf{z})     \\
	                           & = m(\pi f,1)(( z_2, z_3, z_1 )) \\
	                           & = (\pi f)(z_2, b_2, b_3)        \\
	                           & = f(b_3, z_2, b_2)              \\
	                           & = b_3 + z_2^2 + b_2^3
\end{align*}
These two quantities will be equal when $b_1 = b_2 = b_3$. In other words,
the method is anonymous if the baseline value of each feature is identical,
\textit{i.e.}~features are removed in the same way, independently of their position in the arguments.

\section{Additive Functional Decomposition}
\label{sec:additive-functional-decomposition}

In this section,
we will define and study the properties of additive functional decompositions.
We also introduce the concept of a dependency structure,
which is a generalization of the ANOVA structure introduced in \cite{hooker2004a}.
We show that any real-valued function $f$ has a unique minimal dependency structure,
which is related to the concept of minimality in additive functional decompositions.
We then introduce the \textit{canonical additive decomposition} (CAD)
and show that any additive functional decomposition is indeed a CAD.
Next, we study a number of useful properties that a CAD can have, and give a few examples.

\begin{definition}[Additive functional decomposition]
        \label{def:additive-functional-decomposition}
	Assume $\mathcal{F}$ is a linear space of functions
	$f: \mathcal{X} \rightarrow \mathbb{R}$,
	where $\mathcal{X}$ is a (possibly infinite) hyperrectangle in $\mathbb{R}^d$.
	A set of operators $G := \{g_S: \mathcal{F} \rightarrow \mathcal{F}|S \subseteq [d]\}$
	is an \textbf{additive functional decomposition} on $\mathcal{F}$ if,
	for any $f \in \mathcal{F}, S \subseteq [d]$:
	\begin{itemize}
		\item Completeness: $f = \sum_{S \subseteq [d]}g_S(f)$
		\item Independence: $g_S(f)$ is independent of $X_{\overline{S}}$
	\end{itemize}
	We will call the operators $g_S$ the \textbf{decomposition operators} of $G$.
	We denote the set of valid additive functional decompositions on $\mathcal{F}$ as $\mathcal{D}_{\mathcal{F}}$.
	For a given function $f \in \mathcal{F}$,
	we call the set $G(f) := \{g_S(f)|S \subseteq [d]\}$ an \textbf{additive decomposition of $f$.}
\end{definition}
This definition views the decomposition as a set of $2^d$ functions,
one for each subset of features $S \subseteq [d]$.
The sum of these $2^d$ functions must equal the original function in all points $\mathbf{x}$,
which is a minimal requirement for this set to be considered a decomposition of $f$.
Finally, every function $g_S(f)$ can only depend on the corresponding variables $S$.
Note that an additive decomposition of a function $f \in \mathcal{F}$ always exists.
We can simply define:
\begin{equation*}
	g_S(f) := \left\{
	\begin{array}{ll}
		f & \mbox{if } S = [d] \\
		0 & \mbox{otherwise}
	\end{array}
	\right.
\end{equation*}
We will call this decomposition the \textit{trivial decomposition} of $f$.

We now introduce the concept of the \textit{dependency structure}.
A dependency structure of a function $f$ is defined as a collection of sets
that correspond to the maximal non-zero components in a given decomposition.
These sets correspond to non-additive interaction effects in the function $f$,
as modeled by the decomposition.
However, different functional decompositions can result in different
non-zero components,
meaning that some interaction effects are perhaps not fundamental in $f$,
but only when $f$ is decomposed in a particular way.
Therefore, an interesting dependency structure is the \textit{minimal}
dependency structure of $f$.
This is the minimal set of subsets that any additive decomposition must contain
in order to model $f$.
The minimal dependency structure of $f$ does not depend on any specific decomposition,
implying that this dependency structure describes the non-additive interactions that are
fundamentally present in $f$.

\begin{definition}[Dependency structure]
	$f \in \mathcal{F}$ has \textbf{dependency structure}
	$\mathcal{S} \subseteq 2^{[d]}$
	(notation: $\mathcal{S} \in \text{DS}(f)$) if :
	\begin{equation*}
		f = \sum_{S \in \mathcal{S}}g_S(f)
	\end{equation*}
	for $G := \{g_S|S \subseteq [d]\} \in \mathcal{D}_{\mathcal{F}}$ such that:
	\begin{equation*}
		\forall S,T \in \mathcal{S}: S \subseteq T \implies S=T
	\end{equation*}
	\textit{i.e.}~$\mathcal{S}$ is an antichain in the lattice formed by the subsets of $[d]$ and the relation $\subseteq$.
	Given $\mathcal{S},\mathcal{U} \in \text{DS}(f)$.
	$\mathcal{S}$ is a \textbf{sub-dependency structure} of $\mathcal{U}$
	(notation: $\mathcal{S} \leq_\text{DS} \mathcal{U}$)
	if $\forall S \in \mathcal{S}: \exists U \in \mathcal{U}:S \subseteq U$.
	A dependency structure $\mathcal{S}$ for $f$ is \textbf{minimal} if
	$\forall \mathcal{U} \in \text{DS}(f): \mathcal{U} \leq_\text{DS} \mathcal{S} \implies \mathcal{U} = \mathcal{S}$.
\end{definition}

As a simple example, consider the function $f(\mathbf{x}) = x_{1} + x_{2} * x_{3}$.
This function can be written as the sum of two functions $f_{1}(\mathbf{x}) = x_{1}$
and $f_{2,3}(\mathbf{x}) = x_{2} * x_{3}$.
$f_{1}$ only depends on $X_{1}$ and $f_{2,3}$ only depends on $X_{2}$ and $X_{3}$.
Therefore, the collection $\{ \{ 1 \}, \{ 2,3 \} \}$ is a valid dependency structure of $f$.
Alternatively, we could write $f = f'_{1} + f'_{1,2,3}$,
with $f'_{1}(\mathbf{x}) = 2x_{1}$ and $f'_{1,2,3}(\mathbf{x}) = x_{2} * x_{3} - x_{1}$.
However, $\{ \{ 1 \}, \{ 1,2,3 \} \}$ is not a valid dependency structure,
as this collection does not adhere to the required antichain property,
since $\{ 1 \} \subset \{ 1,2,3 \}$.
Note that the existence of the trivial decomposition implies that
$\{[d]\}$ is a valid dependency structure of any function $f \in \mathcal{F}$.
As mentioned above, a more interesting dependency structure therefore
is the minimal dependency structure.
As the sub-dependency structure relation $\leq_{\text{DS}}$ is a partial order
on the set of dependency structures $\text{DS}(f)$
and this set is non-empty and finite,
it follows immediately that at least one minimal element must exist.
In the following proposition, we prove that this minimal element is also unique,
which allows us to speak of \textit{the} minimal dependency structure
of a function $f$.
This also implies that the subsets present in the minimal dependency structure
represent fundamentally irreducible interactions between variables,
implying that this minimal dependency structure is a fundamental property of $f$,
independent of any specific way of decomposing $f$.

\begin{restatable}{proposition}{propmdsunique}
	For any function $f \in \mathcal{F}$, there exists a unique minimal dependency structure $\mathcal{S} := \text{MDS}(f)$.
\end{restatable}

Finally,
the following theorem shows a link between independent variables of a function $f$
and the minimal dependency structure of $f$.

\begin{restatable}{proposition}{propmdsindependent}
	\label{prop:fdfi-prop-mds-independent}
	Given $f \in \mathcal{F}, \text{MDS}(f) = \mathcal{S}$.
	Then $X_{i}$ is an independent variable of $f \iff i \notin \bigcup \mathcal{S}$.
\end{restatable}

\subsection{The Canonical Additive Decomposition}
\label{sec:fdfi-canonical-additive-decomposition}
In this section, we introduce the \textit{canonical additive decomposition} (CAD).
This is a generalization of the \textit{general decomposition formula} introduced in \citet{kuo2010}.
This definition gives a general structure to the additive functional decomposition introduced above.
We then show that any additive functional decomposition is indeed a canonical additive decomposition, justifying the name.
Next, we introduce a number of useful properties that a canonical additive decomposition can have,
which can be used to gain insight into the general characteristics of a specific decomposition.
Finally, we give some examples of additive functional decompositions and discuss their properties.

\begin{definition}[Canonical Additive Decomposition]
	Assume $\mathcal{F}$ is a linear function space.
	Let $\mathcal{P} = \{P_T: \mathcal{F} \rightarrow \mathcal{F}|T \subseteq [d]\}$
	be a set of removal operators on $\mathcal{F}$.

	We then define the \textbf{canonical additive decomposition}
	$G := \{g_S: \mathcal{F} \rightarrow \mathcal{F}|S \subseteq [d]\}$
	with removal operators $\mathcal{P}$
	as follows:
	\begin{align}
		g_S(f) & := P_{\overline{S}}(f) - \sum_{T \subset S}g_T(f) \label{eqn:cad-def-1}             \\
		       & = \sum_{T \subseteq S}(-1)^{|S|-|T|} P_{\overline{T}}(f) \label{eqn:fdfi-cad-def-2}
	\end{align}
\end{definition}
Recall that the removal operators $P_T$ can be viewed as \textit{removing} the features $T$
from the function $f$
by producing a new function $P_T(f)$
which is independent of $X_T$.
This gives an intuitive interpretation to \Cref{eqn:cad-def-1}:
we first remove all features that are not in $S$,
and then subtract all behaviour in $f$ that was already accounted for in strict subcomponents.
From \Cref{eqn:cad-def-1}, we derive the \textit{summation property} of the CAD:
\begin{equation*}
	P_{\overline{T}}(f) = g_T(f) + \sum_{S \subset T}g_S(f) = \sum_{S \subseteq T}g_S(f)
\end{equation*}
Using this property, we can show that the two definitions of the CAD are indeed equivalent:
\begin{align*}
	P_{\overline{T}}(f) & = \sum_{S \subseteq T}g_S(f)                            \\
	g_S(f)              & = \sum_{T \subseteq S}(-1)^{|S|-|T|}P_{\overline{T}}(f)
\end{align*}
where we apply \Cref{thm:fdfi-inclusion-exclusion-principle} using $g(A) := P_{\overline{A}}(f)$ and $f(S) := g_S(f)$.

The following theorem shows that the definition of the canonical additive decomposition
covers precisely all additive functional decompositions,
justifying the name \textit{canonical} additive decomposition.
\begin{restatable}{theorem}{thmcadexistenceuniqueness}
	\label{thm:fdfi-cad-existence-uniqueness}
	A set of operators $G := \{g_S: \mathcal{F} \rightarrow \mathcal{F}|S \subseteq [d]\}$
	is an additive functional decomposition
	if and only if there exists a set of removal operators $\mathcal{P}$
	such that $G$ is the canonical additive decomposition
	with removal operators $\mathcal{P}$.
\end{restatable}

As the removal operators used in the definition of
removal-based attribution methods
are identical to those used in the definition of
the canonical additive decomposition,
we can use them to define a CAD.
We call this CAD the \textit{corresponding functional decomposition}:
\begin{definition}
	Assume $m: \mathcal{F} \times 2^{[d]}$ is a removal-based attribution method
	with removal operators $\{ P^{m}_{T} \mid T \subseteq 2^{[d]} \}$.
	We then call the CAD
	\begin{equation*}
		\forall S \subseteq [d]: g_{S}^{m}(f) := P_{\overline{S}}^{m}(f) - \sum_{T \subset S}g_{T}(f)
	\end{equation*}
	the \textbf{corresponding functional decomposition} of $m$.
\end{definition}

\subsection{Examples}
\label{sec:fdfi-cad-examples}
In this section, we go over a few examples of additive decompositions and their properties.
The first decomposition we consider is the ANOVA decomposition \citep{hoeffding1948}.
The ANOVA decomposition of a function $f \in \mathcal{F}$
can be expressed as a CAD with removal operators:
\begin{equation*}
	P_T^{\text{ANOVA}}(f)(\mathbf{x}) := \int_{[0,1]^{|T|}} f(\mathbf{x}_{\overline{T}},\mathbf{z}_{T}) d \mathbf{z}
\end{equation*}
\textit{i.e.}~the features $X_{j}, j \in T$ are removed from $f$
by taking the integral of $f$ with respect to $X_{T}$,
producing a function that no longer depends on $X_{T}$.
The ANOVA decomposition is a thoroughly-studied decomposition with many interesting properties \citep{hoeffding1948,roosen1995,hooker2007,owen2013b},
and makes the implicit assumption that variables are distributed uniformly and independently between 0 and 1.
This is the corresponding decomposition of the IME method \citep{strumbelj2014}.

Another well-known and related decomposition is the \textit{anchored decomposition} \citep[Appendix A]{owen2013}:
\begin{equation*}
	P^{\text{Anchored}}_{T}(f)(\mathbf{x}) := f(\mathbf{x}_{\overline{T}},\mathbf{z}_{T})
\end{equation*}
for some fixed baseline vector $\mathbf{z} \in \mathcal{X}$.
It is easy to see that the removal operators in the ANOVA decomposition
can be computed as the expected value of the removal operators of the anchored decomposition,
where the baseline vector is sampled from the uniform distribution on the unit hypercube.
Because both decompositions are linear, this implies that the ANOVA components
can also be approximated as the expected value of the anchored components
for baseline vectors sampled from the uniform distribution on the unit hypercube \citep{merrick2020}.
Both the ANOVA and anchored decompositions can also be expressed in the general decomposition formula
framework proposed by \citet{kuo2010}.

The next decomposition is the marginal decomposition for a given reference distribution $\mathcal{D}^{\text{ref}}$.
This is the canonical additive decomposition with removal operators:
\begin{equation*}
	P_T^{\mathcal{D}_{\text{ref}}}(f)(\mathbf{x}) := \mathbb{E}_{X \sim \mathcal{D}^{\text{ref}}}[f(\mathbf{x}_{\overline{T}}, X_T)]
\end{equation*}
This decomposition is a generalization of the ANOVA decomposition:
the ANOVA decomposition can be viewed as the marginal decomposition
where the reference distribution $\mathcal{D}^{\text{ref}}$
is the uniform distribution over the unit hypercube.
Other choices for the reference distribution include:
\begin{itemize}
	\item The input distribution $\mathcal{D}^{\text{inp}}$: leads to the \textit{partial dependence decomposition} \citep{gevaert2022}.
	      This is the corresponding decomposition of SHAP,
	      assuming that the assumption of independence between variables is made
	      to make the computation tractable \citep{lundberg2017}.
	\item Product of marginal distributions $\mathcal{D}^{\text{PM}}$: $\Pr_{\mathcal{D}^{\text{PM}}}[X=\mathbf{x}] = \prod_{i=1}^d \Pr_{\mathcal{D}^{inp}}[X_{i}=x_i]$.
	      This is the corresponding distribution of the QII method \citep{datta2016}.
      \item Singleton distribution $\Pr[X=\mathbf{x}] = \mathbf{1}[\mathbf{x} = \mathbf{z}]$ for some vector $\mathbf{z} \in \mathcal{X}$: leads to the anchored decomposition with baseline $\mathbf{z}$.
\end{itemize}

Note that, analogously to the ANOVA decomposition,
any marginal decomposition can be computed as an average of anchored decompositions,
where the anchor points $\mathbf{z}$ are sampled according to $\mathcal{D}^{\text{ref}}$ \citep{merrick2020}.
Note also the subtle difference between the input distribution and the product of marginal distributions.
The difference between these two removal operators
is the fact that the product of marginals does not take into account any correlations
between the variables that are being removed,
whereas the partial dependence decomposition does.
This means that if multiple variables $j \in T$ are removed from $f$ using these operators,
then $P_T^{\text{PM}}$ will integrate out each variable \textit{separately} using its univariate marginal distribution,
whereas $P_T^{\text{PD}}$ integrates out \textit{all} of the variables in $T$ using the joint marginal distribution.
In practice, this corresponds to sampling each variable separately ($P_T^{\text{PM}}$)
or first sampling a full row from the data and then filling in the corresponding variables ($P_T^{\text{PD}}$).
Note also that, although the marginal decomposition using the product of marginal distributions
can be expressed as a general decomposition formula,
the partial dependence decomposition cannot \citep{kuo2010}.
This is because the removal operator for a subset of features $P^{\mathcal{D}^{\text{inp}}}_{T}$
is not equal to the composition of the removal operators for each feature separately.
This shows that the canonical additive decomposition is indeed a \textit{strict} generalization
of the general decomposition formula.

A final example is the \textit{conditional decomposition.}
This decomposition removes variables from $f$ by taking a conditional expected value,
conditioning on the other variables:
\begin{equation*}
	P^{\text{Cond}}_T(f)(\mathbf{x}) := \mathbb{E}_{X}[f(X) \mid X_{\overline{T}} = \mathbf{x}_{\overline{T}}]
\end{equation*}
This way of removing variables prevents the function $f$ from being evaluated on unrealistic datapoints,
in contrast to $P_j^{\text{PM}}$ and $P_S^{\text{PD}}$.
This is the corresponding decomposition of the conditional Shapley values method,
or SHAP if the assumption of independence between features is not made \citep{lundberg2017}.
Analogously to the partial dependence decomposition,
the conditional decomposition cannot be expressed as a generalized decomposition formula.

\subsection{Properties of Additive Functional Decompositions}
We now go over a few interesting properties
that an additive functional decomposition can have.
These properties will be useful in later sections,
where we connect the CAD to cooperative game theory and removal-based attribution.
The first, most basic property we consider is \textit{linearity}:

\begin{definition}
	An additive functional decomposition $G = \{ g_{S} \mid S \subseteq [d] \}$ is \textbf{linear} if:
	\begin{equation*}
		\forall f_1,f_2 \in \mathcal{F}, \alpha,\beta \in \mathbb{R}, S \subseteq [d]: g_S(\alpha f_1 + \beta f_2) = \alpha g_S(f_1) + \beta g_S(f_2)
	\end{equation*}
\end{definition}

This can be viewed as a sensible property for an additive functional decomposition to have,
and indeed all of the examples given in \Cref{sec:fdfi-cad-examples} are linear.
It is easy to show that a CAD is linear if and only if
its removal operators are linear operators.
Assume first that the subset operators $P_{S}$ of $G$ are linear.
Then, for $\alpha, \beta \in \mathbb{R}, f_{1}, f_{2} \in \mathcal{F}$:
\begin{align*}
	g_S(\alpha f_1 + \beta f_2) & = \sum_{T \subseteq S}(-1)^{|S|-|T|}P_{\overline{T}}(\alpha f_1 + \beta f_2)                                                    \\
	                            & = \sum_{T \subseteq S}(-1)^{|S|-|T|}(\alpha P_{\overline{T}}(f_1) + \beta P_{\overline{T}}(f_2))                                \\
	                            & = \alpha\sum_{T \subseteq S}(-1)^{|S|-|T|}P_{\overline{T}}(f_1) + \beta\sum_{T \subseteq S}(-1)^{|S|-|T|} P_{\overline{T}}(f_2) \\
	                            & = \alpha g_S(f_1) + \beta g_S(f_2)
\end{align*}
Now assume that the decomposition itself is linear.
From the summation property, we have:
\begin{equation*}
	\forall T \subseteq [d]: P_T(f) = \sum_{S \subseteq \overline{T}}g_S(f)
\end{equation*}
Therefore:
\begin{align*}
	P_T(\alpha f_1 + \beta f_2) & = \sum_{S \subseteq \overline{T}}g_S(\alpha f_1 + \beta f_2)                                     \\
	                            & =\sum_{S \subseteq \overline{T}}\left(\alpha g_S(f_1) + \beta g_S(f_2)\right)                                 \\
	                            & = \alpha \sum_{S \subseteq \overline{T}}g_S(f_1) + \beta \sum_{S \subseteq \overline{T}}g_S(f_2) \\
	                            & = \alpha P_T(f_1) + \beta P_T(f_2)                                                               \\
\end{align*}
A more interesting property that a given decomposition may or may not have
is \textit{minimality}.
Intuitively, a minimal decomposition
does not introduce unnecessary terms \citep{kuo2010}.
In other words, the behaviour of $f$ is captured in the lowest possible order terms.
A decomposition that is not minimal can be viewed as ``introducing'' interactions
between variables that are not truly necessary to explain the full behaviour
of $f$. Formally:

\begin{definition}
	For any subset $S \subseteq [d]$ and function $f \in \mathcal{F}$,
	denote the set of components $\{g_T(f):S \subseteq T \subseteq [d]\}$
	as the \textbf{super-$S$ components} of $G(f)$.
	An additive functional decomposition $G \in \mathcal{D}_{\mathcal{F}}$ is \textbf{minimal} if
	for any $S \subseteq [d]$ and $f \in \mathcal{F}$,
	we have that
	if an additive decomposition $H(f)$ of $f$ exists such that
	all super-$S$ components of $H(f)$ are zero,
	then all super-$S$ components of $G(f)$ are also zero.
\end{definition}

The concept of minimality is tightly linked to the minimal dependency structure
of a function $f$.
An example of a decomposition that is not minimal is the conditional decomposition:
\begin{equation*}
	P_{T}(f)(\mathbf{x}) = \mathbb{E}_{X}[f(X) \mid X_{\overline{T}} = \mathbf{x}_{\overline{T}}]
\end{equation*}
To illustrate this, consider the simple function $f(x_{1},x_{2}) = x_{1}$.
Obviously, the minimal dependency structure for this function is $\{ \{ 1 \} \}$.
Now assume that $X_{1} \sim \mathcal{N}(0,1)$,
and $X_{2}$ and $X_{1}$ are identical with probability 1.
It is easy to verify that the conditional decomposition of $f$ is then:
\begin{align*}
	f_\emptyset(x_{1},x_{2}) & = 0    \\
	f_1(x_{1},x_{2})         & = x_1  \\
	f_2(x_{1},x_{2})         & = x_2  \\
	f_{12}(x_{1},x_{2})      & = -x_1
\end{align*}
This is a valid decomposition, as the sum of the functions $f_{S}$ is equal to $f$
and each function $f_{S}$ does not depend on any variables not in $S$.
However, this decomposition is clearly not minimal.
The following lemma shows that
only subsets of sets in the minimal dependency structure of $f$
can correspond to non-zero components in a minimal decomposition $G(f)$.
\begin{restatable}{lemma}{lemmaminimalcadmds}
	\label{lemma:fdfi-minimal-cad-mds}
	Assume $G \in \mathcal{D}_{\mathcal{F}}$ is a minimal additive functional decomposition,
	$f \in \mathcal{F}$.
	Then:
	\begin{equation*}
		\forall S \subseteq [d]:
		g_{S}(f) \neq 0 \implies (\exists T \in \text{MDS}(f): S \subseteq T)
	\end{equation*}
\end{restatable}
From the definition of the minimal dependency structure,
we can easily see that this lemma extends to any dependency structure:
for any $\mathcal{S} \in \text{DS}(f)$,
the non-zero components of a minimal decomposition of $f$
correspond to subsets of sets in $\mathcal{S}$.
This can in turn be used to prove that the components of a minimal decomposition
corresponding to the sets in a minimal dependency structure
\textit{must} be non-zero:

\begin{restatable}{proposition}{propminimalcadmds}
	\label{prop:fdfi-minimal-cad-mds}
	Assume $G \in \mathcal{D}_{\mathcal{F}}$ is a minimal additive functional decomposition, $f \in \mathcal{F}$, $\text{MDS}(f) = \mathcal{S}$. Then:
	\begin{equation*}
		\forall S \in \mathcal{S}: g_S(f) \neq 0
	\end{equation*}
\end{restatable}

Proofs for both of these properties can be found in \Cref{sec:proofs}.
From \Cref{lemma:fdfi-minimal-cad-mds} and \Cref{prop:fdfi-minimal-cad-mds},
we can see that a decomposition $G$ is minimal if and only if
for any $f \in \mathcal{F}$,
the maximal subsets corresponding to nonzero components of $G(f)$
form precisely the minimal dependency structure of $f$.
As minimality is an interesting property for an additive decomposition to have,
we will now derive necessary and sufficient conditions
in order for $G$ to be minimal.
We first introduce the concept of \textit{independence preservation.}

\begin{definition}
	An additive functional decomposition $G$ is \textbf{independence-preserving}
	if for any function $f$ with independent variable $X_{i}$
	and subset $S \subseteq [d] \setminus i$:
	\begin{equation*}
		g_{S \cup i}(f) = 0
	\end{equation*}
\end{definition}

Intuitively, a decomposition is independence-preserving if an independent variable of $f$
also has no direct effects or interaction effects according to $G(f)$.
We can identify the following necessary and sufficient conditions
on the removal operators $P_{T}$ for independence preservation:

\begin{restatable}{proposition}{propnecsuffindependencepreservation}
	Given an additive functional decomposition $G$ with removal operators $\{ P_{T} \mid T \subseteq [d] \}$.
	$G$ is independence-preserving
	if and only if
	for any function $f$ with independent variable $X_{i}$ and any subset $T \subseteq [d] \setminus i$:
	\begin{equation*}
		P_{T}(f) = P_{T \cup i}(f)
	\end{equation*}
\end{restatable}

The following proposition shows that independence preservation
is a necessary condition for the minimality of the decomposition.

\begin{restatable}{proposition}{minimalimpliesip}
	\label{prop:fdfi-minimal-implies-ip}
	If an additive functional decomposition $G$ is minimal, then it is independence-preserving.
\end{restatable}

Additionally, if the decomposition is linear,
then independence preservation of the removal operators
is also a sufficient condition for minimality.

\begin{restatable}{proposition}{linearipimpliesminimal}
	\label{prop:fdfi-linear-ip-implies-minimal}
	If an additive functional decomposition $G$ is independence-preserving and linear,
	then that decomposition is minimal.
\end{restatable}

\Cref{prop:fdfi-minimal-implies-ip,prop:fdfi-linear-ip-implies-minimal}
imply that a linear additive decomposition is minimal
if and only if its removal operators
are independence-preserving.

Up to this point, we have assumed that the removal operators $P_{T}$
are defined for all subsets $T$ of the total feature set $[d]$.
This reflects the fact that multiple features can be removed from the function $f$
as a group,
for example using a joint marginal or conditional distribution.
However, if the removal operators $P_{T}$ remove the features \textit{separately},
then the removal operators only need to be defined for the singletons:
$\{ P_{j} \mid j \in [d] \}$.
We formalize this notion as follows:

\begin{definition}
	Given a set of removal operators $\{ P_{T} \mid T \subseteq [d] \}$.
	The operators $P_{T}$ are called \textbf{separable} if:
	\begin{equation*}
		\forall T,T' \subseteq [d]: P_{T} \circ P_{T'} = P_{T'} \circ P_{T} = P_{T \cup T'}
	\end{equation*}
\end{definition}

This immediately implies that a definition of the removal operators
for the singletons $j \in [d]$ is indeed sufficient,
as $\forall T \subseteq [d]: P_{T} = \Pi_{j \in T}P_{j}$.
As the separable operators are commutative,
the expression $\Pi_{j \in T}P_{j}$ is well-defined.
Note also that separability implies that the removal operators are projections:
\begin{equation*}
	P_{T} \circ P_{T} = P_{T \cup T} = P_{T}
\end{equation*}
If the removal operators are both linear and separable,
then we obtain the general decomposition formula from \citet{kuo2010}.
This shows that the canonical additive decomposition
is indeed a generalization of the general decomposition formula.
This also implies that an additive functional decomposition 
with linear, separable removal operators
is minimal,
as shown in Theorem 3.1 of \citet{kuo2010}.
Although separability greatly simplifies the analysis of additive decompositions,
many existing decompositions are unfortunately not separable.
For example, the marginal decomposition using the joint marginal distribution
and the conditional decomposition
both have inseparable removal operators.

If the components of an additive decomposition
can be viewed as interaction effects,
then it can be interesting to investigate
the \textit{consistency} of those interactions.
One way to formalize this
is to investigate if the interaction effects
remain identical if the decomposition is performed twice.
More specifically, we define \textit{idempotence} of a decomposition $G$
as follows:
\begin{definition}
	An additive functional decomposition $G$ is \textbf{idempotent} if:
	\begin{equation*}
		\forall f \in \mathcal{F}, S, T \subseteq [d]: g_T(g_S(f)) =
		\left\{
		\begin{array}{ll}
			g_S(f) & \mbox{if } T = S  \\
			0      & \mbox{otherwise.}
		\end{array}
		\right.
	\end{equation*}
\end{definition}
If a decomposition $G$ is idempotent,
then decomposing a component $g_{S}(f)$ using $G$
results in a semi-trivial decomposition where $g_{S}(g_{S}(f))$
is the only non-zero component,
implying that the effect between the variables $S$
is in a sense ``pure,'' at least according to the decomposition $G$.
This also implies that the decomposition operators $g_{S}$ are projections,
although the reverse implication does not hold.
Note that idempotence does not necessarily imply
that the interaction effects given by a decomposition $G$
are ``pure'' in the intuitive sense.
A simple example is the trivial decomposition.
It is easy to show that the trivial decomposition is indeed idempotent,
although it can hardly be interpreted as ``pure,''
since it models all of the behaviour of any function $f$
as a single simultaneous interaction effect between all variables.
Additionally imposing minimality on the decomposition does not solve the problem:
one can easily design a decomposition such that the feature subsets
in the minimal dependency structure of any function $f$
correspond to the only non-zero components,
implying that all strict subsets of sets in the minimal dependency structure are zero.
In such a decomposition,
direct effects of variables that are also involved in higher-order interactions
will always be modeled as interaction effects.
Although idempotence is still an interesting property of a decomposition,
a precise definition of \textit{interaction purity}
that fully captures this intuition is still an open problem to the best of our knowledge.
The following proposition introduces necessary and sufficient conditions
for an additive decomposition to be idempotent.

\begin{restatable}{proposition}{idempotencenecsuffconditions}
	\label{prop:fdfi-idempotence-nec-suff-conditions}
	An additive functional decomposition $G$ is idempotent if and only if,
	for any subsets $T,S \subseteq [d]$:
	\begin{equation*}
		P_{T}(g_{S}(f)) = \begin{cases}
			g_{S}(f) & \mbox{ if } S \cap T = \emptyset    \\
			0        & \mbox{ if } S \cap T \neq \emptyset
		\end{cases}
	\end{equation*}
\end{restatable}
Note that the first condition for $P_{T}(g_{S}(f))$ is a weaker form
of independence preservation:
as $g_{S}(f)$ is by definition independent of any variables not in $S$,
it is also independent of all of the variables in $T$.
The first condition states that $P_{T}$ therefore leaves this function unchanged.
The second condition can be linked to the \textit{annihilating property}
of the general decomposition formula from \citet{kuo2010}:
\begin{equation*}
	\forall j \in S: P_j(g_S(f)) = 0
\end{equation*}
\textit{i.e.}~if a variable $j \in S$ removed from $g_S(f)$,
the result is a constant zero function.
This can be seen as a generalized form of centeredness of the functions $g_S(f)$:
if $P_j$ is defined as integrating out the variable $j$,
then this property states that the expected value of each non-empty component must be zero.
It is easy to see that the second condition in \Cref{prop:fdfi-idempotence-nec-suff-conditions}
is equivalent to the annihilating property
if the projections $P_{S}$ are separable.

In general, it can be difficult to prove that a decomposition is idempotent,
as it requires an analysis of the removal operators $P_{S}$
and the decomposition operators $G_{S}$ simultaneously.
Fortunately, if we assume that the removal operators
(and therefore also the decomposition itself) are linear,
then the necessary and sufficient conditions for idempotence
become much simpler.
We show this in the following proposition.
The proof of this proposition can be found in \Cref{sec:proofs}.

\begin{restatable}{proposition}{linearseparableiffidempotent}
	\label{prop:fdfi-linear-separable-iff-idempotent}
	Assume $G$ is a linear additive functional decomposition
	with removal operators $P_{S}$.
	Then the removal operators $P_{S}$ are separable
	if and only if $G$ is idempotent.
\end{restatable}

One example of a decomposition that is not idempotent
and indeed has inseparable removal operators
is the partial dependence decomposition.
The removal operators for the PDD are defined as follows:
\begin{equation*}
	P_{T}(f)(\mathbf{x}) = \mathbb{E}_{X \sim \mathcal{D}^{\text{inp}}}[f(X_{T},\mathbf{x}_{\overline{T}})]
\end{equation*}
Because the joint marginal distribution of all of the variables in $S$
is used to marginalize out the variables,
these removal operators are indeed inseparable.
Assume $f(x_{1},x_{2}) = x_{1} + x_{2} + x_{1}x_{2}$.
For the sake of simplicity, we also assume that
$\mathbb{E}[X_{1}] = \mathbb{E}[X_{2}] = 0$.
We then have:
\begin{align*}
	g_{\emptyset}(f)(x_{1},x_{2}) & = \mathbb{E}[X_{1}X_{2}]              \\
	g_{1}(f)(x_{1},x_{2})         & = x_{1} - \mathbb{E}[X_{1}X_{2}]      \\
	g_{2}(f)(x_{1},x_{2})         & = x_{2} - \mathbb{E}[X_{1}X_{2}]      \\
	g_{1,2}(f)(x_{1},x_{2})       & = x_{1}x_{2} + \mathbb{E}[X_{1}X_{2}] \\
\end{align*}
If we decompose $g_{1,2}(f)$ a second time using the same decomposition,
we get:
\begin{align*}
	g_{\emptyset}(g_{1,2}(f)) & = 2 \mathbb{E}[X_{1}X_{2}]            \\
	g_{1}(g_{1,2}(f))         & = -\mathbb{E}[X_{1}X_{2}]             \\
	g_{2}(g_{1,2}(f))         & = -\mathbb{E}[X_{1}X_{2}]             \\
	g_{1,2}(g_{1,2}(f))       & = x_{1}x_{2} + \mathbb{E}[X_{1}X_{2}] \\
\end{align*}
Note that if we would instead use the product of marginal distributions,
\textit{i.e:}
\begin{equation*}
	P_{T}(f)(\mathbf{x}) = \mathbb{E}_{X \sim \mathcal{D}^{\text{PM}}}[f(X_{T},\mathbf{x}_{\overline{T}})]
\end{equation*}
then the term $\mathbb{E}[X_{1}X_{2}]$ would be replaced by
$\mathbb{E}[X_{1}]\mathbb{E}[X_{2}] = 0$,
and the decomposition would be idempotent.
Indeed, it is easy to verify that
the removal operators in this case are also separable.

As we have already discussed the property of independence preservation,
which states that independent variables of $f$ should remain independent
in the decomposition of $f$,
it seems straightforward to define a similar property for additive variables.
Recall that $X_{i}$ is an additive variable of $f$
if $f$ can be written as a sum of two functions:
\begin{equation*}
	f = f_{i} + f_{\overline{i}}
\end{equation*}
where $f_{i}$ depends only on $X_{i}$, and $f_{\overline{i}}$ is independent of $X_{i}$.
Starting from this definition, it seems reasonable to define the property of
\textit{additivity preservation} as follows:

\begin{definition}
	An additive functional decomposition $G$ is \textbf{additivity-preserving} if, for any $f \in \mathcal{F}$ with additive variable $X_i$, we have:
	\begin{equation*}
		\forall S \subseteq [d] \setminus i: S \neq \emptyset \implies g_{S \cup i}(f) = 0
	\end{equation*}
\end{definition}

Intuitively, additivity preservation means that if $X_i$ is an additive variable of $f$,
then all higher-order terms in the decomposition containing $X_i$ will be zero.
This reflects the fact that there are no interactions between $X_i$ and any other variable.
An example of a decomposition that is not additivity-preserving is the conditional decomposition.
This can be shown using the same example that we used to show that this decomposition
is not minimal.
We can identify necessary and sufficient conditions
on the removal operators $P_{S}$
for additivity preservation:

\begin{restatable}{proposition}{additivitypreservingremovalconditions}
	\label{prop:fdfi-additivity-preserving-removal-conditions}
	An additive functional decomposition $G$ with removal operators
	$\{P_S|S \subseteq [d]\}$
	is additivity-preserving
	if and only if for any function $f \in \mathcal{F}$
	with additive variable $X_i$:
	\begin{equation*}
		\forall S,T \subseteq [d] \setminus i: P_{T \cup i}(f) - P_T(f) = P_{S \cup i}(f) - P_S(f)
	\end{equation*}
	We call a set of removal operators that adheres to these conditions
	\textbf{additivity-preserving removal operators.}
\end{restatable}

Similarly to how minimality of the decomposition implies independence preservation,
it is also a sufficient condition for additivity preservation:

\begin{restatable}{proposition}{minimalimpliesadditivitypreserving}
	\label{prop:fdfi-minimal-implies-additivity-preserving}
	Given an additive functional decomposition $G$. If $G$ is minimal, then $G$ is additivity-preserving.
\end{restatable}

Note that despite the fact that any independent variable of $f$
is also an additive variable of $f$ (with $f_{i} = 0$),
additivity preservation does not imply independence preservation.
We can construct a decomposition that is additivity-preserving
but not independence-preserving as follows:
assume $H$ is a minimal decomposition.
Then $H$ is both independence- and additivity-preserving.
Now define $G$ as follows:
\begin{equation*}
	g_{S}(f) = \begin{cases}
		h_{\emptyset}(f) + d & \mbox{ if } S = \emptyset \\
		h_{S}(f) - 1         & \mbox{ if } |S| = 1       \\
		h_{S}(f)             & \mbox{ otherwise.}
	\end{cases}
\end{equation*}
It is easy to see that this is indeed a valid decomposition.
The decomposition only modifies the terms of $H$ corresponding to singletons
and the empty set.
Therefore, it remains additivity-preserving.
However, for any independent variable $X_{i}$ we have $g_{i}(f) = -1$,
which violates independence preservation.

The following property concerns symmetric variables.
\begin{definition}
	An additive functional decomposition $G$ is \textbf{symmetry-preserving} if,
	for any $f$ with symmetric variables $X_i,X_j$
	and subset $S \subseteq [d] \setminus \{i,j\}$:
	\begin{equation*}
		\forall \mathbf{x} \in \mathcal{X}:
		x_i = x_j \implies
		g_{S \cup i}(f)(\mathbf{x}) = g_{S \cup j}(f)(\mathbf{x})
	\end{equation*}
\end{definition}

Intuitively, a decomposition is symmetry-preserving
if the symmetric variables $i$ and $j$
are interchangeable in any component $g_{S}(f)$.
Note that this does not necessarily imply
that any component containing both $i$ and $j$ is in turn
symmetric in these variables.
We can again identify necessary and sufficient conditions
for symmetry preservation
in terms of the removal operators
(proof given in \Cref{sec:proofs}).

\begin{restatable}{proposition}{symmetrypreservationnecsuffconditions}
	\label{prop:fdfi-symmetry-preservation-nec-suff-conditions}
	An additive functional decomposition $G$
	with removal operators $\{ P_{S} \mid S \subseteq [d] \}$
	is symmetry-preserving if and only if
	for any function $f$ with symmetric variables $X_{i},X_{j}$
	and subset $S \subseteq [d] \setminus \{ i,j \}$:
	\begin{equation*}
		\forall \mathbf{x} \in \mathcal{X}:
		x_{i} = x_{j} \implies P_{S \cup i}(f)(\mathbf{x}) = P_{S \cup j}(f)(\mathbf{x})
	\end{equation*}
	We call a set of removal operators that adheres to these conditions
	\textbf{symmetry-preserving removal operators.}
\end{restatable}

Finally, we can generalize symmetry preservation to \textit{anonymity}.
Intuitively, anonymity states that the decomposition is invariant
to a re-ordering or relabeling of the input features of any function $f$.
More specifically, reordering the variables in $\mathbf{x}$ according to $\pi$
and then decomposing the function $\pi f$
results in the same decomposition of $f$
with the components reordered according to $\pi$.

\begin{definition}
	An additive functional decomposition $G$ is \textbf{anonymous} if,
	for any permutation $\pi \in \Pi([d])$,
	function $f \in \mathcal{F}$,
	subset $S \subseteq [d]$
	and point $\mathbf{x} \in \mathcal{X}$:
	\begin{equation*}
		g_S(\pi f)(\pi \mathbf{x}) = g_{\pi S}(f)(\mathbf{x})
	\end{equation*}
\end{definition}

This can be viewed as a type of symmetry in the removal of features by $P_S$,
in the sense that all inputs are removed in a similar way,
regardless of their index.
Indeed, this is also the intuition
behind the following necessary and sufficient conditions
for anonymity in terms of the removal operators
(proof given in \Cref{sec:proofs}):

\begin{restatable}{proposition}{anonymityremovalconditions}
	An additive functional decomposition $G$ with removal operators $\{P_S|S \subseteq [d]\}$ is anonymous if and only if,
	for any permutation $\pi \in \Pi([d])$,
	function $f \in \mathcal{F}$,
	subset $S \subseteq [d]$
	and point $\mathbf{x} \in \mathcal{X}$:
	\begin{equation*}
		P_S(\pi f)(\pi \mathbf{x}) = P_{\pi S}(f)(\mathbf{x})
	\end{equation*}
\end{restatable}

For example, using this property
it is easy to verify that the marginal decomposition is anonymous
if and only if the marginal distributions for all variables
are identical.
This aligns with the intuition that anonymity implies that
all inputs are removed in a similar way.
To show that anonymity implies symmetry preservation,
assume $X_{i},X_{j}$ are symmetric in $f$.
Denote $\pi_{ij}$ as the permutation that simply swaps these two variables.
Then $\pi_{ij}(f) = f$.
If $x_{i} = x_{j}$, then we also have $\pi_{ij}\mathbf{x} = \mathbf{x}$.
Using anonymity, we then have
\begin{equation*}
	g_{S}(\pi_{ij} f)(\pi_{ij} \mathbf{x}) = g_{S}(f)(\mathbf{x}) = g_{\pi_{ij} S}(f)(\mathbf{x})
\end{equation*}
Symmetry preservation then follows from the fact that
$\pi_{ij}(S \cup i) = S \cup j$ for any $S \subseteq [d] \setminus i,j$,
and vice versa.

\section{A unifying framework}
\label{sec:unifying-framework}

In \Cref{sec:additive-functional-decomposition},
we have seen that the removal operators of a removal-based attribution method
can also be used to define an additive functional decomposition,
which we call the \textit{corresponding additive decomposition}
of that attribution method.
In the following sections, we will exploit this connection
to develop a general theory of RBAMs.
This general theory then allows us to
build a taxonomy of RBAMs based on cooperative game theory,
develop a general approximation scheme for RBAMs,
and derive sufficient conditions for the adherence of RBAMs
to the functional axioms introduced in \Cref{sec:fdfi-functional-axioms}.
This section will proceed as follows.
In \Cref{sec:fdfi-rbam-representation-theorem},
we define the \textit{pointwise cooperative game},
a class of cooperative games that is generated by a given combination
of additive decomposition and behaviour mapping.
We then introduce the RBAM representation theorem,
which shows that an attribution method
is a removal-based attribution method
\textit{if and only if} it is a value or interaction index
for a (possibly constant-shifted) pointwise cooperative game.
This implies that a removal-based attribution method
can alternatively be defined by specifying a definition
of a pointwise cooperative game
and a value or interaction index for that game,
instead of the original definition based on removal operators,
behaviour mapping and aggregation coefficients.
Next, we use this connection between removal-based attribution
and cooperative game theory to develop a taxonomy of RBAMs
in \Cref{sec:fdfi-taxonomy}.
In \Cref{sec:fdfi-efficient-computation}, we exploit the mathematical link between RBAMs
and additive functional decomposition
to develop a general class of approximation methods for attribution methods
based on the explicit modeling of the additive decomposition.
Subsequently, in \Cref{sec:fdfi-suff-cond-functional-axioms}, we derive sufficient conditions
for the functional axioms that were previously introduced in \Cref{sec:fdfi-functional-axioms}.
Finally, we revisit some of the examples given in \Cref{sec:fdfi-rbam-examples}
and study their corresponding decompositions and functional axioms in \Cref{sec:fdfi-framework-examples}.

\subsection{RBAM Representation Theorem}
\label{sec:fdfi-rbam-representation-theorem}
In this section, we will develop a fundamental link
between removal-based attribution methods
and cooperative game theory.
We begin by defining the \textit{pointwise cooperative game}.
Then, we will show that any RBAM generates a specific
pointwise cooperative game,
which we will call the \textit{corresponding pointwise cooperative game.}
Finally, we will show that
\textit{any} removal-based attribution method
is a linear function of a possibly constant-shifted
version of its corresponding pointwise cooperative game.
Furthermore, the method is internally consistent
if and only if it is a linear combination of discrete derivatives
of its corresponding pointwise cooperative game,
\textit{i.e.}~it is a marginal contribution value or interaction index
for this game.
As the pointwise cooperative game is entirely defined by
an additive decomposition and a behaviour mapping,
this theorem mathematically shows that the specification of
any internally consistent RBAM can alternatively be viewed
as a choice of pointwise cooperative game
and a set of constants that define the linear function.
We begin with the definition of a pointwise cooperative game:
\begin{definition}
        Given an additive functional decomposition $G \in \mathcal{D}_{\mathcal{F}}$
        with removal operators $\{ P_{T} \mid T \subseteq [d] \}$
	and a behaviour mapping $\Phi: \mathcal{F} \rightarrow \mathcal{F}$.
	Denote $\mathcal{G}(S)$ as the set of cooperative games on the set of players $S$.
	The \textbf{pointwise cooperative game}
	$v_{G}^{\Phi}: \mathcal{F} \times \mathcal{X} \rightarrow \mathcal{G}([d])$
	of $G$ and $\Phi$
	is then defined as:
	\begin{align*}
                v_{G}^{\Phi}(f,\mathbf{x})(S) 
                &= \Phi \left(P_{\overline{S}}(f)\right)(\mathbf{x}) - \Phi\left(P_{[d]}(f)\right)(\mathbf{x})\\
                &= \Phi \left( \sum_{T \subseteq S} g_{T}(f) \right)(\mathbf{x}) - \Phi(g_{\emptyset}(f))(\mathbf{x})
	\end{align*}
\end{definition}
The equality in this definition follows from the
summation property of the canonical additive decomposition:
\begin{equation*}
	P_{\overline{T}}(f) = \sum_{S \subseteq T}g_{S}(f)
\end{equation*}
Note that if $\Phi$ is a global behaviour mapping,
the cooperative game $v_{G}^{\Phi}(f,\mathbf{x})$
also does not depend on the argument $\mathbf{x}$,
and can therefore be written more succinctly as
$v_{G}^{\Phi}(f)$.
It is easy to see that the pointwise cooperative game
for a given decomposition $G$ and behaviour mapping $\Phi$
at a specific function $f$ and input point $\mathbf{x}$
is indeed a valid cooperative game,
\textit{i.e.}~$v_{G}^{\Phi}(f,\mathbf{x})(\emptyset) = 0$.
For a given functional decomposition $G$
and behaviour mapping $\Phi$,
we denote the set of pointwise cooperative games as:
\begin{equation*}
	\text{PCG}(G,\Phi) :=
	\left\{v_G^{\Phi}(f,\mathbf{x}) \vert
	f \in \mathcal{F},
	\mathbf{x} \in \mathcal{X} \right\}
\end{equation*}
As an example of a pointwise cooperative game,
consider the function:
\begin{equation*}
	f(x_{1},x_{2},x_{3}) = x_{1} + x_{2} + x_{2}x_{3}
\end{equation*}
Assume $G$ is the anchored decomposition using the baseline $\mathbf{b} = (0,0,0)$
(see \Cref{sec:fdfi-cad-examples}),
$\Phi$ is the identity function,
and $\mathbf{x} = (3,4,5)$.
We can then compute the pointwise cooperative game $v := v_{G}^{\Phi}(f,\mathbf{x})$ as follows.
\begin{align*}
	v(S) & = \Phi \left( P_{\overline{S}}(f) \right)(\mathbf{x}) - \Phi(P_{[d]}(f))(\mathbf{x}) \\
	     & = P_{\overline{S}}(f)(\mathbf{x}) - f(0,0,0)                                                         \\
\end{align*}
Using this simplified formulation, we can easily compute the values
$v(S)$ for all subsets $S$:
\begin{align*}
	v(\emptyset) & = 0 & v(\{ 1,3 \})   & = 3  \\
	v(\{ 1 \})   & = 3 & v(\{ 2,3 \})   & = 24 \\
	v(\{ 2 \})   & = 4 & v(\{ 1,2 \})   & = 7  \\
	v(\{ 3 \})   & = 0 & v(\{ 1,2,3 \}) & = 27 \\
\end{align*}
The following proposition shows that if a functional decomposition $G$ is idempotent,
then the pointwise cooperative game $v_{G}^{\Phi}(g_{S}(f),\mathbf{x})$
corresponding to $G$ applied to one of its own components $g_{S}(f)$
is a weighted unanimity game
for all $S \subseteq [d], f \in \mathcal{F}, \mathbf{x} \in \mathcal{X}$.
This reinforces the idea that idempotence of $G$ is a type of ``purity''
in the interaction effects modeled by $G$,
as a unanimity game is the game-theoretic counterpart of a ``pure''
interaction between players.

\begin{restatable}{proposition}{idempotentimpliesunanimitygame}
	\label{fdfi:prop-idempotent-implies-unanimity-game}
	Assume $G$ is an idempotent additive decomposition on $\mathcal{F}$.
	Then for any $S \subseteq [d], f \in \mathcal{F}, \mathbf{x} \in \mathcal{X}$,
	$\Phi: \mathcal{F} \rightarrow \mathcal{F}$ we have:
	\begin{equation*}
		v_{G}^{\Phi}(g_{S}(f),\mathbf{x}) =  c v_{S}
	\end{equation*}
	where $c = \Phi(g_{S}(f))(\mathbf{x}) - \Phi(f_{0})(\mathbf{x})$,
	$\forall \mathbf{x} \in \mathcal{X}: f_{0}(\mathbf{x}) = 0$
	and $v_{S}$ is the unanimity game for the set $S$:
	\begin{equation*}
		\forall T \subseteq [d]: v_{S}(T) = \begin{cases}
			1 & \mbox{ if } S \subseteq T \\
			0 & \mbox{ otherwise.}
		\end{cases}
	\end{equation*}
\end{restatable}

Note that the pointwise cooperative game $v_{G}^{\Phi}$ is entirely defined
by its additive functional decomposition $G$ and the behaviour mapping $\Phi$.
Since any removal-based attribution method is also based on a behaviour mapping $\Phi$,
and its removal operators define an additive functional decomposition
as shown in \Cref{sec:fdfi-canonical-additive-decomposition},
we can introduce the following definition:

\begin{definition}[Corresponding Pointwise Cooperative Game]
	Given a removal-based attribution method $m$
	with behaviour mapping $\Phi$
	and corresponding additive decomposition $G$.
	The pointwise cooperative game $v_{G}^{\Phi}$
	is then called the \textbf{corresponding pointwise cooperative game}
	of $m$.
\end{definition}

We can now state the RBAM representation theorem,
which forms the fundamental connection between the class of
removal-based attribution methods
and cooperative game theory.

\begin{restatable}[RBAM Representation Theorem]{theorem}{rbamrepresentationtheorem}
	\label{thm:fdfi-rbam-representation-theorem}
	Assume $m: \mathcal{F} \times 2^{[d]} \rightarrow \mathcal{F}$ 
        is a removal-based attribution method
        with aggregation coefficients $\left\{ \alpha_{T}^{S} \mid S,T \subseteq [d] \right\}$,
        behaviour mapping $\Phi: \mathcal{F} \rightarrow \mathcal{F}$
        and corresponding functional decomposition $G$.
	Then for any $f \in \mathcal{F}, \mathbf{x} \in \mathcal{X}, S \subseteq [d]$:
	\begin{equation*}
                m(f,S)(\mathbf{x}) = \sum_{T \subseteq [d]}\alpha_{\overline{T}}^{S} \left[v_{G}^{\Phi}(f,\mathbf{x})(T) + \Phi(g_{\emptyset}(f))(\mathbf{x}) \right]
	\end{equation*}
	where $v_{G}^{\Phi}$ is the corresponding pointwise cooperative game
	of $m$.
\end{restatable}

The surprising consequence of the RBAM representation theorem
is that it shows that the entire class of removal-based attribution methods
can be viewed as values or interaction indices for a set of 
(possibly constant-shifted) cooperative games.
This implies that there is a fundamental link between each RBAM
and cooperative game theory,
including for a large number of methods that were not designed
with cooperative game theory in mind,
such as LIME \citep{ribeiro2016}, occlusion \citep{zeiler2014} or LOCO \citep{kohavi1997}.
Because the RBAM representation theorem shows that any removal-based attribution method
is a linear function of a possibly constant-shifted version
of its corresponding pointwise cooperative game,
we can view this linear function itself as a value or interaction index.
We formalize this notion in the following definition.

\begin{definition}[Pointwise Value/Interaction Index]
	Given a removal-based attribution method $m$ with behaviour mapping $\Phi$,
	corresponding functional decomposition $G$,
	and aggregation coefficients $\{ \alpha_{T}^{S} \mid S,T \subseteq [d] \}$.
	Assume $S \subseteq [d]$.
	The \textbf{pointwise value} (if $|S| = 1$) or \textbf{pointwise interaction index} (if $|S| > 1$)
	$\phi_{S}^{m}: \text{PCG}(G,\Phi) \rightarrow \mathbb{R}$ of $m$ is then defined as
	\begin{equation*}
		\phi_{S}^{m}(v) = \sum_{T \subseteq [d]} \alpha_{\overline{T}}^{S}v(T)
	\end{equation*}
\end{definition}

It is easy to see that any RBAM $m$ is equal to its pointwise value or interaction index
applied to a possibly constant-shifted version of
its corresponding pointwise cooperative game.

\subsection{Taxonomy of removal-based attribution methods}
\label{sec:fdfi-taxonomy}
Because any RBAM can be linked to a value
or interaction index for the set of pointwise cooperative games $\text{PCG}(G, \Phi)$,
we can define analogues of the game-theoretic axioms from \Cref{sec:game-theory-value-problem}.
We call these axioms the \textit{pointwise axioms} for removal-based attribution methods:

\begin{definition}
	Given a RBAM $m$ with corresponding functional decomposition $G$
	and behaviour mapping $\Phi$.
	For a given axiom for values and/or interaction indices $A$,
	we say that $m$ adheres to the \textbf{pointwise axiom $A$} if
	its pointwise value/interaction index $\phi_{S}^{m}$
	adheres to axiom $A$
	on $\text{PCG}(G,\Phi)$.
\end{definition}

For example,
$m$ adheres to the \textbf{Pointwise Dummy} axiom
if for any subset $S \subseteq [d]$,
the value or interaction index $\phi_{S}^{m}: \text{PCG}(G,\Phi) \rightarrow \mathbb{R}$
adheres to the Dummy axiom.
Analogously, we can define the Pointwise Linearity, (Interaction) Null,
(Interaction) Monotonicity,
(Interaction) Efficiency, 2-Efficiency, Anonymity, Symmetry and
Dummy Partnership axioms.
Because these axioms are entirely analogous to those for values and interaction indices
for cooperative games in general,
we can extend the taxonomies for values and interaction indices
from \Cref{sec:game-theory-value-problem,sec:game-theory-interaction-indices}
to removal-based attribution methods.
Specifically, we will call a given removal-based attribution method
a \textit{marginal contribution} (MC) attribution method if its pointwise value or interaction index
is an MC value or MC interaction index, respectively.
Analogously, we define a \textit{probabilistic, cardinal-probabilistic, random-order, Shapley, Banzhaf}
or \textit{Shapley-Taylor attribution method}
as an attribution method with a pointwise value or interaction index
that is of the corresponding type.

Note that this taxonomy also makes it immediately clear how the uniqueness of certain values
or interaction indices is lost when they are used to create attribution methods:
for any attribution method that adheres to a set of pointwise axioms,
the corresponding functional decomposition and behaviour mapping can be modified freely
without changing any of the pointwise axioms the method adheres to.

The following proposition shows that
a removal-based attribution method
is completely internally consistent,
\textit{i.e.}~both feature-level and interaction-level,
if and only if it is an MC attribution method
applied to its corresponding pointwise cooperative game.
This also implies that the constant shift from \Cref{thm:fdfi-rbam-representation-theorem}
vanishes for any internally consistent RBAM.

\begin{restatable}{proposition}{internalconsistencynecsuffconditions}
        \label{prop:fdfi-internal-consistency-nec-suff-conditions}
	\sloppy A removal-based attribution method $m$ with aggregation coefficients
        $\left\{ \alpha_{T}^{S} \mid S, T \subseteq [d] \right\}$,
	corresponding functional decomposition $G$
	and behaviour mapping $\Phi$
	is feature- and interaction-level consistent
	if and only if
	\begin{equation*}
		m(f,S)(\mathbf{x}) = \sum_{T \subseteq [d] \setminus S}\beta_{T}^{S} \Delta_{S}v_{G}^{\Phi}(f,\mathbf{x})(T)
	\end{equation*}
        with
        \begin{equation*}
                \forall S \subseteq [d], T \subseteq [d] \setminus S: \beta_{T}^{S} = \alpha_{\overline{T \cup S}}^{S}
        \end{equation*}
\end{restatable}

This proposition also implies that the three defining choices
of an internally consistent removal-based attribution method,
namely explained behaviour, feature removal and aggregation coefficients
(see \Cref{sec:removal-based-attribution}),
can equivalently be viewed as two defining choices:
a pointwise cooperative game and a set of aggregation coefficients.
Recall that the three defining choices for RBAMs correspond to the three choices of
behaviour, removal and aggregation in the XBR framework \citep{covert2021}, respectively.
Note that the choice of aggregation in XBR is defined more broadly,
which increases the scope of the resulting unifying framework.
For example, XBR includes aggregation methods that return subsets of features
that are deemed important,
rather than an importance score per feature.
As these methods still rely on the same concepts of behaviour mapping and feature removal,
many of the theoretical developments in this work surrounding additive decomposition
should translate to this setting.
We leave this as an interesting direction for further research.

\subsection{Efficient computation of simple attribution methods}
\label{sec:fdfi-efficient-computation}
In this section,
we will show that if $m$ is a simple MC attribution method,
\textit{i.e.}~the behaviour mapping is simply the identity mapping,
then $m(f,S)$ can be computed as a linear combination
of functional components $g_{T}(f)$
where $S \subseteq T$ and $\{ g_{S} \mid S \subseteq [d] \}$
is the corresponding additive decomposition of $m$.
This implies that any MC attribution method
for which the explained behaviour is the output of the model $f$
is a linear combination of functional components.
Although computing the coefficients of this linear combination
scales exponentially with the number of features,
we will proceed to show that
if we additionally assume that $m$ is a cardinal-probabilistic
attribution method,
then this collapses into a linear computation.
This idea was exploited in \citet{gevaert2022} and \citet{gevaert2023}
to develop the PDD-SHAP and A-PDD-SHAP algorithms,
which drastically reduce the amortized cost
of computing (asymmetric) Shapley values
by approximating them using a functional decomposition.
The fact that the same idea holds for the much more general
MC attribution methods shows the potential of this theoretical framework,
as it can easily be generalized to achieve similar performance gains
in other methods such as BANZHAF \citep{karczmarz2022},
causal Shapley values \citep{heskes2020},
and many others (see \Cref{sec:fdfi-rbam-examples}).

\begin{restatable}{proposition}{mcvaluecomputation}
	\label{prop:fdfi-mc-value-computation}
	Assume $m$ is a
	simple, internally consistent MC attribution method
	with corresponding functional decomposition $G$:
	\begin{equation*}
		m(f,S)(\mathbf{x})
		= \sum_{T \subseteq [d] \setminus S}\beta_{T}^{S} \Delta_S v_G^{I}(f,\mathbf{x})(T)
	\end{equation*}
	where $I$ is the identity mapping. Then:
	\begin{align*}
		m(f,S)
		                       & = \sum_{T \subseteq D \setminus S} \overline{\beta}_{T}^S g_{T \cup S}(f) \\
		\overline{\beta}_{T}^S & := \sum_{T \subseteq U \subseteq [d] \setminus S}\beta_U^S
	\end{align*}
\end{restatable}

Without making any further assumptions about the family of constants
$\{\beta_T^S|T \subseteq [d] \setminus S\}$, this expression cannot be simplified further.
If $m$ is univariate, the formula simplifies to:
\begin{align*}
	m(f,i)
	                     & = \sum_{S \subseteq [d] \setminus i} \overline{\beta}_S^i g_{S \cup i}(f) \\
	\overline{\beta}_S^i & = \sum_{S \subseteq T \subseteq [d] \setminus i}\beta_T^i
\end{align*}
The alternative form of $m(f,S)(\mathbf{x})$
given by \Cref{prop:fdfi-mc-value-computation} can be exploited
by approximating each component of the functional decomposition separately.
If such an approximation is available,
then the computation of any MC attribution method is reduced
to a simple weighted sum of the outputs of each component.
However, the total number of components grows exponentially with the number of features.
If we make the \textit{factor sparsity} assumption \citep[Appendix A]{owen2013},\citep{cox1984},
which roughly states that most of the behaviour of the model
can be explained using relatively low-order interactions,
then this problem can be mitigated by only approximating low-order terms.
This is exactly the idea behind PDD-SHAP \citep{gevaert2022}
and A-PDD-SHAP \citep{gevaert2023}.

As any probabilistic attribution method is also an MC attribution method,
it can be computed entirely analogously.
The fact that the coefficients $\{\beta_T^S \mid T \subseteq [d] \setminus S\}$
form a probability distribution does not allow any further simplification of the computation.
However, if $m$ is a cardinal-probabilistic attribution method, some simplification is possible.
As the coefficients only depend on the cardinalities $t$ and $s$ of the sets $T$ and $S$, respectively,
we can simplify the computation for $\overline{\beta}_T^S$:
\begin{align*}
	\overline{\beta}_T^S & = \sum_{T \subseteq U \subseteq [d] \setminus S} \beta_{|U|}^{|S|} \\
	                     & = \sum_{i=0}^{d-|S|-|T|} \binom{d-|S|-|T|}{i} \beta_{|T|+i}^{|S|}
\end{align*}
An important consequence of this simplification is that
the computation for $\overline{\beta}_T^S$ is now linear instead of exponential
in the total number of variables $d$.

If $m$ is a Shapley attribution method, then the computation for $\overline{\beta}_T^S$ can be simplified further to constant complexity.
As a Shapley attribution method is by definition univariate, we get the following:
\begin{align*}
	m(f,i)
	                       & = \sum_{S \subseteq [d] \setminus i} \overline{\beta}_{|S|} g_{S \cup i}(f) \\
	\overline{\beta}_{|S|} & = \sum_{T \subseteq [d] \setminus i} \beta_{|T|}
\end{align*}
with:
\begin{equation*}
	\beta_{|S|} := \frac{1}{d}\binom{d-1}{|S|}^{-1}
\end{equation*}
It can be shown that inserting this expression for $\beta_{|S|}$
into the expression for $\overline{\beta}_{|S|}$ leads to \citep{gevaert2022}:
\begin{align*}
	\overline{\beta}_{|S|} & = \frac{1}{|S|+1}                                                 \\
	m(f,i)                 & = \sum_{S \subseteq [d] \setminus i}\frac{g_{S \cup i}(f)}{|S|+1}
\end{align*}

\subsection{Sufficient Conditions for Functional Axioms}
\label{sec:fdfi-suff-cond-functional-axioms}
Using the unifying framework for removal-based attribution methods,
we will now derive sufficient conditions for the functional axioms introduced in \Cref{sec:removal-based-attribution}.
These conditions allow us to study the behaviour of explanation methods
based on the behaviour of their corresponding additive decompositions.
Proofs for all propositions in this section are given in \Cref{sec:proofs}.
We first consider the Functional Dummy axiom.
The following proposition shows that this functional axiom
is tightly linked with the property of additivity preservation in the functional decomposition.

\begin{restatable}{proposition}{suffcondfuncdummy}
	\label{prop:fdfi-suff-cond-func-dummy}
	Given a simple removal-based attribution method $m$
	with corresponding functional decomposition $G$.
	If $m$ adheres to the Pointwise Dummy axiom
	and $G$ is additivity-preserving,
	then $m$ adheres to the Functional Dummy axiom.
\end{restatable}

Analogously, the following proposition shows that the Functional Null axiom
is related to independence preservation in the functional decomposition.

\begin{restatable}{proposition}{pointwisenullipimpliesfuncnull}
	\label{prop:fdfi-pointwise-null-ip-implies-func-null}
	Given a removal-based attribution method $m$
	with corresponding functional decomposition $G$.
	If $m$ adheres to the Pointwise Null axiom
	and $G$ is independence-preserving,
	then $m$ adheres to the Functional Null axiom.
\end{restatable}

As any minimal decomposition is also independence- and additivity-preserving,
the previous two propositions allow us to link the Functional Null
and Dummy axioms to minimality:

\begin{corollary}
	Given a simple removal-based attribution method $m$
	with corresponding functional decomposition $G$.
	If $m$ is a probabilistic attribution method
	and $G$ is minimal,
	then $m$ adheres to the Functional Dummy axiom.
\end{corollary}

\begin{proof}
	This follows directly from \Cref{prop:fdfi-suff-cond-func-dummy,prop:fdfi-minimal-implies-additivity-preserving}
	and the fact that any probabilistic attribution method adheres to the Pointwise Dummy axiom.
\end{proof}

\begin{corollary}
	Given an MC attribution method $m$
	with corresponding functional decomposition $G$.
	If $G$ is minimal,
	then $m$ adheres to the Functional Null axiom.
\end{corollary}

\begin{proof}
	This follows directly from \Cref{prop:fdfi-minimal-implies-ip,prop:fdfi-pointwise-null-ip-implies-func-null},
	and the fact that any MC attribution method adheres to the Pointwise Null axiom.
\end{proof}

Finally, the following two propositions
link Functional Symmetry and Anonymity to the analogous properties
in the corresponding decomposition.

\begin{restatable}{proposition}{suffcondfuncsymmetry}
	Given a simple removal-based attribution method $m$
	with corresponding functional decomposition $G$.
	If $m$ adheres to the Pointwise Symmetry axiom
	and $G$ is symmetry-preserving,
	then $m$ adheres to the Functional Symmetry axiom.
\end{restatable}

\begin{restatable}{proposition}{suffcondfuncanonymity}
	Given a simple removal-based attribution method $m$
	with corresponding functional decomposition $G$.
	If $m$ adheres to the Pointwise Anonymity axiom
	and $G$ is anonymous,
	then $m$ adheres to the Functional Anonymity axiom.
\end{restatable}

\subsection{Examples}
\label{sec:fdfi-framework-examples}
In this section, we revisit some of the examples from \Cref{sec:fdfi-rbam-examples}
of existing removal-based attribution methods
using their corresponding functional decompositions
and the properties and functional axioms that they do or do not adhere to.

\subsubsection{Conditional and Marginal Shapley Values}
Consider the conditional Shapley value attribution method.
This method is defined as the Shapley value for the cooperative game
$v(S)$ defined using the conditional expectation of the model $f$:
\begin{equation*}
	v(S) = \mathbb{E}_{X}[f(X) \mid X_{S} = \mathbf{x}_{S}] - \mathbb{E}_{X}[f(X)]
\end{equation*}
where $\mathbf{x}$ is the input instance for which an explanation is computed.
Writing this as a pointwise cooperative game:
\begin{align*}
	v_{G}^{\Phi}(f,\mathbf{x})(S) & = \Phi \left( \sum_{T \subseteq S}g_{T}(f)\right)(\mathbf{x}) - \Phi(g_{\emptyset}(f))(\mathbf{x}) \\
	                              & = \Phi \left( P_{\overline{S}}(f) \right)(\mathbf{x}) - \Phi(g_{\emptyset}(f))(\mathbf{x})         \\
	                              & = \mathbb{E}_{X}[f(X) \mid X_{S} = \mathbf{x}_{S}] - \mathbb{E}_{X}[f(X)]
\end{align*}
we immediately see that the behaviour mapping $\Phi$ is the identity mapping,
and the removal operators $P_{S}$ are defined as:
\begin{equation*}
	P_{S}(f)(\mathbf{x}) = \mathbb{E}_{X}[f(X) \mid X_{\overline{S}} = \mathbf{x}_{\overline{S}}]
\end{equation*}
This implies that the corresponding decomposition of this method is the conditional decomposition:
\begin{align*}
	g_{S}(f) & = P_{\overline{S}}(f) - \sum_{T \subset S} g_{T}(f)                          \\
	         & = \mathbb{E}_{X}[f(X) \mid X_{S} = \mathbf{x}_{S}] - \sum_{T \subset S} g_{T}(f)
\end{align*}
The two choices of identity behaviour mapping and conditional decomposition
define the pointwise cooperative game of this method.
In turn, this pointwise cooperative game and the pointwise Shapley value
completely define the conditional Shapley value as a removal-based attribution method.
We know that the conditional decomposition is not minimal,
and therefore also not independence-preserving.
Indeed, as we have shown in practical examples before, this method does not adhere
to the Functional Null axiom: an independent variable of $f$ that is correlated with
a dependent variable of $f$ can receive non-zero attribution.
Analogously, we can show that this method does not adhere to the Functional Symmetry,
Functional Dummy or Functional Anonymity axioms.

Swapping out the conditional decomposition for the marginal decomposition,
we get a different set of properties.
First, we know that the marginal decomposition is minimal and therefore also
independence-preserving.
As the Shapley value adheres to the Null axiom,
we know that both the conditional and marginal Shapley values
adhere to the Pointwise Null axiom.
Therefore, the marginal Shapley values
also adhere to the Functional Null axiom,
\textit{i.e.}~the attribution value for an independent variable of $f$
is \textit{guaranteed to never be nonzero.}
As this decomposition is only anonymous if the marginal distributions
are identical,
we can only guarantee Functional Anonymity if this is the case.

\subsubsection{Shapley-Taylor Interaction Index}
The Shapley-Taylor interaction index \citep{sundararajan2020} is defined as a pointwise value only,
and can therefore be applied to any pointwise cooperative game.
Whereas the Shapley values are of order $1$,
the Shapley-Taylor interaction index is of arbitrary order $k$:
\begin{equation*}
	\phi^{STII}(v) = \begin{cases}
		\Delta_{S}v(\emptyset)                                                             & \text{ if } |S| < k \\
		\frac{k}{n} \sum_{T \subseteq [d] \setminus S} \binom{n-1}{|T|}^{-1}\Delta_{S}v(T) & \text{ if } |S| = k \\
	\end{cases}
\end{equation*}
Because the Shapley-Taylor interaction index can be applied to any pointwise cooperative game,
it can be applied to the same PCGs as the marginal or conditional Shapley values.
This allows us to extrapolate some of our findings about these methods to the Shapley-Taylor interaction index.
For example, if the marginal decomposition is chosen as a corresponding decomposition,
then we again have that this decomposition is minimal and the Shapley-Taylor interaction index
will also satisfy the Functional Null axiom,
whereas it will not if the conditional decomposition is used.
Analogously, Functional Anonymity of this method when using the marginal decomposition will only be
guaranteed if the marginal distributions of the variables are identical.

\subsubsection{Occlusion}
As a final example, we revisit the occlusion method.
Recall that this method removes features by replacing them with a fixed value,
and computes the attributions by removing each feature from the original instance separately.
The corresponding decomposition of this method is the anchored decomposition
for some baseline vector $\mathbf{c}$,
and the behaviour mapping is again the identity mapping.
These choices define the pointwise cooperative game of the occlusion method.
If the patch size is 1, the pointwise value of this method is:
\begin{equation*}
	\phi(v) = \Delta_{i}v([d] \setminus i) = v([d]) - v([d] \setminus i)
\end{equation*}
It is easy to verify that this is a cardinal-probabilistic value.
Therefore, this method satisfies the Pointwise Dummy, Null,
and Anonymity axioms.

The removal operators of the anchored decomposition are separable and linear,
implying that the decomposition is minimal as well.
This implies that the occlusion method adheres to the Functional Null axiom.
Furthermore, if the mean vector is a constant vector, \textit{i.e.}
$\mathbf{c} = (c, \dots, c)$ for some $c \in \mathbb{R}$,
then the anchored decomposition is also anonymous
and the occlusion method therefore adheres to the Functional Anonymity axiom.
Note that, as the choices of aggregation function and removal operators are
entirely independent, we could easily design new ``versions'' of occlusion,
where feature are removed by \textit{e.g.}~marginalizing over the input
distribution or the conditional distribution.
In such cases, we could again transfer the analysis we have already done
on these removal operators
to draw conclusions and provide guarantees about the behaviour
on these ``new'' methods.

\section{Conclusions \& Future Work}
\label{sec:conclusions-future-work}

In this work, we have introduced a formal unifying framework for univariate and multivariate removal-based attribution methods
based on additive functional decomposition.
We have also introduced the canonical additive decomposition (CAD), and have shown that
any valid additive decomposition can be expressed as a CAD.
This unifying framework enables us to formalize the behaviour of and differences between methods,
as well as their similarities.
We believe that such a formal framework is necessary to advance the research in this field beyond the development of different heuristics
and into a rigorous form of science.

We have already demonstrated in previous work how the ideas in this framework can be used to develop new, faster algorithms
for computing Shapley value-based attributions \citep{gevaert2022,gevaert2023}.
As we have now shown that the same core ideas apply to a much broader range of methods,
this opens up paths to further research into faster approximation algorithms for different explanations.

We have also shown how this theoretical framework can be used to study and explain surprising behaviour of methods
in the form of functional axioms.
The sufficient conditions that we derived for these properties allow researchers to taxonomize methods more effectively by their \textit{behaviour},
rather than their computational properties.
Such a behaviour-based taxonomy is much more useful for practitioners to decide which particular method is the most well-suited for a particular use case.

Another interesting avenue for further research is in the domain of objective evaluation of explanation methods.
As recent work has shown \citep{gevaert2024,tomsett2020}, current quality metrics for attribution methods measure different underlying properties of explanations,
although it is still unclear what these underlying properties are exactly.
This makes it difficult to rank methods by their quality for a given application when these metrics disagree.
The formal framework introduced in this work can be used to guide the search towards formal and exact descriptions of these underlying properties,
which can in turn help in the development of new metrics and benchmarking protocols.

Finally, a possible direction for further research is the extension of the framework to an even broader set of methods,
such as counterfactual explanations
or explanations that return a subset of important features instead of attribution scores.
As these methods often also rely on a notion of the removal of features,
there is reason to believe that this type of methods can also be incorporated into a similar framework.
Such a development would allow us to formally study the differences and similarities between attribution and counterfactual methods.
Also, the study of attribution methods that do not fit in the current framework, such as the integrated gradients method,
can be interesting to provide new insights into the fundamental differences between these methods and removal-based attributions.

To conclude, we believe that formal, unifying frameworks are necessary tools to transform the field of explainable machine learning
into a truly rigorous science.
With this work, we develop a first step towards this formalization, and provide a basis for further advancement in this direction.


\newpage

\appendix
\section{Proofs of theorems and propositions}
\label{sec:proofs}

\proprbamsufficientconditions*

\begin{proof}
	We need to prove that for any $S \subseteq [d]$,
	the vector function $m_{S}$ is a linear transformation
	from (a subspace of) $\mathbb{R}^{2^{d}}$ to $\mathbb{R}$.
	Then $m_{S}$ can be written as a vector product,
	and the result follows.
	For a given point $\mathbf{x} \in \mathcal{X}$, define the set $\Lambda_\mathbf{x}$ as follows:
	\begin{equation*}
		\Lambda_\mathbf{x} := \{(P_{T}(f)(\mathbf{x}) \mid T \subseteq [d]) \mid f \in \mathcal{F}\}
	\end{equation*}
	Then the domain of $m_{S}$ is $\bigcup_{\mathbf{x} \in \mathcal{X}} \Lambda_\mathbf{x}$.
	Note that this is not necessarily a linear subspace of $\mathbb{R}^{2^{d}}$,
	nor does this subset necessarily span $\mathbb{R}^{2^{d}}$.
	Therefore, we will show that there exists a unique linear extension
	of $m_{S}$ to $\text{span}(\bigcup_{\mathbf{x} \in \mathcal{X}} \Lambda_\mathbf{x})$.

	Because the removal operators are linear,
	we know that
	each $\Lambda_\mathbf{x}$ is a linear subspace of $\mathbb{R}^{2^d}$.
	From the linearity of $m(\cdot, S)$, we also know that $m_{S}$ is linear on each of the subspaces $\Lambda_\mathbf{x}$.
	Any vector $v$ in $\text{span}(\bigcup_{\mathbf{x} \in \mathcal{X}} \Lambda_\mathbf{x})$
	can be written as a sum of vectors
	$v_\mathbf{x}$ where each $v_\mathbf{x} \in \Lambda_{\mathbf{x}}$:
	\begin{equation*}
		v = \sum_{\mathbf{x} \in \mathcal{X}} v_\mathbf{x}
	\end{equation*}
	Therefore, we can linearly extend $m_{S}$
	to $\text{span}(\bigcup_{\mathbf{x} \in \mathcal{X}} \Lambda_\mathbf{x})$ as follows:
	\begin{align*}
		m_{S}(v) & = m_{S}\left(\sum_{\mathbf{x} \in \mathcal{X}} v_\mathbf{x}\right) \\
		         & = \sum_{\mathbf{x} \in \mathcal{X}}m_{S}( v_\mathbf{x})            \\
	\end{align*}
	It is easy to verify that this linear extension is unique
	(\textit{i.e.}~if $v$ can be written as a sum of vectors $v_x$ in multiple ways,
	then summing the corresponding values of $m_{S}$ will give the same result for $v$).
	Therefore, $m_{S}$ is a linear transformation from a linear subspace of $\mathbb{R}^{2^d}$ to $\mathbb{R}$.
\end{proof}

\begin{lemma}
	\label{lemma:fdfi-ceiling-dependency-structure}
	Given a collection of sets $\mathcal{S}$. We define the \textit{ceiling} of $\mathcal{S}$ as follows:
	\begin{equation*}
		\lceil \mathcal{S} \rceil := \{S \in \mathcal{S}|\nexists S' \in \mathcal{S}: S \subset S'\}
	\end{equation*}
	\textit{i.e.}~the ceiling of $\mathcal{S}$
	consists of all the maximal elements of $\mathcal{S}$
	with respect to the partial order $\subseteq$.
	Assume $f \in \mathcal{F}$,
	$\mathcal{S} \subseteq 2^{[d]}$,
	$\{g_S(f)|S \in \mathcal{S}\}$ is an additive decomposition of $f$:
	\begin{equation*}
		f = \sum_{S \in \mathcal{S}} g_S(f)
	\end{equation*}
	Then $\lceil \mathcal{S} \rceil \in \text{DS}(f)$.
\end{lemma}
\begin{proof}
	From the assumption, we can write $f = \sum_{S \in \mathcal{S}} g_S(f)$ for some set of functions $g_S(f)$.
	We can then define a new decomposition as follows:
	\begin{equation*}
		g'_S(f) =
		\left\{
		\begin{array}{ll}
			\sum_{T \in \mathcal{S}, T \subseteq S} \frac{g_T(f)}{\alpha_T} & \mbox{if } S \in \lceil \mathcal{S} \rceil \\
			0                                                               & \mbox{otherwise }
		\end{array}
		\right.
	\end{equation*}
	where $\alpha_T := |\{U \in \lceil \mathcal{S} \rceil: T \subseteq U\}|$.
	It is easy to see that this decomposition indeed adds up to $f$ and only the components in $\lceil \mathcal{S} \rceil$ are used.
	Also, it follows directly from the definition of $\lceil \mathcal{S} \rceil$
	that $\forall S, T \in \lceil \mathcal{S} \rceil: S \subseteq T \implies S = T$.
	Therefore, $\lceil \mathcal{S} \rceil \in \text{DS}(f)$.
\end{proof}

\propmdsunique*

\begin{proof}
	It is easy to see that $\leq_\text{DS}$ defines a partial order on the elements of $\text{DS}(f)$.
	Also, $\forall f \in \mathcal{F}: \text{DS}(f) \neq \emptyset$,
	as the trivial dependency structure $\mathcal{S} = \{[d]\} \in \text{DS}(f), \forall f \in \mathcal{F}$.
	Because $(\text{DS}(f), \leq_\text{DS})$ is a non-empty finite poset,
	it must contain at least one minimal element.
	We will now prove that this element is indeed unique.

	Assume $\mathcal{S}, \mathcal{U} \in \text{DS}(f)$
	are both minimal and $\mathcal{S} \neq \mathcal{U}$.
	Since $\mathcal{S}, \mathcal{U} \in \text{DS}(f)$, we have
	\begin{equation*}
		f = \sum_{S \in \mathcal{S}}g_S(f) = \sum_{U \in \mathcal{U}}g'_U(f)
	\end{equation*}
	for some decompositions $\{g_S|S \in \mathcal{S}\}$ and $\{g'_U|U \in \mathcal{U}\}$.
	We also have $\exists U \in \mathcal{U}: \forall S \in \mathcal{S}: U \nsubseteq S$.
	Otherwise $\mathcal{U} \leq_{\text{DS}} \mathcal{S}$, and $\mathcal{S}$ would not be minimal.
	Now choose $U \in \mathcal{U}$ such that $\forall S \in \mathcal{S}: U \nsubseteq S$.
	We can write:
	\begin{align*}
		g'_U(f) & = f - \sum_{U' \in \mathcal{U}'}g'_{U'}(f)                              \\
		        & = \sum_{S \in \mathcal{S}}g_S(f) - \sum_{U' \in \mathcal{U}'}g'_{U'}(f)
	\end{align*}
	where $\mathcal{U}' := \mathcal{U} \setminus U$.
	From \Cref{lemma:fdfi-ceiling-dependency-structure},
	we therefore have $\lceil \mathcal{S} \cup \mathcal{U}' \rceil \in \text{DS}(g'_U(f))$.
	Denote $\mathcal{V} := \lceil \mathcal{S} \cup \mathcal{U}' \rceil$.
	Then there exists a valid decomposition $\{h_V: V \in \mathcal{V}\}$ of $g'_U(f)$:
	\begin{equation*}
		g'_U(f) = \sum_{V \in \mathcal{V}} h_V(g'_U(f))
	\end{equation*}
	From our assumption about $U$,
	we know that $(\forall S \in \mathcal{S})(U \nsubseteq S)$.
	Because $\mathcal{U} \in \text{DS}(f)$,
	we also have $(\forall U' \in \mathcal{U}')(U \nsubseteq U')$.
	Therefore, $(\forall V \in \mathcal{V})(U \nsubseteq V)$.

	We also know that $g'_U(f) = \sum_{V \in \mathcal{V}} h_V(g'_U(f))$ is independent of $X_{\overline{U}}$.
	Therefore, if we choose an arbitrary value for $\mathbf{x}_{\overline{U}}$, we get
	\begin{equation*}
		\sum_{V \in \mathcal{V}} h_V(g'_U(f))(\mathbf{x}_{\overline{U}}, \cdot) = g'_{U}(f)
	\end{equation*}
	where we use the notation $f(\mathbf{x}_S, \cdot)$ to signify partial application:
	\begin{equation*}
		f(\mathbf{x}_S, \cdot):
		\mathbb{R}^{|\overline{S}|} \rightarrow \mathbb{R}:
		\mathbf{x}'_{\overline{S}} \mapsto f\left(\mathbf{x}_S, \mathbf{x}'_{\overline{S}}\right)
	\end{equation*}
	By definition, $\forall V \in \mathcal{V}: h_V(g'_U(f))(\mathbf{x}_{\overline{U}}, \cdot)$
	is independent of $X_{\overline{U} \cup \overline{V}} = X_{\overline{U \cap V}}$.
	This implies that $\mathcal{V}' := \lceil \{U \cap V|V \in \mathcal{V}\} \rceil \in \text{DS}(g'_U(f))$.
	By definition, $(\forall V' \in \mathcal{V}')(V' \subseteq U)$.
	Also we know that $(\forall V \in \mathcal{V})(U \nsubseteq V)$.
	Therefore, $(\forall V' \in \mathcal{V}')(V' \subset U)$.
	This in turn means that $g'_U(f)$ can be decomposed into a set of functions that only depend on strict subsets of $U$.
	However, this contradicts the assumption that $\mathcal{U}$ was minimal:
	construct $\mathcal{U}' := \lceil (\mathcal{U} \setminus U) \cup \mathcal{V}'\rceil$.
	Then $\mathcal{U}' \in \text{DS}(f)$ and $\mathcal{U}' \leq_{\text{DS}} \mathcal{U}$.
	From this contradiction, we conclude that $\mathcal{S} = \mathcal{U}$.
\end{proof}

\propmdsindependent*

\begin{proof}
	We start by proving that if $f$ depends on $X_i$,
	then $i \in \bigcup \mathcal{S}$.
	Assume that $f$ depends on $X_i$ and $i \notin \bigcup \mathcal{S}$.
	Then $f = \sum_{S \in \mathcal{S}}g_S(f)$.
	Since $i \notin \bigcup \mathcal{S}$, we have $\forall S \in \mathcal{S}: i \notin S$.
	But then $\forall S \in \mathcal{S}: g_S(f)$ is independent of $X_{i}$,
	and therefore $f$ is independent of $X_i$,
	which is a contradiction.

	We now prove the other direction.
	Assume $f \in \mathcal{F}$,
	$\text{MDS}(f) = \mathcal{S}$
	and $X_{i}$ is an independent variable of $f$.
	We must then prove that $i \notin \bigcup \mathcal{S}$.
	As $f$ does not depend on $X_{i}$,
	the following decomposition of $f$ is valid:
	\begin{equation*}
		g_{S}(f) = \begin{cases}
			f & \mbox{ if } S = [d] \setminus i \\
			0 & \mbox{ otherwise.}
		\end{cases}
	\end{equation*}
	Therefore, $\{ [d] \setminus i \}$ is a dependency structure of $f$.
	From the minimality of $\mathcal{S}$,
	we know that $\mathcal{S}$ must be a sub-dependency structure
	of $\{ [d] \setminus i \}$.
	Therefore, $\forall S \in \mathcal{S}: S \subseteq [d] \setminus i$,
	and $i \notin \bigcup \mathcal{S}$.

\end{proof}

\thmcadexistenceuniqueness*

\begin{proof}
	We begin by proving that any CAD is a valid additive decomposition.
	Completeness follows immediately from the summation property of the CAD:
	\begin{equation*}
		f = P_\emptyset(f) = \sum_{S \subseteq \overline{\emptyset}} g_{S}(f) = \sum_{S \subseteq [d]}g_{S}(f)
	\end{equation*}
	We prove the independence property using induction.
	Consider $g_{\emptyset}(f) = P_{[d]}(f)$.
	From the definition of $P_S$, we know that $P_{[d]}(f)$ is independent of $X_{[d]} = X_{\overline{\emptyset}}$.
	Now assume the independence property holds for any strict subset $T \subset S$.
	Then we have $g_{S}(f) = P_{\overline{S}}(f) - \sum_{T \subset S}g_{T}(f)$.
	Again from the definition of $P_S$ we know that $P_{\overline{S}}(f)$ is independent of $X_{\overline{S}}$.
	From the induction assumption we know $g_{T}(f)$ is independent of $X_{\overline{T}}, \forall T \subset S$.
	Since $T \subset S$, we have $\overline{S} \subset \overline{T}$,
	and therefore $g_{T}(f)$ is independent of $X_{\overline{S}},
		\forall T \subset S$.
	$g_{S}(f)$ is therefore a sum of functions which are all individually independent of $X_{\overline{S}}$.
	Therefore, $g_{S}(f)$ is independent of $X_{\overline{S}}$.

	We now prove the reverse implication.
	Assume $G: \{g_S:S \subseteq [d]\}$ is an additive functional decomposition on $\mathcal{F}$.
	We then have:
	\begin{itemize}
		\item $f = \sum_{S \subseteq [d]} g_S(f)$
		\item $\forall S \subseteq [d]: g_S(f) $ is independent of $X_{\overline{S}}$.
	\end{itemize}
	Then we need to show that there exists a set of operators $P = \{P_S: S \subseteq [d]\}$
	such that $\forall f \in \mathcal{F}, S \subseteq [d]$:
	\begin{enumerate}
		\item $P_\emptyset = I$
		\item $P_S(f)$ is independent of $X_{S}$
		\item $g_S(f) = P_{\overline{S}}(f) - \sum_{T \subset S}f_T$
	\end{enumerate}

	Consider $P_S(f) = \sum_{T \subseteq \overline{S}} g_T(f)$.

	\begin{enumerate}
		\item $P_\emptyset(f) = \sum_{T \subseteq [d]}g_T(f)$.
		      Because the decomposition is valid, we therefore have $P_\emptyset(f) = f \implies P_\emptyset = I$.
		\item $P_S(f) = \sum_{T \subseteq \overline{S}}g_T(f)$.
		      Because the decomposition is valid, we know that $g_T(f)$ is independent of $X_{\overline{T}}, \forall T \subseteq [d]$.
		      We also know that $T \subseteq \overline{S} \implies S \subseteq \overline{T}$.
		      Therefore $g_T(f)$ is independent of $X_{S}, \forall T \subseteq \overline{S}$.
		      $P_S(f)$ is then a sum of functions which are all individually independent of $X_{S}$.
		      Therefore, $P_S(f)$ is independent of $X_{S}$.
		\item From the definition of $P_S$, we get:
		      \begin{align*}
			      \sum_{T \subseteq S}g_T(f) & = P_{\overline{S}}(f)                            \\
			      g_S(f)                     & = P_{\overline{S}}(f) - \sum_{T \subset S}g_T(f)
		      \end{align*}
	\end{enumerate}
\end{proof}

\lemmaminimalcadmds*

\begin{proof}
	Choose $S \subseteq [d]$ such that $\nexists T \in \text{MDS}(f): S \subseteq T$.
	From the definition of a dependency structure,
	there exists some additive functional decomposition $H \in \mathcal{D}_{\mathcal{F}}$
	such that $\forall T \subseteq [d]: T \notin \text{MDS}(f) \implies h_T(f) = 0$.
	Therefore, $h_S(f) = 0$ and $\forall R \supseteq S: h_R(f) = 0$.
	However, this means that all super-$S$ terms in $H$ are zero.
	From the definition of minimality,
	this implies that all super-$S$ terms in $G$ are also zero.
\end{proof}

\propminimalcadmds*

\begin{proof}
	Assume $\exists S \in \mathcal{S}: g_S(f) = 0$.
	Define $\mathcal{S}' := \{S \subseteq [d]: g_S(f) \neq 0\}$.
	Then from Lemma \ref{lemma:fdfi-ceiling-dependency-structure}
	we have $\lceil \mathcal{S}' \rceil \in \text{DS}(f)$.
	From Lemma \ref{lemma:fdfi-minimal-cad-mds},
	we have $\forall S \in \mathcal{S}': \exists T \in \mathcal{S}: S \subseteq T$,
	which implies that $\lceil \mathcal{S}' \rceil$ is a sub-dependency structure of $\mathcal{S}$.
	However, $S \notin \lceil \mathcal{S}' \rceil$.
	Therefore, $\lceil \mathcal{S}' \rceil \leq_{\text{DS}} \mathcal{S}$,
	which contradicts the assumption $\mathcal{S} = \text{MDS}(f)$.
\end{proof}

\propnecsuffindependencepreservation*

\begin{proof}
	Assume first $G$ is independence-preserving.
	Then, for any $S \subseteq [d] \setminus i$:
	\begin{align*}
		P_{S}(f) & = \sum_{T \subseteq \overline{S}} g_{T}(f)                               \\
		         & = \sum_{T \subseteq \overline{S} \setminus i} g_{T}(f) + g_{T \cup i}(f) \\
		         & = \sum_{T \subseteq \overline{S \cup i}} g_{T}(f) + g_{T \cup i}(f)      \\
		         & = \sum_{T \subseteq \overline{S \cup i}} g_{T}(f)                        \\
		         & = P_{S \cup i}(f)
	\end{align*}
	Now assume $P_{S}(f) = P_{S \cup i}(f)$, for any $S \subseteq [d] \setminus i$.
	Then:
	\begin{align*}
		g_{i}(f) & = P_{\overline{i}}(f) - g_{\emptyset}(f) \\
		         & = P_{[d]}(f) - g_{\emptyset}(f)          \\
		         & = g_{\emptyset}(f) - g_{\emptyset}(f)    \\
		         & = 0
	\end{align*}
	Now assume $S \subseteq [d] \setminus i$
	and $g_{T \cup i}(f) = 0$,
	for any strict subset $T \subseteq S$.
	We then have:
	\begin{align*}
		g_{S \cup i}(f) & = P_{\overline{S \cup i}}(f) - \sum_{T \subseteq S \cup i}g_{T}(f)     \\
		                & = P_{\overline{S}}(f) - \sum_{T \subseteq S}g_{T}(f) + g_{T \cup i}(f) \\
		                & = P_{\overline{S}}(f) - \sum_{T \subseteq S}g_{T}(f)                   \\
		                & = P_{\overline{S}}(f) - P_{\overline{S}}(f) = 0                        \\
	\end{align*}
	The required property follows from induction.
\end{proof}

\minimalimpliesip*

\begin{proof}
	Assume $f \in \mathcal{F}$, $X_{i}$ is an independent variable of $f$.
	From \Cref{thm:fdfi-cad-existence-uniqueness},
	we know that $f$ has a unique minimal dependency structure $\mathcal{S} = \text{MDS}(f)$.
	From \Cref{prop:fdfi-minimal-cad-mds},
	we know that $\forall S \subseteq [d]: (\nexists T \in \mathcal{S}: S \subseteq T) \implies g_S(f) = 0$.
	From \Cref{prop:fdfi-prop-mds-independent},
	we know that $f$ is independent to $X_i \implies i \notin \bigcup \mathcal{S}$.
	Therefore, $\forall {S} \in \mathcal{S}: i \notin S$.
	This implies that $\forall S \subseteq [d]: i \in S \implies (\nexists T \in \mathcal{S}: S \subseteq T)$,
	which implies $g_S(f) = 0$.
\end{proof}

\linearipimpliesminimal*

\begin{proof}
	Assume $G$ is independence-preserving, $f \in \mathcal{F}$.
	Suppose that $\exists H = \{h_S: S \subseteq [d]\}$,
	such that $H$ is valid and $H(f)$ has all super-$Z$ components equal to zero for some $Z \subseteq [d]$.
	We then need to prove that all super-$Z$ components of $G(f)$ are also equal to zero.
	Assume $S \subseteq [d]$ with $Z \subseteq S$.
	We need to prove that $g_S(f) = 0$.
	\begin{align*}
		g_S(f) & = \sum_{T \subseteq S}(-1)^{|S|-|T|}P_{\overline{T}}(f)                                                                   \\
		       & = \sum_{T \subseteq S}(-1)^{|S|-|T|}P_{\overline{T}}\left(\sum_{U \subseteq [d]}h_U(f)\right) & \text{(Validity of $H$)}  \\
		       & = \sum_{U \subseteq [d]}\sum_{T \subseteq S}(-1)^{|S|-|T|}P_{\overline{T}}(h_U(f))            & \text{(Linearity of $G$)}
	\end{align*}
	For any term in the outer sum, we have one of two possibilities:
	\begin{enumerate}
		\item $S \subseteq U$
		\item $S \not \subseteq U$
	\end{enumerate}
	In the first case, we have:
	$$
		S \subseteq U \implies Z \subseteq U \implies h_U(f) = 0 \implies P_{\overline{T}}(h_U(f)) = 0, \forall T \subseteq [d]
	$$
	Because every term in the inner sum is 0, the term in the outer sum also becomes 0.
	In the second case, we have $S \not \subseteq U \implies \exists k \in S: k \notin U$. In this case, we can divide all the subsets of $T \subseteq S$ in 2 groups: $T_1 := \{T \subseteq S \setminus \{k\}\}$ and $T_2 := \{T \cup \{k\}: T \in T_1\}$. Obviously, we have $|T_1| = |T_2|$, $T_1 \cup T_2 = 2^S$, and $T_1 \cap T_2 = \emptyset$. In other words, $T_1$ and $T_2$ form a partition of $2^S$.
	Choose an arbitrary $V \in T_1$ and corresponding $V' := V \cup \{k\}$, so that $V' \in T_2$. Because $k \notin U$, we have $h_U(f)$ is independent of $X_k$. Because of independence preservation and because $k \notin V \implies k \in \overline{V}$, we therefore get $P_{\overline{V}}(h_U(f)) = P_{\overline{V} \setminus \{k\}}(h_U(f)) = P_{\overline{V'}}(h_U(f))$. Also, because $|V'| = |V| + 1$, the corresponding coefficient $(-1)^{|S|-|V'|}$ is equal to $-(-1)^{|S|-|V|}$. Therefore, all terms in the inner sum cancel out and the term in the outer sum again becomes 0.
\end{proof}

\idempotencenecsuffconditions*

\begin{proof}
	Assume $G$ is a CAD with removal operators $P_{S}$ that adhere to the given assumption.
	We then need to prove that $G$ is idempotent:
	\begin{equation*}
		\forall f \in \mathcal{F}, S, T \subseteq [d]: g_T(g_S(f)) =
		\left\{
		\begin{array}{ll}
			g_S(f) & \mbox{if } T = S  \\
			0      & \mbox{otherwise.}
		\end{array}
		\right.
	\end{equation*}
	Assume first $S = T$.
	We need to prove that $g_S(g_S(f)) = g_S(f)$.
	From the definition of $G$, we have the following:
	\begin{equation*}
		g_S(g_S(f)) = \sum_{T \subseteq S}(-1)^{|T|-|S|}P_{\overline{T}}(g_S(f))
	\end{equation*}
	From the definition of strict subsets, we also have:
	\begin{equation*}
		\forall T \subset S: \exists i \in S: i \in \overline{T}
	\end{equation*}
	Therefore, the second case of the assumption on $P_{T}$ implies:
	\begin{equation*}
		\forall T \subset S: P_{\overline{T}}(g_S(f)) = 0
	\end{equation*}
	and the sum collapses to a single term:
	\begin{equation*}
		g_S(g_S(f)) = P_{\overline{S}}(g_S(f))
	\end{equation*}
	By definition, $S \cap \overline{S} = \emptyset$.
	The first case of the assumption on $P_{T}$ therefore implies that $P_{\overline{S}}(g_S(f)) = g_S(f)$.
	Now assume $S \neq T$.
	We need to prove that $g_T(g_S(f)) = 0$.
	From the definition of $G$:
	\begin{equation*}
		g_T(g_S(f)) = P_{\overline{T}}(g_S(f)) - \sum_{R \subset T}g_R(g_S(f))
	\end{equation*}
	We will prove that this equals zero using induction on the cardinality of $T$.
	Assume first $T = \emptyset$.
	Then $g_T(g_S(f)) = P_{[d]}(g_S(f))$.
	Since $S \subseteq [d]$, the assumption implies that $g_T(g_S(f)) = 0$.
	Assume now that the idempotence property holds for all $T'$ with $|T'| < |T|$:
	\begin{equation*}
		\forall f \in \mathcal{F}, S, T' \subseteq [d], |T'| < |T| \implies g_{T'}(g_S(f)) =
		\left\{
		\begin{array}{ll}
			g_S(f) & \mbox{if } T' = S \\
			0      & \mbox{otherwise.}
		\end{array}
		\right.
	\end{equation*}
	We now have two possible cases in the definition of $g_T(g_S(f))$:
	\begin{itemize}
		\item $S \subset T$.
		      In this case, the induction hypothesis implies $\sum_{R \subset T}g_R(g_S(f)) = g_S(f)$.
		      Since $S \subset T$,
		      we also have $S \cap \overline{T} = \emptyset \implies P_{\overline{T}}(g_S(f)) = g_S(f)$.
		      Therefore, $g_T(g_S(f)) = g_S(f) - g_S(f) = 0$.
		\item $S \not \subset T \implies S \cap \overline{T} \neq \emptyset \implies P_{\overline{T}}(g_S(f)) = 0$.
		      In this case, the induction hypothesis implies $\sum_{R \subset T}g_R(g_S(f)) = 0$.
		      Therefore, $g_T(g_S(f)) = 0 - 0 = 0$.
	\end{itemize}
	Now assume $G$ is idempotent.
	We must prove that:
	\begin{equation*}
		P_{T}(g_{S}(f)) = \begin{cases}
			g_{S}(f) & \mbox{ if } S \cap T = \emptyset    \\
			0        & \mbox{ if } S \cap T \neq \emptyset
		\end{cases}
	\end{equation*}
	Assume first $S \cap T = \emptyset \iff S \subseteq \overline{T}$.
	There are 2 options: either $S = \overline{T}$ or $S \subset \overline{T}$.
	If $S = \overline{T}$, then idempotence implies that $g_{\overline{T}}(g_{S}(f)) = g_{S}(f)$.
	We also have:
	\begin{align*}
		g_{\overline{T}}(g_S(f)) & = P_{T}(g_S(f)) - \sum_{R \subset S}g_R(g_S(f)) \\
		                         & = P_{T}(g_S(f))                                 \\
	\end{align*}
	Therefore, $P_{T}(g_{S}(f)) = g_{S}(f)$.
	Now assume $S \subset \overline{T}$.
	Then idempotence implies that $g_{\overline{T}}(g_{S}(f)) = 0$.
	We also have:
	\begin{align*}
		g_{\overline{T}}(g_S(f)) & = P_T(g_S(f)) - \sum_{R \subset \overline{T}}g_R(g_S(f)) \\
		                         & = P_T(g_S(f)) - g_S(f)                                   \\
	\end{align*}
	Therefore, $P_T(g_S(f)) = g_S(f)$.
	Now assume $S \cap T \neq \emptyset \iff S \nsubseteq \overline{T}$.
	We must prove that $P_T(g_S(f)) = 0$.
	From idempotence, we know that $g_{\overline{T}}(g_{S}(f)) = 0$.
	We also have:
	\begin{align*}
		g_{\overline{T}}(g_S(f)) & = P_T(g_S(f)) - \sum_{R \subset \overline{T}}g_R(g_S(f)) \\
		                         & = P_T(g_S(f))                                            \\
	\end{align*}
	Therefore, $P_T(g_S(f)) = 0$.
\end{proof}

\linearseparableiffidempotent*

\begin{proof}
	Assume $G$ is linear with separable removal operators $\{ P_{S} \mid S \subseteq d] \}$.
	We will prove that:
	\begin{equation*}
		P_{T}(g_{S}(f)) = \begin{cases}
			0        & \mbox{ if } T \cap S \neq \emptyset \\
			g_{S}(f) & \mbox{ if } T \cap S = \emptyset    \\
		\end{cases}
	\end{equation*}
	for any function $f$ and subset $T \subseteq [d]$,
	which is equivalent to idempotence (\Cref{prop:fdfi-idempotence-nec-suff-conditions}).
	First assume $S = \emptyset$ (which implies $T \cap S = \emptyset$).
	Then:
	\begin{align*}
		P_{T}(g_{S}(f)) & = P_{T}(P_{[d]}(f))             \\
		                & = P_{[d]}(f) = g_{\emptyset}(f)
	\end{align*}
	since $T \subseteq [d]$ and $P_{T} \circ P_{[d]} = P_{T \cup [d]} = P_{[d]}$.
	Now assume that $T \cap S' = \emptyset \implies P_{T}(g_{S'}(f)) = g_{S'}(f)$
	for all strict subsets $S' \subset S$ and $S \cap T = \emptyset$.
	Then:
	\begin{align*}
		P_{T}(g_{S}(f)) & = P_{T}(P_{\overline{S}}(f)) - \sum_{S' \subset S}P_{T}(g_{S'}(f)) \\
		                & = P_{T}(P_{\overline{S}}(f)) - \sum_{S' \subset S}g_{S'}(f)        \\
		                & = P_{\overline{S}}(f) - \sum_{S' \subset S}g_{S'}(f)               \\
		                & = g_{S}(f)
	\end{align*}
	since $S \cap T = \emptyset \implies T \subseteq \overline{S} \implies P_{T} \circ P_{\overline{S}} = P_{\overline{S}}$.
	The result then follows from induction.
	We will now show the case $T \cap S \neq \emptyset \implies P_{T}(g_{S}(f) = 0$.
	First assume $|S| = 1$.
	Then $T = S \implies P_{T} \circ P_{\overline{S}} = P_{[d]}$.
	Therefore:
	\begin{align*}
		P_{T}(g_{S}(f)) & = P_{T}(P_{\overline{S}}(f)) - \sum_{S' \subset S} P_{T}(g_{S'}(f)) \\
		                & = P_{[d]}(f) - P_{T}(g_{\emptyset}(f))                              \\
		                & = P_{[d]}(f) - P_{[d]}(f) = 0
	\end{align*}
	Now assume that the required property holds for all strict subsets $S' \subset S$
	and $S \cap T \neq \emptyset$.
	Then we have:
	\begin{align*}
		\sum_{S' \subset S}P_{T}(g_{S'}(f)) & = \sum_{S' \subseteq S \setminus T} P_{T}(g_{S'}(f))           \\
		                                    & = \sum_{S' \subseteq S \setminus T}g_{S'}(f)                   \\
		                                    & = P_{\overline{S \setminus T}}(f) = P_{T \cup \overline{S}}(f)
	\end{align*}
	Therefore:
	\begin{align*}
		P_{T}(g_{S}(f)) & = P_{T}(P_{\overline{S}}(f)) - P_{T \cup \overline{S}}(f) \\
		                & = 0
	\end{align*}
	This proves that linearity and separability implies idempotence.
	We will now proceed to prove the reverse implication.
	Assume $G$ is idempotent and linear.
	We must prove that for any $S, T \subseteq [d]: P_{S} \circ P_{T} = P_{S \cup T}$.
	For any $f \in \mathcal{F}$:
	\begin{align*}
		P_{S}(P_{T}(f)) & = \sum_{U \subseteq \overline{S}} \sum_{V \subseteq \overline{T}} g_{U}(g_{V}(f)) \\
		                & = \sum_{U \subseteq \overline{S} \cap \overline{T}}g_{U(f)}                       \\
		                & = P_{\overline{\overline{S} \cap \overline{T}}}(f) = P_{S \cup T}(f)
	\end{align*}
\end{proof}

\additivitypreservingremovalconditions*

\begin{proof}
	First assume $G$ is additivity-preserving,
	and $f$ is a function with additive variable $X_{i}$.
	Then for any subset $S \subseteq [d] \setminus i$:
	\begin{align*}
		P_{S \cup i}(f) - P_{S}(f) & = \sum_{T \subseteq  \overline{S \cup i}}g_{S}(f) - \sum_{T \subseteq \overline{S}}g_{T}(f)     \\
		                           & = \sum_{T \subseteq \overline{S} \setminus i}g_{T}(f) - \sum_{T \subseteq \overline{S}}g_{T}(f) \\
		                           & = -\sum_{T \subseteq \overline{S} \setminus i}g_{T \cup i}(f)                                   \\
		                           & = -g_{i}(f)
	\end{align*}
	As this expression does not depend on $S$, we get the required property.
	Now assume that the property on the removal operators holds,
	\textit{i.e.}~for any function $f \in \mathcal{F}$ with additive variable $x_i$:
	\begin{equation*}
		\forall S,T \subseteq [d] \setminus i: P_{T \cup i}(f) - P_T(f) = P_{S \cup i}(f) - P_S(f)
	\end{equation*}
	We must prove that $G$ is additivity-preserving.
	Let $S$ be a non-empty subset of $[d] \setminus i$.
	Then, using the given property on the subset operators and the definition of the canonical additive decomposition:
	\begin{align*}
		g_{S \cup i}(f) & = \sum_{T \subseteq S \cup i}(-1)^{|T|-|S|-1}P_{\overline{T}}(f)                                          \\
		                & = \sum_{T \subseteq S}(-1)^{|T|-|S|-1}\left(P_{\overline{T}}(f) - P_{\overline{T \cup i}}(f) \right)      \\
		                & = \sum_{T \subseteq S}(-1)^{|T|-|S|-1}\left(P_{\overline{T}}(f) - P_{\overline{T} \setminus i}(f) \right) \\
	\end{align*}
	From the assumption, we know that there exists a function $h \in \mathcal{F}$
	such that for any subset $T \subseteq [d]$ with $i \in T$:
	\begin{equation*}
		P_{T}(f) - P_{T \setminus i}(f) = h
	\end{equation*}
	As $S \subseteq [d] \setminus i$, we know that $\forall T \subseteq S: i \in \overline{T}$.
	Therefore, we can rewrite the sum as:
	\begin{align*}
		\sum_{T \subseteq S}(-1)^{|T|-|S|-1}\left(P_{\overline{T}}(f) - P_{\overline{T} \setminus i}(f) \right)
		 & = h \sum_{T \subseteq S}(-1)^{|T|-|S|-1} \\
		 & = - h \sum_{T \subseteq S}(-1)^{|T|-|S|} \\
		 & = 0
	\end{align*}
\end{proof}

\minimalimpliesadditivitypreserving*

\begin{proof}
	Assume $x_i$ is an additive variable for $f$.
	Then there exist functions $f_{i}, f_{-i} \in \mathcal{F}$ such that:
	\begin{equation*}
		f = f_i + f_{-i}
	\end{equation*}
	where $f_i$ is independent of $X_{[d]\setminus i}$ and $f_{-i}$ is independent of $X_{i}$.
	This implies that
	$\mathcal{U} := \{[d]\setminus i, \{i\}\}$
	is a dependency structure for $f$.
	Therefore, the minimal dependency structure $\mathcal{S}$ for $f$
	contains no strict supersets of $\{i\}$,
	since $\mathcal{S} \leq_{\text{DS}} \mathcal{U}$.
	Using \Cref{lemma:fdfi-minimal-cad-mds},
	we get that $g_S(f) = 0$ for every subset $S \subseteq [d]$ such that $i \in S, |S| > 1$.
	Therefore, $G$ is additivity-preserving.
\end{proof}

\symmetrypreservationnecsuffconditions*

\begin{proof}
	Assume that $G$ is a CAD with removal operators that have the assumed property.
	We will prove that $G$ is symmetry-preserving by induction.
	Assume $f \in \mathcal{F}$ is a function with symmetric variables $X_{i},X_{j}$,
	$\mathbf{x} \in \mathcal{X}$ with $x_{i} = x_{j}$
	and $S = \emptyset$.
	Then:
	\begin{align*}
		g_{S \cup i}(f)(\mathbf{x}) & = P_{\overline{i}}(f)(\mathbf{x}) - g_{\emptyset}(f)(\mathbf{x}) \\
		g_{S \cup j}(f)(\mathbf{x}) & = P_{\overline{j}}(f)(\mathbf{x}) - g_{\emptyset}(f)(\mathbf{x})
	\end{align*}
	Denote $T := [d] \setminus \{ i,j \}$.
	Then $P_{\overline{i}} = P_{T \cup j}$ and $P_{\overline{j}} = P_{T \cup i}$.
	Therefore, $P_{\overline{i}}(f)(\mathbf{x}) = P_{\overline{j}}(f)(\mathbf{x})$,
	and the required equality follows.
	Now assume that the proposition holds for all strict subsets $T \subset S$,
	for any $S \subseteq [d] \setminus \{ i,j \}$.
	Then:
	\begin{align*}
		g_{S \cup i}(f)(\mathbf{x})
		 & = P_{\overline{S \cup i}}(f)(\mathbf{x}) - \sum_{T \subset S \cup i}g_{T}(f)(\mathbf{x}) \\
		 & = P_{\overline{S} \setminus i}(f)(\mathbf{x})
		- \sum_{T \subseteq S} g_{T}(f)(\mathbf{x})
		- \sum_{T \subset S} g_{T \cup i}(f)(\mathbf{x})                                            \\
		 & = P_{\overline{S} \setminus j}(f)(\mathbf{x})
		- \sum_{T \subseteq S} g_{T}(f)(\mathbf{x})
		- \sum_{T \subset S} g_{T \cup j}(f)(\mathbf{x})                                            \\
		 & = P_{\overline{S} \setminus j}(f)(\mathbf{x})
		- \sum_{T \subset S \cup j} g_{T}(f)(\mathbf{x})                                            \\
		 & = g_{S \cup j}(f)(\mathbf{x})
	\end{align*}
	Now assume that $G$ is symmetry-preserving,
	$f \in \mathcal{F}$ is a function with symmetric variables $X_{i},X_{j}$,
	and $\mathbf{x} \in \mathcal{X}$ with $x_{i} = x_{j}$.
	We must prove that for any
	$S \subseteq [d] \setminus \{ i,j \}: P_{S \cup i}(f)(\mathbf{x}) = P_{S \cup j}(f)(\mathbf{x})$
	or equivalently, for any
	$S \subseteq [d], \{ i,j \} \subseteq S: P_{S \setminus i}(f)(\mathbf{x}) = P_{S \setminus j}(f)(\mathbf{x})$.
	Assume $S \subseteq [d]$ with $\{ i,j \} \subseteq S$. Then:
	\begin{align*}
		P_{S \setminus i}(f)(\mathbf{x})
		 & = \sum_{T \subseteq \overline{S \setminus i}}g_{T}(f)(\mathbf{x})                                  \\
		 & = \sum_{T \subseteq \overline{S} \cup i}g_{T}(f)(\mathbf{x})                                       \\
		 & = \sum_{T \subseteq \overline{S}}\left( g_{T}(f)(\mathbf{x}) + g_{T \cup i}(f)(\mathbf{x}) \right) \\
		 & = \sum_{T \subseteq \overline{S}}\left(g_{T}(f)(\mathbf{x}) + g_{T \cup j}(f)(\mathbf{x})\right)   \\
		 & = \sum_{T \subseteq \overline{S} \cup j}g_{T}(f)(\mathbf{x})                                       \\
		 & = P_{S \setminus j}(f)(\mathbf{x})
	\end{align*}
\end{proof}

\anonymityremovalconditions*

\begin{proof}
	We first prove $P_S(\pi f)(\pi \mathbf{x}) = P_{\pi S}(f)(\mathbf{x}) \implies g_S(\pi f)(\pi \mathbf{x}) = g_{\pi S}(f)(\mathbf{x})$:
	\begin{align*}
		g_S(\pi f)(\pi \mathbf{x}) & = \sum_{T \subseteq S} (-1)^{|T|-|S|} P_T(\pi f)(\pi \mathbf{x}) \\
		                           & = \sum_{T \subseteq S} (-1)^{|T|-|S|} P_{\pi T}(f)(\mathbf{x})   \\
		                           & = \sum_{T \subseteq \pi S} (-1)^{|T|-|S|} P_{T}(f)(\mathbf{x})   \\
		                           & = g_{\pi S}(f)(\mathbf{x})
	\end{align*}
	We now prove the reverse direction:
	\begin{align*}
		g_S(\pi f)(\pi \mathbf{x}) & = g_{\pi S}(f)(\mathbf{x})                                                              \\
		P_{\overline{S}}(\pi f)(\pi \mathbf{x})
		- \sum_{T \subset S} g_T(\pi f)(\pi \mathbf{x})
		                           & = P_{\overline{\pi S}}(f)(\mathbf{x}) - \sum_{T \subset \pi S} g_T(f)(\mathbf{x})       \\
		                           & = P_{\overline{\pi S}}(f)(\mathbf{x}) - \sum_{T \subset S} g_{\pi T}(f)(\mathbf{x})     \\
		                           & = P_{\overline{\pi S}}(f)(\mathbf{x}) - \sum_{T \subset S} g_{T}(\pi f)(\pi \mathbf{x}) \\
		P_{\overline{S}}(\pi f)(\pi \mathbf{x})
		                           & = P_{\overline{\pi S}}(f)(\mathbf{x})
	\end{align*}
\end{proof}

\idempotentimpliesunanimitygame*

\begin{proof}
	Assume $G$ is an idempotent additive decomposition on $\mathcal{F}$,
	$\Phi: \mathcal{F} \rightarrow \mathcal{F}, f \in \mathcal{F}, S \subseteq [d], \mathbf{x} \in \mathcal{X}$.
	Then for any subset $T \subseteq [d]$:
	\begin{equation*}
		\sum_{V \subseteq T}g_{V}(g_{S}(f)) = \begin{cases}
			g_{S}(f) & \mbox{ if } S \subseteq T \\
			0        & \mbox{ otherwise.}
		\end{cases}
	\end{equation*}
	Therefore:
	\begin{align*}
		v_{G}^{\Phi}(g_{S}(f),\mathbf{x})(T)
		 & = \Phi \left( \sum_{V \subseteq T}g_{V}(g_{T}(f)) \right)(\mathbf{x}) - \Phi(g_{\emptyset}(g_{S}(f))(\mathbf{x}) \\
		 & = \begin{cases}
			     \Phi \left( g_{S}(f) \right)(\mathbf{x}) - \Phi(f_{0})(\mathbf{x}) & \mbox{ if } S \subseteq T \\
			     0                                          & \mbox{ otherwise.}        \\
		     \end{cases}
	\end{align*}
	where $f_{0}(\mathbf{x}) = 0, \forall \mathbf{x} \in \mathcal{X}$.
\end{proof}

\rbamrepresentationtheorem*

\begin{proof}
	Assume $m$ is a removal-based attribution method
	on a function space $\mathcal{F}$
	with domain $\mathcal{X}$,
        behaviour mapping $\Phi$,
        aggregation coefficients $\left\{ \alpha_{T}^{S} \mid S,T \subseteq [d] \right\}$,
	and corresponding functional decomposition $G$.
        We then have:
	\begin{align*}
                m(f,S) &= \sum_{T \subseteq [d]} \alpha_{T}^{S}\Phi\left(P_{T}(f)\right)\\
                       &= \sum_{T \subseteq [d]} \alpha_{\overline{T}}^{S} \Phi\left(P_{\overline{T}}(f)\right)\\
                       &= \sum_{T \subseteq [d]} \alpha_{\overline{T}}^{S}\left[\Phi\left(P_{\overline{T}}(f)\right) - \Phi\left(P_{[d]}(f)\right) + \Phi\left(P_{[d]}(f)\right)\right]\\
                m(f,S)(\mathbf{x}) &= \sum_{T \subseteq [d]} \alpha_{\overline{T}}^{S}\left[\Phi\left(P_{\overline{T}}(f)\right)(\mathbf{x}) - \Phi\left(P_{[d]}(f)\right)(\mathbf{x}) + \Phi\left(P_{[d]}(f)\right)(\mathbf{x})\right]\\
                                   &= \sum_{T \subseteq [d]}\alpha_{\overline{T}}^{S} \left[v_{G}^{\Phi}(f,\mathbf{x})(T) + \Phi(g_{\emptyset}(f))(\mathbf{x}) \right]
	\end{align*}
\end{proof}

\begin{lemma}
	\label{lemma:fdfi-suff-cond-derivative-aggregation-constants}
	Assume $m$ is a removal-based attribution method
	with corresponding functional decomposition $G$,
	behaviour mapping $\Phi$,
	and aggregation coefficients $\left\{ \alpha_{T}^{S} \mid S,T \subseteq [d] \right\}$.
	If:
	\begin{equation*}
		\forall S \subseteq [d], i \in S, T \subseteq [d] \setminus i: \alpha_{T}^{S} = -\alpha_{T \cup i}^{S}
	\end{equation*}
	Then:
	\begin{equation*}
		m(f,S)(\mathbf{x})
		= \sum_{T \subseteq [d] \setminus S}\alpha_{\overline{T \cup S}}^{S} \Delta_{S}v_{G}^{\Phi}(f,\mathbf{x})(T)
	\end{equation*}
	for any $f \in \mathcal{F}, S \subseteq [d], \mathbf{x} \in \mathcal{X}$.
\end{lemma}

\begin{proof}
	This proof is based on the proof for Proposition 2 in \citet{grabisch1999}.

	Assume $f \in \mathcal{F}, S \subseteq [d], \mathbf{x} \in \mathcal{X}$.

	First, using the RBAM representation theorem (\Cref{thm:fdfi-rbam-representation-theorem}),
	we can write the attribution method as follows:
        \begin{equation*}
                m(f,S)(\mathbf{x}) = \sum_{T \subseteq [d]} \alpha_{\overline{T}}^{S} \left[v_{G}^{\Phi}(f,\mathbf{x})(T) + \Phi(g_{\emptyset}(f))(\mathbf{x})\right]
        \end{equation*}
        From the assumption that $\forall i \in S, T \subseteq [d] \setminus i: \alpha_{T}^{S} = -\alpha_{T \cup i}^{S}$,
        we get:
	\begin{align*}
		\forall L \subseteq S, T \subseteq [d] \setminus S:
                \alpha_{\overline{T \cup L}}^S & = (-1)^{|L|} \alpha_{\overline{T}}^S                \\
                                              & = (-1)^{|S|-|L|} \alpha_{\overline{T \cup S}}^{S}
        \end{align*}
	Substituting this in the expression for $m$:
        
        \begin{align*}
                m(f,S)(\mathbf{x}) &= \sum_{T \subseteq [d]} \alpha_{\overline{T}}^{S} \left[v_{G}^{\Phi}(f,\mathbf{x})(T) + \Phi(g_{\emptyset}(f))(\mathbf{x})\right]\\
                                   &= \sum_{T \subseteq [d] \setminus S} \sum_{L \subseteq S} \alpha_{\overline{T \cup L}}^{S}\left[v_{G}^{\Phi}(f,\mathbf{x})(T \cup L) + \Phi(g_{\emptyset}(f))(\mathbf{x})\right]\\
                                   &= \sum_{T \subseteq [d] \setminus S}\alpha_{\overline{T \cup S}}^{S} \sum_{L \subseteq S}(-1)^{|S|-|L|}[v_{G}^{\Phi}(f,\mathbf{x})(T \cup L) + \Phi(g_{\emptyset}(f))(\mathbf{x})]\\
                                   &= \sum_{T \subseteq [d] \setminus S} \alpha_{\overline{T \cup S}}^{S} \left[\Delta_{S}v_{G}^{\Phi}(f,\mathbf{x})(T) + \sum_{L \subseteq S}(-1)^{|S|-|L|}\Phi(g_{\emptyset}(f))(\mathbf{x}) \right]\\
                                   &= \sum_{T \subseteq [d] \setminus S} \alpha_{\overline{T \cup S}}^{S} \Delta_{S}v_{G}^{\Phi}(f,\mathbf{x})(T)
        \end{align*}
\end{proof}

\internalconsistencynecsuffconditions*

\begin{proof}
	Assume $m$ is a removal-based attribution method
	with corresponding functional decomposition $G$,
	behaviour mapping $\Phi$
	and aggregation coefficients $\left\{ \alpha_{T}^{S} \mid S,T \subseteq [d] \right\}$
	that is both feature-level and interaction-level consistent.
	For a fixed subset $S \subseteq [d]$, we can view the RBAM $m$
	as a vector function $\mathbb{R}^{2^{[d]}} \rightarrow \mathbb{R}$.
	To do this, we first assign an index $i(T)$ to each of the subsets $T \subseteq [d]$.
	As the elements of $[d]$ are the integers $1, \dots, d$,
	we can do this by \textit{e.g.}~representing each subset $T$
	as a binary vector $\mathbf{b}^{T} = (b^{T}_{j} \mid j \in 1, \dots, d)$
	such that
	\begin{equation*}
		b^{T}_{j} = \begin{cases}
			1 & \mbox{ if } j \in T \\
			0 & \mbox{ otherwise.}
		\end{cases}
	\end{equation*}
	We can then define $i(T)$ as the number for which $\mathbf{b}^{T}$ is the binary representation,
	\textit{i.e.}~$i(T) = \sum_{j \in [d]} 2^{j-1}b^{T}_{j}$.
	The index $i: 2^{[d]} \rightarrow [2^{d}]$ can be used to denote elements of vectors
	$\mathbf{z} \in \mathbb{R}^{2^{[d]}}$,
	\textit{i.e.}~$\mathbf{z} = (z_{i(T)} \mid T \subseteq [d])$.
	We will simplify this notation by writing $z_{T} := z_{i(T)}$.

	If we can show that
	\begin{equation*}
		\forall S \subseteq [d], i \in S, T \subseteq [d] \setminus i: \alpha_{T}^{S} = -\alpha_{T \cup i}^{S}
	\end{equation*}
	then the desired form follows from
	\Cref{lemma:fdfi-suff-cond-derivative-aggregation-constants}.
	However, it is possible for a RBAM to be internally consistent without having this property.
	For example, if the corresponding decomposition of $m$ is the trivial decomposition
	and $\Phi$ is the identity mapping,
	then it is easy to see that $m$ is necessarily internally consistent:
	\begin{itemize}
		\item If $f(\mathbf{x}) = 0$,
		      then all $\forall S \subseteq [d]: \Phi(P_{S}(f))(\mathbf{x}) = 0$,
		      and therefore $\forall S \subseteq [d]: m(f,S)(\mathbf{x}) = 0$.
		\item If $f(\mathbf{x}) \neq 0$,
		      then $\forall i \in [d]: \Phi(P_{\emptyset}(f))(\mathbf{x}) = f(\mathbf{x}) \neq 0 = \Phi(P_{i}(f))(\mathbf{x})$.
		      Therefore, there are no locally independent variables for $f$ at $\mathbf{x}$.
	\end{itemize}
	For a given subset $S \subseteq [d]$,
	we can view the attribution method $m$
	as a linear real-valued vector function $m_{S}: \mathbb{R}^{2^{d}} \rightarrow \mathbb{R}$:
	\begin{equation*}
		m_{S}(\mathbf{z}) = \sum_{T \subseteq [d]} \alpha_{T}^{S}\mathbf{z}_{T}
	\end{equation*}
	where each vector $\mathbf{z}$ contains the values $P_{T}(f)(\mathbf{x})$
	for all subsets $T$,
	for some function $f \in \mathcal{F}$ and point $\mathbf{x} \in \mathcal{X}$.
	We denote the set of such vectors as:
	\begin{equation*}
		\Lambda := \{ \left( \Phi(P_{T}(f))(\mathbf{x}) \mid T \subseteq [d] \right) \mid f \in \mathcal{F} \}
	\end{equation*}
	Note that the set $\Lambda$ does not necessarily span $\mathbb{R}^{2^{d}}$ entirely.
	In the previous example,
	any $\mathbf{z} \in \Lambda$ can only have non-zero entries
	corresponding to $T = \emptyset$.

	We will therefore construct a linear extension $\tilde{m}_S$ of $m_{S}$
	to $\mathbb{R}^{2^{d}}$ such that:
	\begin{align*}
		\tilde{m}_S(\mathbf{z})                                   & = \sum_{S \subseteq [d]}\beta_{T}^{S} \mathbf{z}_{T} \\
		\forall \mathbf{z} \in \Lambda: m_{S}(\mathbf{z})           & = \tilde{m}_S(\mathbf{z})                          \\
		\forall i \in S, T \subseteq [d] \setminus i: \beta_{T}^{S} & = -\beta_{T \cup i}^{S}
	\end{align*}
	then this linear extension fulfills the requirements
	for \Cref{lemma:fdfi-suff-cond-derivative-aggregation-constants},
	and the required result follows.

	We first define
	$\tilde{m}_S(\mathbf{z}) := m_{S}(\mathbf{z}), \forall \mathbf{z} \in \Lambda$.
	For a given point $\mathbf{x} \in \mathcal{X}$,
	define the set $\Lambda_\mathbf{x}$ as follows:
	\begin{equation*}
		\Lambda_\mathbf{x}
		:= \{(\Phi(P_{T}(f))(\mathbf{x}) \mid T \subseteq [d]) \mid f \in \mathcal{F}\}
	\end{equation*}
	Then $\Lambda = \bigcup_{\mathbf{x} \in \mathcal{X}} \Lambda_\mathbf{x}$,
	and $\tilde{m}_S$ has a unique linear extension to
	$\overline{\Lambda} := \text{span}(\bigcup_{\mathbf{x} \in \mathcal{X}} \Lambda_\mathbf{x})$.
	We will therefore reuse the notation $\tilde{m}_S$
	to denote this linear extension $\overline{\Lambda} \rightarrow \mathbb{R}$.
	By construction, $\overline{\Lambda}$ is a subspace of $\mathbb{R}^{2^d}$.
	We will now extend $\tilde{m}_S$ to $\mathbb{R}^{2^{d}}$.

	For $i \in S$, define the set $V_i$ as follows:
	\begin{equation*}
		V_i = \left\{\mathbf{v} \in \mathbb{R}^{2^d}
		\mid \forall T \subseteq [d] \setminus i: \mathbf{v}_T = \mathbf{v}_{T \cup i}\right\}
	\end{equation*}
	$V_i \cap \Lambda$ contains all vectors that correspond to a function $f$
	and point $\mathbf{x}$ at which $X_{i}$ is a locally independent variable.
	Therefore, internal consistency of $m$ implies that
	$V_i \cap \Lambda \subseteq \text{ker}(\tilde{m}_{S}), \forall i \in S$.
	For any $i \in S$, $V_i$ is a linear subspace of $\mathbb{R}^{2^d}$.
	This can be seen by constructing a basis for $V_i$.
	Define the vector $\mathbf{e}_i^T$ as follows:
	\begin{equation*}
		(\mathbf{e}_i^T)_R =
		\left\{
		\begin{array}{ll}
			1 & \mbox{if } R \in \{T, T \cup i\} \\
			0 & \mbox{otherwise}
		\end{array}
		\right.
	\end{equation*}
	Then it is easy to see that
	$\mathcal{B}_i := \{\mathbf{e}_i^T \mid T \subseteq [d] \setminus i\}$
	constitutes a basis for $V_i$.
	Because $V_i$ and $\overline{\Lambda}$ are both subspaces of $\mathbb{R}^{2^d}$,
	their intersection $V_i \cap \Lambda$ is also a subspace.

	We will now extend $\tilde{m}_S$ by defining its value on a basis for $\mathbb{R}^{2^d}$:
	\begin{enumerate}
		\item Choose a basis $\mathcal{B}_1$ for $\overline{\Lambda}$
		      and define $\tilde{m}_S(\mathbf{e}) = m_{S}(\mathbf{e})$
		      for any $\mathbf{e} \in \mathcal{B}_1$.
		      This ensures that $\tilde{m}_S$ is indeed an extension of $m_{S}$.
		      Note that this also implies that
		      $\tilde{m}_S(\mathbf{z}) = 0$
		      for any $\mathbf{z} \in \Lambda \cap V_i, i \in S$.
		\item Extend $\mathcal{B}_1$
		      to a basis $\mathcal{B}_2$
		      for $\Lambda + \sum_{i \in S} V_i$,
		      where $+$ and $\sum$ denote the sum of \textit{subspaces},
		      \textit{i.e.}~the span of the union of two subspaces.
		      Define $\tilde{m}_S(\mathbf{e}) = 0$
		      for any $\mathbf{e} \in \mathcal{B}_2 \setminus \mathcal{B}_1$.
		\item Finally, extend $\mathcal{B}_2$ to a basis $\mathcal{B}_3$
		      for $\mathbb{R}^{2^d}$.
		      The value for $\overline{m}_S(\mathbf{e})$
		      can be any arbitrary value for any
		      $\mathbf{e} \in \mathcal{B}_3 \setminus \mathcal{B}_2$.
	\end{enumerate}

	By construction, we have:
	\begin{itemize}
		\item $\tilde{m}_S$ is an extension of $m_S$.
		\item $\tilde{m}_S(\mathbf{z}) = 0$ for any $\mathbf{z} \in V_i, i \in S$.
		\item $\tilde{m}_S$ is linear:
		      $\tilde{m}_S(\mathbf{z}) = \sum_{T \subseteq [d]}\beta^S_T \mathbf{z}_T$
	\end{itemize}

	We will now use these properties to show that
	$\forall i \in S, T \subseteq [d] \setminus i: \beta_{T}^{S} = -\beta_{T \cup i}^{S}$.
	Choose $i \in S, T \subseteq [d] \setminus i$. Consider the vector $\mathbf{z}$:
	\begin{equation*}
		\mathbf{z}_R =
		\left\{
		\begin{array}{ll}
			1 & \mbox{if } R \in \{T, T \cup i\} \\
			0 & \mbox{otherwise}
		\end{array}
		\right.
	\end{equation*}
	Then $\mathbf{z} \in V_i$. Therefore:
	\begin{align*}
		\tilde{m}_S(\mathbf{z}) & = \sum_{T \subseteq [d]}\beta^S_T \mathbf{z}_T \\
		                          & = \beta_{T}^S + \beta_{T \cup i}^S = 0         \\
		\iff \beta_{T}^S          & = -\beta_{T \cup i}^S
	\end{align*}
	the required result now follows from
	\Cref{lemma:fdfi-suff-cond-derivative-aggregation-constants}.

	Now assume $m$ can be written as:
	\begin{equation*}
		m(f,S)(\mathbf{x}) = \sum_{T \subseteq [d] \setminus S}\beta_{T}^{S}\Delta_{S}v_{G}^{\Phi}(f,\mathbf{x})(T)
	\end{equation*}
        for some set of constants $\{ \beta_{T}^{S} \mid S \subseteq [d], T \subseteq [d] \setminus S \}$.
	We will then show that $m$ is internally consistent.
	Assume $i$ is a locally independent variable
	for $f \in \mathcal{F}$ at $\mathbf{x}$.
	We must show that, for any $S \subseteq [d]: i \in S \implies m(f,S)(\mathbf{x}) = 0$.
	From the definition of local independence
	and the pointwise cooperative game,
	we have for any $T \subseteq [d] \setminus i$:
	\begin{align*}
		\Phi(P_{T \cup i}(f))(\mathbf{x})
		 & = \Phi(P_{T}(f))(\mathbf{x})                                                        \\
		v_{G}^{\Phi}(f,\mathbf{x})(T)
		 & = \Phi( P_{\overline{T}}(f))(\mathbf{x}) - \Phi(P_{[d]}(f))(\mathbf{x})             \\
		 & = \Phi( P_{\overline{T} \setminus i}(f))(\mathbf{x}) - \Phi(P_{[d]}(f))(\mathbf{x}) \\
		 & = \Phi( P_{\overline{T \cup i}}(f))(\mathbf{x}) - \Phi(P_{[d]}(f))(\mathbf{x})      \\
		 & = v_{G}^{\Phi}(f,\mathbf{x})(T \cup i)
	\end{align*}
	Now assume $S \subseteq [d], i \in S$.
	Combining the previous equality with the definition of the discrete derivative
	$\Delta_{S}v_{G}^{\Phi}(f,\mathbf{x})(T)$,
	we obtain for any $T \subseteq [d] \setminus S$:
	\begin{align*}
		\Delta_{S}v_{G}^{\Phi}(f,\mathbf{x})(T)
		 & = \sum_{L \subseteq S}(-1)^{|S|-|L|} v_{G}^{\Phi}(f,\mathbf{x})(T \cup L) \\
		 & = \sum_{L \subseteq S \setminus i}
		\left[
		(-1)^{|S|-|L|}v_{G}^{\Phi}(f,\mathbf{x})(T \cup L)
		+ (-1)^{|S|-|L|-1}v_{G}^{\Phi}(f,\mathbf{x})((T \cup L) \cup i)
		\right]                                                                      \\
		 & = \sum_{L \subseteq S \setminus i}
		\left[
		(-1)^{|S|-|L|}v_{G}^{\Phi}(f,\mathbf{x})(T \cup L)
		+ (-1)^{|S|-|L|-1}v_{G}^{\Phi}(f,\mathbf{x})(T \cup L)
		\right]                                                                      \\
		 & = 0
	\end{align*}
	Therefore, every term in the definition for $m(f,S)(\mathbf{x})$ is zero,
	which implies that $m(f,S)(\mathbf{x}) = 0$.
\end{proof}

\mcvaluecomputation*

\begin{proof}
	Assume $m$ is a simple MC attribution method
	with corresponding functional decomposition $G$.
	Since the behaviour mapping $\Phi$ is simply the identity mapping,
	we will denote the pointwise cooperative game as $v_{G}$.
	From the definition of the discrete derivative, we get:
	\begin{align*}
		\Delta_{S}v_{G}(f,\mathbf{x})(\emptyset)
		 & = \sum_{T \subseteq S}(-1)^{|T|-|S|}v_{G}(f,\mathbf{x})(\emptyset)   \\
		 & = \sum_{T \subseteq S}(-1)^{|S|-|T|}
		\left(P_{\overline{T}}(f)(\mathbf{x}) - P_{[d]}(f)(\mathbf{x})\right)              \\
		 & = \sum_{T \subseteq S}(-1)^{|S|-|T|} P_{\overline{T}}(f)(\mathbf{x})
		- P_{[d]}(f)(\mathbf{x})\left(\sum_{T \subseteq S}(-1)^{|S|-|T|}\right)            \\
		 & = \sum_{T \subseteq S}(-1)^{|S|-|T|} P_{\overline{T}}(f)(\mathbf{x}) \\
		 & = g_{S}(f)(\mathbf{x})
	\end{align*}
	Combining this with \Cref{eqn:fdfi-discrete-derivative-emptyset},
	we obtain for any $T,S \subseteq [d]$:
	\begin{align*}
		\Delta_{S}v_{G}(f,\mathbf{x})(T)
		 & = \sum_{L \subseteq T}\Delta_{L \cup S}v_{G}(f,\mathbf{x})(\emptyset) \\
		 & = \sum_{L \subseteq T}g_{L \cup S}(f)(\mathbf{x})                     \\
	\end{align*}
	Since $\Delta_S v_{G}(f,\mathbf{x})(T)$
	is a linear combination of functional components $g_S(f)$,
	and $m(f,S)$ is a linear combination of $\Delta_S v_{G}(f,\mathbf{x})(T)$,
	we can already see that $m(f,S)$
	must be a linear combination of functional components $g_S(f)$ as well.
	We will now derive the exact coefficients of this linear combination.
	\begin{align*}
		m(f,S) & = \sum_{T \subseteq [d] \setminus S}\beta_T^S \Delta_S v_{G}(f,\mathbf{x})(T)                                                          \\
		       & = \sum_{T \subseteq [d] \setminus S}\beta_T^S \left[ \sum_{L \subseteq T} g_{L \cup S}(f)(\mathbf{x})\right]                           \\
		       & = \sum_{T \subseteq [d] \setminus S} \left[ \sum_{T \subseteq U \subseteq [d] \setminus S}\beta_U^S\right] g_{L \cup S}(f)(\mathbf{x}) \\
		       & = \sum_{T \subseteq [d] \setminus S} \overline{\beta}_T^S g_{L \cup S}(f)(\mathbf{x})
	\end{align*}
	where we define $\beta_{T}^{S} := \sum_{T \subseteq U \subseteq [d] \setminus S}\beta_U^S$.
\end{proof}

\begin{lemma}
	\label{lemma:additivity-preservation-dummy}
	Given a canonical additive decomposition $G \in \mathcal{D}_{\mathcal{F}}$,
	function $f \in \mathcal{F}$,
	$i \in [d]$ such that $X_{i}$ is an additive variable of $f$,
	and let $I$ be the identity mapping.
	If $G$ is additivity-preserving,
	then $i$ is a dummy player
	for the pointwise cooperative game
	$v_G^{I}(f,\mathbf{x})$ at any point $\mathbf{x} \in \mathcal{X}$.
\end{lemma}
\begin{proof}
	The pointwise cooperative game $v_{G}^{I}(f,\mathbf{x})(S)$
	is defined as:
	\begin{equation*}
		v_G^{I}(f,\mathbf{x})(S) = \sum_{T \subseteq S} g_T(f)(\mathbf{x}) - g_\emptyset(f)
	\end{equation*}
	Assume $S \subseteq [d] \setminus i$.
	Then:
	\begin{align*}
		v_G^{I}(f,\mathbf{x})(S \cup i)
		 & = \sum_{T \subseteq S \cup i}g_{T}(f)(\mathbf{x}) - g_\emptyset(f)(\mathbf{x})  \\
		 & = \sum_{T \subseteq S} g_T(f)(\mathbf{x})
		+ \sum_{T \subseteq S} g_{T \cup i}(f)(\mathbf{x})
		- g_\emptyset(f)                                                                   \\
		 & = \sum_{T \subseteq S} g_T(f)(\mathbf{x}) + g_i(f)(\mathbf{x}) - g_\emptyset(f) \\
		 & = \sum_{T \subseteq S} g_T(f)(\mathbf{x})
		- g_\emptyset(f)
		+ \sum_{T \subseteq \{i\}} g_T(f)(\mathbf{x})- g_\emptyset(f)                      \\
		 & = v_{G}^{I}(f,\mathbf{x})(S) + v_G^{I}(f,\mathbf{x})(i)
	\end{align*}
\end{proof}

\suffcondfuncdummy*

\begin{proof}
	Given $f \in \mathcal{F}$,
	$i \in S \subseteq [d]$,
	$f$ is additive in $x_i$.
	Assume $m$ is a simple removal-based attribution method
	that adheres to the Pointwise Dummy axiom,
	and the corresponding functional decomposition
	$G_m := \{g_{S}|S \subseteq [d]\}$ is additivity-preserving.
	We must prove that $m(f,i)$ is independent of $X_{[d] \setminus i}$ and:
	\begin{equation*}
		\forall S \subseteq [d]: S \neq \emptyset \implies m(f,S \cup i) = 0
	\end{equation*}
	Let $\mathbf{x} \in \mathcal{X}$.
	Consider the corresponding pointwise cooperative game $v_G^I(f,\mathbf{x})$.
	From Lemma \ref{lemma:additivity-preservation-dummy},
	we know that $i$ is a dummy player for this game.
	Because $m$ adheres to the Pointwise Dummy axiom
	and $i$ is a dummy player for the pointwise cooperative game $v_G^I(f,\mathbf{x})$,
	we have:
	\begin{align*}
		m(f,i)(\mathbf{x})
		                                        & = v_G^I(f,\mathbf{x})(i)                          \\
		                                        & = g_i(f)(\mathbf{x}) - g_\emptyset(f)(\mathbf{x}) \\
		\forall S \neq \emptyset: m(f,S \cup i) & = 0
	\end{align*}
	Since $g_i(f)$ is independent of $\mathbf{X}_{[d]\setminus i}$,
	we get that $m(f,i)$ is independent of $\mathbf{X}_{[d] \setminus i}$.
\end{proof}

\begin{lemma}
	\label{lemma:fdfi-independence-preservation-null}
	Given a canonical additive decomposition $G \in \mathcal{D}_{\mathcal{F}}$,
	function $f \in \mathcal{F}$,
	behaviour mapping $\Phi$,
	$i \in [d]$ such that $X_{i}$ is an independent variable of $f$.
	If $G$ is independence-preserving,
	then $i$ is a null player for the pointwise cooperative game $v_G^\Phi(f,\mathbf{x})$
	at any point $\mathbf{x} \in \mathcal{X}$.
\end{lemma}

\begin{proof}
	Assume $S \subseteq [d] \setminus i$.
	Then:
	\begin{align*}
		v_G^{\Phi}(f,\mathbf{x})(S \cup i)
		 & = \Phi \left( \sum_{T \subseteq S \cup i}g_{T}(f) \right)(\mathbf{x})
		- \Phi(g_{\emptyset}(f))(\mathbf{x})                                     \\
		 & = \Phi \left(
		\sum_{T \subseteq S} g_T(f) + \sum_{T \subseteq S} g_{T \cup i}(f)\right)(\mathbf{x})
		- \Phi(g_\emptyset(f))(\mathbf{x})                                       \\
		 & = \Phi \left(
		\sum_{T \subseteq S} g_T(f) \right)(\mathbf{x})
		- \Phi(g_\emptyset(f))(\mathbf{x})                                       \\
		 & = v_G^\Phi(f,\mathbf{x})(S)
	\end{align*}
\end{proof}

\pointwisenullipimpliesfuncnull*

\begin{proof}
	Given $f \in \mathcal{F}, S \subseteq [d] \setminus i, f$ is independent of $X_i$.
	We must prove that $m(f,S \cup i) = 0$,
	\textit{i.e.}~$\forall \mathbf{x} \in \mathcal{X}: m(f,S \cup i)(\mathbf{x}) = 0$.
	Choose any value for $\mathbf{x}$.
	From the independence preservation property of $G$
	and the fact that $f$ is independent of $X_i$,
	we know that $i$ is a null player
	in the pointwise cooperative game $v_G^{\phi}(f,\mathbf{x})$
	(\Cref{lemma:fdfi-independence-preservation-null}).
	Since $i$ is a null player in $v_G^\Phi(f,\mathbf{x})$
	and $m$ adheres to the Pointwise Null axiom,
	we know that:
	\begin{equation*}
		m(f,S \cup i)(\mathbf{x}) = 0
	\end{equation*}
\end{proof}

\begin{lemma}
	\label{lemma:fdfi-symmetry-preservation-symmetry}
	Given a canonical additive decomposition $G \in \mathcal{D}_{\mathcal{F}}$,
	function $f \in \mathcal{F}$,
	$i,j \in [d]: X_{i} \sim_{f} X_{j}$.
	If $G$ is symmetry-preserving,
	then $i$ and $j$ are symmetric
	in the pointwise cooperative game
	$v_{G}^{I}(f,\mathbf{x})$
	at any point $\mathbf{x} \in \mathcal{X}$ with $x_{i}=x_{j}$.
\end{lemma}

\begin{proof}
	Assume $S \subseteq [d] \setminus \{ i,j \}$.
	Then:
	\begin{align*}
		v_G^{I}(f,\mathbf{x})(S \cup i)
		 & = \sum_{T \subseteq S \cup i}g_{T}(f)(\mathbf{x})
		- g_{\emptyset}(f)(\mathbf{x})                       \\
		 & = \sum_{T \subseteq S}g_{T}(f)
		+ \sum_{T \subseteq S}g_{T \cup i}(f)
		(\mathbf{x})
		- g_{\emptyset}(f)(\mathbf{x})                       \\
		 & = \sum_{T \subseteq S}g_{T}(f)
		+ \sum_{T \subseteq S}g_{T \cup j}(f)
		(\mathbf{x})
		- g_{\emptyset}(f)(\mathbf{x})                       \\
		 & = \sum_{T \subseteq S \cup j}g_{T}(f)(\mathbf{x})
		- g_{\emptyset}(f)(\mathbf{x})                       \\
		 & = v_G^{I}(f,\mathbf{x})(S \cup j)
	\end{align*}
\end{proof}

\suffcondfuncsymmetry*

\begin{proof}
	Given $f \in \mathcal{F}, \mathbf{x} \in \mathcal{X}, X_{i} \sim_f X_j, x_{i} = x_{j}$,
	$m$ is a simple RBAM that adheres to the Pointwise Symmetry axiom and
	$G$ is symmetry-preserving.
	Assume $S \subseteq [d] \setminus \{i,j\}$.
	We must prove:
	\begin{equation*}
		m(f,S \cup i)(\mathbf{x}) = m(f,S \cup j)(\mathbf{x})
	\end{equation*}
	From Lemma \ref{lemma:fdfi-symmetry-preservation-symmetry}
	we know that $i$ and $j$ are symmetric
	in the pointwise cooperative game $v_{G}^I(f,\mathbf{x})$.
	The claim then follows from this fact
	and the fact that $m$ adheres to the Pointwise Symmetry axiom.
\end{proof}

\suffcondfuncanonymity*

\begin{proof}
	Given a function $f \in \mathcal{F}$,
	permutation $\pi \in \Pi([d])$,
	subset $S \subseteq [d]$
	and point $\mathbf{x} \in \mathcal{X}$.
	We must prove:
	\begin{equation*}
		m(\pi f, S)(\pi \mathbf{x}) = m(f, \pi S)(\mathbf{x})
	\end{equation*}
	From the definition of the pointwise cooperative game
	and the fact that $m$ is a simple RBAM, we have:
	\begin{align*}
		v_G^I(f,\mathbf{x})(S)
		 & = \sum_{T \subseteq S}g_{T}(f)(\mathbf{x})
		- g_{\emptyset}(f)(\mathbf{x})                \\
	\end{align*}
	Using the definition of a permuted game:
	\begin{align*}
		\pi v_G^I(f,\mathbf{x})(S)
		 & = v_G^I(f,\mathbf{x})(\pi^{-1} S)                   \\
		 & = \sum_{T \subseteq \pi^{-1} S}g_{T}(f)(\mathbf{x})
		- g_{\emptyset}(f)(\mathbf{x})                         \\
	\end{align*}
	Combining this with the anonymity of $G$, we get:
	\begin{align*}
		\pi v_{G}^I(\pi f,\pi \mathbf{x})(S)
		 & = \sum_{T \subseteq \pi^{-1} S}g_{T}(\pi f)(\pi \mathbf{x})
		- g_{\emptyset}(\pi f)(\mathbf{x})                             \\
		 & = \sum_{T \subseteq \pi^{-1} S}g_{\pi T}(f)(\mathbf{x})
		- g_{\emptyset}(f)(\mathbf{x})                                 \\
		 & = \pi v_{G}^I(f,\mathbf{x})(S)
	\end{align*}
	Consider the pointwise interaction index $\phi_{S}^{m}(v)$ of $m$:
	\begin{equation*}
		m(f,S)(\mathbf{x}) = \phi_{S}^{m}\left(v_{G}^{I}(f,\mathbf{x})\right)
	\end{equation*}
	Since $m$ adheres to the Pointwise Anonymity axiom,
	this implies that $\phi_{S}^{m}$ is anonymous:
	\begin{align*}
		m(f,S)(\mathbf{x})
		 & = \phi_{S}^{m}\left(v_{G}^{I}(f,\mathbf{x})\right) \\
		 & = \phi_{\pi S}^{m}(\pi v_{G}^I(f,\mathbf{x}))      \\
	\end{align*}
	Therefore:
	\begin{align*}
		m(\pi f,S)(\pi \mathbf{x}) & = \phi_{S}^{m}(v_{G}^I(\pi f,\pi \mathbf{x}))                                        \\
		                           & = \phi_{\pi S}^{m}(\pi v_{G}^I(\pi f,\pi \mathbf{x})) & \text{(Pointwise Anonymity)} \\
		                           & = \phi_{\pi S}^{m}(v_{G}^I(f,\mathbf{x}))             & \text{(anonymity (CAD))}     \\
		                           & = m(f,\pi S)(\mathbf{x})
	\end{align*}
\end{proof}

\bibliography{bibliography}

\begin{thebibliography}{75}
\providecommand{\natexlab}[1]{#1}
\providecommand{\url}[1]{\texttt{#1}}
\expandafter\ifx\csname urlstyle\endcsname\relax
  \providecommand{\doi}[1]{doi: #1}\else
  \providecommand{\doi}{doi: \begingroup \urlstyle{rm}\Url}\fi

\bibitem[Ancona et~al.(2018)Ancona, Ceolini, Öztireli, and Gross]{ancona2018}
Marco Ancona, Enea Ceolini, Cengiz Öztireli, and Markus Gross.
\newblock Towards better understanding of gradient-based attribution methods for deep neural networks.
\newblock In \emph{International Conference on Learning Representations}, 2018.
\newblock URL \url{https://openreview.net/forum?id=Sy21R9JAW}.

\bibitem[Banzhaf(1965)]{banzhaf1964}
John~F. Banzhaf.
\newblock Weighted voting doesn't work: A mathematical analysis.
\newblock \emph{Rutgers Law Review}, 19:\penalty0 317--343, 1965.

\bibitem[{Barredo Arrieta} et~al.(2020){Barredo Arrieta}, D\'{\i}az-Rodr\'{\i}guez, {Del Ser}, Bennetot, Tabik, Barbado, Garc\'{\i}a, Gil-L\'{o}pez, Molina, Benjamins, Chatila, and Herrera]{arrieta2019}
Alejandro {Barredo Arrieta}, Natalia D\'{\i}az-Rodr\'{\i}guez, Javier {Del Ser}, Adrien Bennetot, Siham Tabik, Alberto Barbado, Salvador Garc\'{\i}a, Sergio Gil-L\'{o}pez, Daniel Molina, Richard Benjamins, Raja Chatila, and Francisco Herrera.
\newblock Explainable artificial intelligence (xai): Concepts, taxonomies, opportunities and challenges toward responsible ai.
\newblock \emph{Information Fusion}, 58:\penalty0 82--115, 2020.
\newblock ISSN 1566-2535.
\newblock \doi{https://doi.org/10.1016/j.inffus.2019.12.012}.
\newblock URL \url{https://www.sciencedirect.com/science/article/pii/S1566253519308103}.

\bibitem[Bordt and von Luxburg(2023)]{bordt2023}
Sebastian Bordt and Ulrike von Luxburg.
\newblock From shapley values to generalized additive models and back.
\newblock In Francisco Ruiz, Jennifer Dy, and Jan-Willem van~de Meent, editors, \emph{Proceedings of The 26th International Conference on Artificial Intelligence and Statistics}, volume 206 of \emph{Proceedings of Machine Learning Research}, pages 709--745. PMLR, 25--27 Apr 2023.
\newblock URL \url{https://proceedings.mlr.press/v206/bordt23a.html}.

\bibitem[Breiman(2001)]{breiman2001a}
Leo Breiman.
\newblock Random forests.
\newblock \emph{Machine Learning}, 45:\penalty0 5--32, 2001.
\newblock ISSN 08856125.
\newblock \doi{10.1023/A:1010933404324}.
\newblock URL \url{http://link.springer.com/10.1023/A:1010933404324}.

\bibitem[Casalicchio et~al.(2019)Casalicchio, Molnar, and Bischl]{casalicchio2019}
Giuseppe Casalicchio, Christoph Molnar, and Bernd Bischl.
\newblock Visualizing the feature importance for black box models.
\newblock \emph{arXiv:1804.06620 [cs, stat]}, 11051:\penalty0 655--670, 2019.
\newblock \doi{10.1007/978-3-030-10925-7\_40}.
\newblock URL \url{http://arxiv.org/abs/1804.06620}.

\bibitem[Chen et~al.(2020)Chen, Janizek, Lundberg, and Lee]{chen2020}
Hugh Chen, Joseph~D. Janizek, Scott Lundberg, and Su-In Lee.
\newblock True to the model or true to the data?
\newblock \emph{arXiv preprint arXiv:2006.16234}, 6 2020.
\newblock URL \url{http://arxiv.org/abs/2006.16234}.

\bibitem[Covert et~al.(2020)Covert, Lundberg, and Lee]{covert2020}
Ian Covert, Scott Lundberg, and Su-In Lee.
\newblock Understanding global feature contributions with additive importance measures.
\newblock \emph{arXiv:2004.00668 [cs, stat]}, 10 2020.
\newblock URL \url{http://arxiv.org/abs/2004.00668}.

\bibitem[Covert et~al.(2021)Covert, Lundberg, and Lee]{covert2021}
Ian Covert, Scott Lundberg, and Su-In Lee.
\newblock Explaining by removing: A unified framework for model explanation.
\newblock \emph{Journal of Machine Learning Research}, 22:\penalty0 1--90, 2021.
\newblock ISSN 1533-7928.
\newblock URL \url{http://jmlr.org/papers/v22/20-1316.html}.

\bibitem[Cox(1984)]{cox1984}
D.~R. Cox.
\newblock Interaction.
\newblock \emph{International Statistical Review / Revue Internationale de Statistique}, 52:\penalty0 1--24, 1984.
\newblock ISSN 0306-7734.
\newblock \doi{10.2307/1403235}.
\newblock URL \url{https://www.jstor.org/stable/1403235}.

\bibitem[Datta et~al.(2016)Datta, Sen, and Zick]{datta2016}
Anupam Datta, Shayak Sen, and Yair Zick.
\newblock Algorithmic transparency via quantitative input influence: Theory and experiments with learning systems.
\newblock In \emph{2016 IEEE Symposium on Security and Privacy (SP)}, pages 598--617, 2016.
\newblock \doi{10.1109/SP.2016.42}.

\bibitem[Dubey(1975)]{dubey1975}
P.~Dubey.
\newblock On the uniqueness of the shapley value.
\newblock \emph{International Journal of Game Theory}, 4:\penalty0 131--139, 9 1975.
\newblock \doi{10.1007/bf01780630}.
\newblock URL \url{http://dx.doi.org/10.1007/bf01780630}.

\bibitem[Dubey et~al.(1981)Dubey, Neyman, and Weber]{dubey1981}
Pradeep Dubey, Abraham Neyman, and Robert~James Weber.
\newblock Value theory without efficiency.
\newblock \emph{Mathematics of Operations Research}, 6:\penalty0 122--128, 1981.
\newblock ISSN 0364-765X.
\newblock URL \url{https://www.jstor.org/stable/3689271}.

\bibitem[Frye et~al.(2020)Frye, Rowat, and Feige]{frye2020}
Christopher Frye, Colin Rowat, and Ilya Feige.
\newblock Asymmetric shapley values: incorporating causal knowledge into model-agnostic explainability.
\newblock In H.~Larochelle, M.~Ranzato, R.~Hadsell, M.F. Balcan, and H.~Lin, editors, \emph{Advances in Neural Information Processing Systems}, volume~33, pages 1229--1239. Curran Associates, Inc., 2020.
\newblock URL \url{https://proceedings.neurips.cc/paper_files/paper/2020/file/0d770c496aa3da6d2c3f2bd19e7b9d6b-Paper.pdf}.

\bibitem[Fujimoto et~al.(2006)Fujimoto, Kojadinovic, and Marichal]{fujimoto2006}
Katsushige Fujimoto, Ivan Kojadinovic, and Jean-Luc Marichal.
\newblock Axiomatic characterizations of probabilistic and cardinal-probabilistic interaction indices.
\newblock \emph{Games and Economic Behavior}, 55:\penalty0 72--99, 4 2006.
\newblock ISSN 0899-8256.
\newblock \doi{10.1016/j.geb.2005.03.002}.
\newblock URL \url{https://www.sciencedirect.com/science/article/pii/S0899825605000278}.

\bibitem[Gevaert and Saeys(2022)]{gevaert2022}
Arne Gevaert and Yvan Saeys.
\newblock Pdd-shap: Fast approximations for shapley values using functional decomposition.
\newblock In \emph{International Workshops of ECML PKDD 2022, Proceedings}, page~12, 2022.
\newblock URL \url{https://hal.science/hal-03773430/document}.

\bibitem[Gevaert et~al.(2023)Gevaert, Saranti, Holzinger, and Saeys]{gevaert2023}
Arne Gevaert, Anna Saranti, Andreas Holzinger, and Yvan Saeys.
\newblock Efficient approximation of asymmetric shapley values using functional decomposition.
\newblock In Andreas Holzinger, Peter Kieseberg, Federico Cabitza, Andrea Campagner, A.~Min Tjoa, and Edgar Weippl, editors, \emph{Machine Learning and Knowledge Extraction}, pages 13--30, Cham, 2023. Springer Nature Switzerland.
\newblock ISBN 978-3-031-40837-3.

\bibitem[Gevaert et~al.(2024)Gevaert, Rousseau, Becker, Valkenborg, De~Bie, and Saeys]{gevaert2024}
Arne Gevaert, Axel-Jan Rousseau, Thijs Becker, Dirk Valkenborg, Tijl De~Bie, and Yvan Saeys.
\newblock Evaluating feature attribution methods in the image domain.
\newblock \emph{Machine Learning}, May 2024.
\newblock ISSN 1573-0565.
\newblock \doi{10.1007/s10994-024-06550-x}.
\newblock URL \url{http://dx.doi.org/10.1007/s10994-024-06550-x}.

\bibitem[Grabisch and Roubens(1999)]{grabisch1999}
Michel Grabisch and Marc Roubens.
\newblock An axiomatic approach to the concept of interaction among players in cooperative games.
\newblock \emph{International Journal of Game Theory}, 28:\penalty0 547--565, 11 1999.
\newblock ISSN 1432-1270.
\newblock \doi{10.1007/s001820050125}.
\newblock URL \url{https://doi.org/10.1007/s001820050125}.

\bibitem[Grabisch et~al.(2000)Grabisch, Marichal, and Roubens]{grabisch2000}
Michel Grabisch, Jean-Luc Marichal, and Marc Roubens.
\newblock Equivalent representations of set functions.
\newblock \emph{Mathematics of Operations Research}, 25:\penalty0 157--178, 2000.
\newblock ISSN 0364-765X.
\newblock URL \url{https://www.jstor.org/stable/3690575}.

\bibitem[Graham et~al.(1995)Graham, Gr\"{o}tschel, and Lov\'{a}sz]{graham1995}
Ronald~L. Graham, Martin Gr\"{o}tschel, and L\'{a}szl\'{o} Lov\'{a}sz.
\newblock \emph{Handbook of combinatorics}.
\newblock Elsevier : MIT Press, 1995.
\newblock ISBN 978-0-444-88002-4 978-0-262-07169-7 978-0-444-82346-5 978-0-262-07170-3 978-0-444-82351-9 978-0-262-07171-0.

\bibitem[Guyon and Elisseeff(2003)]{guyon2003}
Isabelle Guyon and Andre Elisseeff.
\newblock An introduction to variable and feature selection.
\newblock \emph{Journal of Machine Learning Research}, 3:\penalty0 1157--1182, 2003.
\newblock ISSN 1533-7928.
\newblock URL \url{https://www.jmlr.org/papers/volume3/guyon03a/guyon03a.pdf?trk=public_post_comment-text}.

\bibitem[Harsanyi(1963)]{harsanyi1963}
John~C. Harsanyi.
\newblock A simplified bargaining model for the n-person cooperative game.
\newblock \emph{International Economic Review}, 4:\penalty0 194, 5 1963.
\newblock \doi{10.2307/2525487}.
\newblock URL \url{http://dx.doi.org/10.2307/2525487}.

\bibitem[Hedstr{\"o}m et~al.(2023)Hedstr{\"o}m, Bommer, Wickstr{\o}m, Samek, Lapuschkin, and H{\"o}hne]{hedstrom2023}
Anna Hedstr{\"o}m, Philine~Lou Bommer, Kristoffer~Knutsen Wickstr{\o}m, Wojciech Samek, Sebastian Lapuschkin, and Marina~MC H{\"o}hne.
\newblock The meta-evaluation problem in explainable {AI}: Identifying reliable estimators with metaquantus.
\newblock \emph{Transactions on Machine Learning Research}, 2023.
\newblock ISSN 2835-8856.
\newblock URL \url{https://openreview.net/forum?id=j3FK00HyfU}.

\bibitem[Hedstr\"{o}m et~al.(2023)Hedstr\"{o}m, Weber, Krakowczyk, Bareeva, Motzkus, Samek, Lapuschkin, and H\"{o}hne]{hedstrom2023b}
Anna Hedstr\"{o}m, Leander Weber, Daniel Krakowczyk, Dilyara Bareeva, Franz Motzkus, Wojciech Samek, Sebastian Lapuschkin, and Marina M.-C. H\"{o}hne.
\newblock Quantus: An explainable ai toolkit for responsible evaluation of neural network explanations and beyond.
\newblock \emph{Journal of Machine Learning Research}, 24\penalty0 (34):\penalty0 1--11, 2023.
\newblock URL \url{http://jmlr.org/papers/v24/22-0142.html}.

\bibitem[Herren and Hahn(2022)]{herren2022}
Andrew Herren and P.~Richard Hahn.
\newblock Statistical aspects of shap: Functional anova for model interpretation, 2022.
\newblock URL \url{https://arxiv.org/abs/2208.09970}.

\bibitem[Heskes et~al.(2020)Heskes, Sijben, Bucur, and Claassen]{heskes2020}
Tom Heskes, Evi Sijben, Ioan~Gabriel Bucur, and Tom Claassen.
\newblock Causal shapley values: Exploiting causal knowledge to explain individual predictions of complex models.
\newblock \emph{arXiv:2011.01625 [cs]}, 11 2020.
\newblock URL \url{http://arxiv.org/abs/2011.01625}.

\bibitem[Hiabu et~al.(2023)Hiabu, Meyer, and Wright]{hiabu2023}
Munir Hiabu, Joseph~T. Meyer, and Marvin~N. Wright.
\newblock Unifying local and global model explanations by functional decomposition of low dimensional structures.
\newblock In Francisco Ruiz, Jennifer Dy, and Jan-Willem van~de Meent, editors, \emph{Proceedings of The 26th International Conference on Artificial Intelligence and Statistics}, volume 206 of \emph{Proceedings of Machine Learning Research}, pages 7040--7060. PMLR, 25--27 Apr 2023.
\newblock URL \url{https://proceedings.mlr.press/v206/hiabu23a.html}.

\bibitem[Hoeffding(1948)]{hoeffding1948}
Wassily Hoeffding.
\newblock A class of statistics with asymptotically normal distribution.
\newblock \emph{The Annals of Mathematical Statistics}, 19:\penalty0 293--325, 9 1948.
\newblock ISSN 0003-4851, 2168-8990.
\newblock \doi{10.1214/aoms/1177730196}.
\newblock URL \url{https://projecteuclid.org/journals/annals-of-mathematical-statistics/volume-19/issue-3/A-Class-of-Statistics-with-Asymptotically-Normal-Distribution/10.1214/aoms/1177730196.full}.

\bibitem[Holzinger(2021)]{holzinger2021c}
Andreas Holzinger.
\newblock The next frontier: Ai we can really trust.
\newblock In Michael Kamp, Irena Koprinska, Adrien Bibal, Tassadit Bouadi, Beno{\^i}t Fr{\'e}nay, Luis Gal{\'a}rraga, Jos{\'e} Oramas, Linara Adilova, Yamuna Krishnamurthy, Bo~Kang, Christine Largeron, Jefrey Lijffijt, Tiphaine Viard, Pascal Welke, Massimiliano Ruocco, Erlend Aune, Claudio Gallicchio, Gregor Schiele, Franz Pernkopf, Michaela Blott, Holger Fr{\"o}ning, G{\"u}nther Schindler, Riccardo Guidotti, Anna Monreale, Salvatore Rinzivillo, Przemyslaw Biecek, Eirini Ntoutsi, Mykola Pechenizkiy, Bodo Rosenhahn, Christopher Buckley, Daniela Cialfi, Pablo Lanillos, Maxwell Ramstead, Tim Verbelen, Pedro~M. Ferreira, Giuseppina Andresini, Donato Malerba, Ib{\'e}ria Medeiros, Philippe Fournier-Viger, M.~Saqib Nawaz, Sebastian Ventura, Meng Sun, Min Zhou, Valerio Bitetta, Ilaria Bordino, Andrea Ferretti, Francesco Gullo, Giovanni Ponti, Lorenzo Severini, Rita Ribeiro, Jo{\~a}o Gama, Ricard Gavald{\`a}, Lee Cooper, Naghmeh Ghazaleh, Jonas Richiardi, Damian Roqueiro, Diego Saldana~Miranda, Konstantinos Sechidis, and Guilherme Gra{\c{c}}a, editors, \emph{Machine Learning and Principles and Practice of Knowledge Discovery in Databases}, pages 427--440, Cham, 2021. Springer International Publishing.
\newblock ISBN 978-3-030-93736-2.

\bibitem[Hooker(2004)]{hooker2004a}
Giles Hooker.
\newblock \emph{Diagnostics and extrapolation in machine learning}.
\newblock PhD thesis, Stanford University, Stanford, CA, USA, 2004.
\newblock AAI3145521.

\bibitem[Hooker(2007)]{hooker2007}
Giles Hooker.
\newblock Generalized functional anova diagnostics for high-dimensional functions of dependent variables.
\newblock \emph{Journal of Computational and Graphical Statistics}, 16:\penalty0 709--732, 9 2007.
\newblock ISSN 1061-8600, 1537-2715.
\newblock \doi{10.1198/106186007X237892}.
\newblock URL \url{http://www.tandfonline.com/doi/abs/10.1198/106186007X237892}.

\bibitem[Janizek et~al.(2021)Janizek, Sturmfels, and Lee]{janizek2021}
Joseph~D. Janizek, Pascal Sturmfels, and Su-In Lee.
\newblock Explaining explanations: Axiomatic feature interactions for deep networks.
\newblock \emph{Journal of Machine Learning Research}, 22\penalty0 (104):\penalty0 1--54, 2021.
\newblock URL \url{http://jmlr.org/papers/v22/20-1223.html}.

\bibitem[Karczmarz et~al.(2022)Karczmarz, Michalak, Mukherjee, Sankowski, and Wygocki]{karczmarz2022}
Adam Karczmarz, Tomasz Michalak, Anish Mukherjee, Piotr Sankowski, and Piotr Wygocki.
\newblock Improved feature importance computation for tree models based on the banzhaf value.
\newblock In James Cussens and Kun Zhang, editors, \emph{Proceedings of the Thirty-Eighth Conference on Uncertainty in Artificial Intelligence}, volume 180 of \emph{Proceedings of Machine Learning Research}, pages 969--979. PMLR, 01--05 Aug 2022.
\newblock URL \url{https://proceedings.mlr.press/v180/karczmarz22a.html}.

\bibitem[Kohavi and John(1997)]{kohavi1997}
Ron Kohavi and George~H. John.
\newblock Wrappers for feature subset selection.
\newblock \emph{Artificial Intelligence}, 97:\penalty0 273--324, 12 1997.
\newblock ISSN 0004-3702.
\newblock \doi{10.1016/S0004-3702(97)00043-X}.
\newblock URL \url{https://www.sciencedirect.com/science/article/pii/S000437029700043X}.

\bibitem[Kumar et~al.(2020)Kumar, Venkatasubramanian, Scheidegger, and Friedler]{kumar2020}
I.~Elizabeth Kumar, Suresh Venkatasubramanian, Carlos Scheidegger, and Sorelle Friedler.
\newblock Problems with shapley-value-based explanations as feature importance measures.
\newblock \emph{arXiv:2002.11097 [cs, stat]}, 6 2020.
\newblock URL \url{http://arxiv.org/abs/2002.11097}.

\bibitem[Kuo et~al.(2010)Kuo, Sloan, Wasilkowski, and Wo\'{z}niakowski]{kuo2010}
F.~Kuo, I.~Sloan, G.~Wasilkowski, and H.~Wo\'{z}niakowski.
\newblock On decompositions of multivariate functions.
\newblock \emph{Mathematics of Computation}, 79:\penalty0 953--966, 4 2010.
\newblock ISSN 0025-5718, 1088-6842.
\newblock \doi{10.1090/S0025-5718-09-02319-9}.
\newblock URL \url{https://www.ams.org/mcom/2010-79-270/S0025-5718-09-02319-9/}.

\bibitem[Kwon and Zou(2022)]{kwon2022a}
Yongchan Kwon and James~Y Zou.
\newblock Weightedshap: analyzing and improving shapley based feature attributions.
\newblock In S.~Koyejo, S.~Mohamed, A.~Agarwal, D.~Belgrave, K.~Cho, and A.~Oh, editors, \emph{Advances in Neural Information Processing Systems}, volume~35, pages 34363--34376. Curran Associates, Inc., 2022.
\newblock URL \url{https://proceedings.neurips.cc/paper_files/paper/2022/file/de1739eba209c682a90ec3669229ab2d-Paper-Conference.pdf}.

\bibitem[Lipovetsky and Conklin(2001)]{lipovetsky2001}
Stan Lipovetsky and Michael Conklin.
\newblock Analysis of regression in game theory approach.
\newblock \emph{Applied Stochastic Models in Business and Industry}, 17\penalty0 (4):\penalty0 319--330, 2001.
\newblock \doi{https://doi.org/10.1002/asmb.446}.
\newblock URL \url{https://onlinelibrary.wiley.com/doi/abs/10.1002/asmb.446}.

\bibitem[Lundberg and Lee(2017)]{lundberg2017}
Scott Lundberg and Su-In Lee.
\newblock A unified approach to interpreting model predictions.
\newblock \emph{Advances in Neural Information Processing Systems}, 30:\penalty0 4766--4775, 2017.
\newblock URL \url{http://arxiv.org/abs/1705.07874}.

\bibitem[Lundberg et~al.(2019{\natexlab{a}})Lundberg, Erion, Chen, DeGrave, Prutkin, Nair, Katz, Himmelfarb, Bansal, and Lee]{lundberg2019}
Scott~M. Lundberg, Gabriel Erion, Hugh Chen, Alex DeGrave, Jordan~M. Prutkin, Bala Nair, Ronit Katz, Jonathan Himmelfarb, Nisha Bansal, and Su-In Lee.
\newblock Explainable ai for trees: From local explanations to global understanding.
\newblock \emph{arXiv:1905.04610 [cs, stat]}, 5 2019{\natexlab{a}}.
\newblock URL \url{http://arxiv.org/abs/1905.04610}.

\bibitem[Lundberg et~al.(2019{\natexlab{b}})Lundberg, Erion, and Lee]{lundberg2019a}
Scott~M. Lundberg, Gabriel~G. Erion, and Su-In Lee.
\newblock Consistent individualized feature attribution for tree ensembles, 2019{\natexlab{b}}.
\newblock URL \url{https://arxiv.org/abs/1802.03888}.

\bibitem[Marichal et~al.(2007)Marichal, Kojadinovic, and Fujimoto]{marichal2007}
Jean-Luc Marichal, Ivan Kojadinovic, and Katsushige Fujimoto.
\newblock Axiomatic characterizations of generalized values.
\newblock \emph{Discrete Applied Mathematics}, 155\penalty0 (1):\penalty0 26--43, 2007.
\newblock ISSN 0166-218X.
\newblock \doi{https://doi.org/10.1016/j.dam.2006.05.002}.
\newblock URL \url{https://www.sciencedirect.com/science/article/pii/S0166218X06002071}.

\bibitem[Merrick and Taly(2020)]{merrick2020}
Luke Merrick and Ankur Taly.
\newblock The explanation game: Explaining machine learning models using shapley values.
\newblock \emph{arXiv:1909.08128 [cs, stat]}, 6 2020.
\newblock URL \url{http://arxiv.org/abs/1909.08128}.

\bibitem[Molnar(2022)]{molnar2022}
Christoph Molnar.
\newblock \emph{Interpretable Machine Learning}.
\newblock 2 edition, 2022.
\newblock URL \url{https://christophm.github.io/interpretable-ml-book}.

\bibitem[Nowak(1997)]{nowak1997}
Andrzej~S. Nowak.
\newblock On an axiomatization of the banzhaf value without the additivity axiom.
\newblock \emph{International Journal of Game Theory}, 26\penalty0 (1):\penalty0 137–141, March 1997.
\newblock ISSN 1432-1270.
\newblock \doi{10.1007/bf01262517}.
\newblock URL \url{http://dx.doi.org/10.1007/BF01262517}.

\bibitem[Owen(2013{\natexlab{a}})]{owen2013}
Art~B. Owen.
\newblock \emph{Monte Carlo theory, methods and examples}.
\newblock \url{https://artowen.su.domains/mc/}, 2013{\natexlab{a}}.

\bibitem[Owen(2013{\natexlab{b}})]{owen2013b}
Art~B. Owen.
\newblock Variance components and generalized sobol' indices.
\newblock \emph{SIAM/ASA Journal on Uncertainty Quantification}, 1\penalty0 (1):\penalty0 19--41, 2013{\natexlab{b}}.
\newblock \doi{10.1137/120876782}.
\newblock URL \url{https://doi.org/10.1137/120876782}.

\bibitem[Owen(2014)]{owen2014}
Art~B. Owen.
\newblock Sobol' indices and shapley value.
\newblock \emph{SIAM/ASA Journal on Uncertainty Quantification}, 2:\penalty0 245--251, 1 2014.
\newblock ISSN 2166-2525.
\newblock \doi{10.1137/130936233}.
\newblock URL \url{http://epubs.siam.org/doi/10.1137/130936233}.

\bibitem[Patel et~al.(2021)Patel, Strobel, and Zick]{patel2021}
Neel Patel, Martin Strobel, and Yair Zick.
\newblock High dimensional model explanations: An axiomatic approach.
\newblock In \emph{Proceedings of the 2021 ACM Conference on Fairness, Accountability, and Transparency}, FAccT '21, page 401–411, New York, NY, USA, 2021. Association for Computing Machinery.
\newblock ISBN 9781450383097.
\newblock \doi{10.1145/3442188.3445903}.
\newblock URL \url{https://doi.org/10.1145/3442188.3445903}.

\bibitem[Pearl(2012)]{pearl2012}
Judea Pearl.
\newblock The do-calculus revisited.
\newblock In \emph{Proceedings of the Twenty-Eighth Conference on Uncertainty in Artificial Intelligence}, UAI'12, page 3–11, Arlington, Virginia, USA, 2012. AUAI Press.
\newblock ISBN 9780974903989.

\bibitem[Penrose(1946)]{penrose1946}
L.~S. Penrose.
\newblock The elementary statistics of majority voting.
\newblock \emph{Journal of the Royal Statistical Society}, 109\penalty0 (1):\penalty0 53--57, 1946.
\newblock ISSN 09528385.
\newblock URL \url{http://www.jstor.org/stable/2981392}.

\bibitem[Ribeiro et~al.(2016)Ribeiro, Singh, and Guestrin]{ribeiro2016}
Marco~Tulio Ribeiro, Sameer Singh, and Carlos Guestrin.
\newblock "why should i trust you?": Explaining the predictions of any classifier.
\newblock In \emph{Proceedings of the 22nd ACM SIGKDD International Conference on Knowledge Discovery and Data Mining}, KDD '16, page 1135–1144, New York, NY, USA, 2016. Association for Computing Machinery.
\newblock ISBN 9781450342322.
\newblock \doi{10.1145/2939672.2939778}.
\newblock URL \url{https://doi.org/10.1145/2939672.2939778}.

\bibitem[Roosen(1995)]{roosen1995}
Charles~Benjamin Roosen.
\newblock \emph{Visualization And Exploration Of High-Dimensional Functions Using The Functional ANOVA Decomposition}.
\newblock PhD thesis, Stanford University, 1995.

\bibitem[Rota(1964)]{rota1964}
Gian-Carlo Rota.
\newblock On the foundations of combinatorial theory i. theory of m\"obius functions.
\newblock \emph{Zeitschrift f\"ur Wahrscheinlichkeitstheorie und Verwandte Gebiete}, 2:\penalty0 340--368, 1964.
\newblock \doi{10.1007/bf00531932}.
\newblock URL \url{http://dx.doi.org/10.1007/bf00531932}.

\bibitem[Rudin(2019)]{rudin2019}
Cynthia Rudin.
\newblock Stop explaining black box machine learning models for high stakes decisions and use interpretable models instead.
\newblock \emph{Nature Machine Intelligence}, 1\penalty0 (5):\penalty0 206–215, May 2019.
\newblock ISSN 2522-5839.
\newblock \doi{10.1038/s42256-019-0048-x}.
\newblock URL \url{http://dx.doi.org/10.1038/s42256-019-0048-x}.

\bibitem[Sculley et~al.(2015)Sculley, Holt, Golovin, Davydov, Phillips, Ebner, Chaudhary, Young, Crespo, and Dennison]{sculley2015}
D.~Sculley, Gary Holt, Daniel Golovin, Eugene Davydov, Todd Phillips, Dietmar Ebner, Vinay Chaudhary, Michael Young, Jean-Fran\c{c}ois Crespo, and Dan Dennison.
\newblock Hidden technical debt in machine learning systems.
\newblock In C.~Cortes, N.~Lawrence, D.~Lee, M.~Sugiyama, and R.~Garnett, editors, \emph{Advances in Neural Information Processing Systems}, volume~28. Curran Associates, Inc., 2015.
\newblock URL \url{https://proceedings.neurips.cc/paper\_files/paper/2015/file/86df7dcfd896fcaf2674f757a2463eba-Paper.pdf}.

\bibitem[Shapley(1953)]{shapley1953}
Lloyd~S Shapley.
\newblock A value for n-person games.
\newblock \emph{Contributions to the Theory of Games}, 2:\penalty0 307--317, 1953.

\bibitem[Shapley and Roth(1988)]{shapley1988}
Lloyd~S. Shapley and Alvin~E. Roth.
\newblock \emph{The Shapley value: essays in honor of Lloyd S. Shapley}.
\newblock Cambridge University Press, 1988.
\newblock ISBN 978-0-521-36177-4.

\bibitem[Sobol\ensuremath{'}(2001)]{sobol2001}
I.M Sobol\ensuremath{'}.
\newblock Global sensitivity indices for nonlinear mathematical models and their monte carlo estimates.
\newblock \emph{Mathematics and Computers in Simulation}, 55:\penalty0 271--280, 2 2001.
\newblock ISSN 03784754.
\newblock \doi{10.1016/S0378-4754(00)00270-6}.
\newblock URL \url{https://linkinghub.elsevier.com/retrieve/pii/S0378475400002706}.

\bibitem[Song et~al.(2016)Song, Nelson, and Staum]{song2016}
Eunhye Song, Barry~L. Nelson, and Jeremy Staum.
\newblock Shapley effects for global sensitivity analysis: Theory and computation.
\newblock \emph{SIAM/ASA Journal on Uncertainty Quantification}, 4:\penalty0 1060--1083, 1 2016.
\newblock \doi{10.1137/15M1048070}.
\newblock URL \url{https://epubs.siam.org/doi/abs/10.1137/15M1048070}.

\bibitem[Strobl et~al.(2008)Strobl, Boulesteix, Kneib, Augustin, and Zeileis]{strobl2008}
Carolin Strobl, Anne-Laure Boulesteix, Thomas Kneib, Thomas Augustin, and Achim Zeileis.
\newblock Conditional variable importance for random forests.
\newblock \emph{BMC Bioinformatics}, 9\penalty0 (1), July 2008.
\newblock ISSN 1471-2105.
\newblock \doi{10.1186/1471-2105-9-307}.
\newblock URL \url{http://dx.doi.org/10.1186/1471-2105-9-307}.

\bibitem[Sundararajan and Najmi(2020)]{sundararajan2020}
Mukund Sundararajan and Amir Najmi.
\newblock The many shapley values for model explanation.
\newblock In Hal~Daumé III and Aarti Singh, editors, \emph{Proceedings of the 37th International Conference on Machine Learning}, volume 119 of \emph{Proceedings of Machine Learning Research}, pages 9269--9278. PMLR, 13--18 Jul 2020.
\newblock URL \url{https://proceedings.mlr.press/v119/sundararajan20b.html}.

\bibitem[Sundararajan et~al.(2020)Sundararajan, Dhamdhere, and Agarwal]{sundararajan2020a}
Mukund Sundararajan, Kedar Dhamdhere, and Ashish Agarwal.
\newblock The shapley taylor interaction index.
\newblock In Hal~Daumé III and Aarti Singh, editors, \emph{Proceedings of the 37th International Conference on Machine Learning}, volume 119 of \emph{Proceedings of Machine Learning Research}, pages 9259--9268. PMLR, 13--18 Jul 2020.
\newblock URL \url{https://proceedings.mlr.press/v119/sundararajan20a.html}.

\bibitem[Tomsett et~al.(2020)Tomsett, Harborne, Chakraborty, Gurram, and Preece]{tomsett2020}
Richard Tomsett, Dan Harborne, Supriyo Chakraborty, Prudhvi Gurram, and Alun Preece.
\newblock Sanity checks for saliency metrics.
\newblock \emph{Proceedings of the AAAI conference on artificial intelligence}, 34, 2020.
\newblock URL \url{www.aaai.org}.

\bibitem[Tonekaboni et~al.(2019)Tonekaboni, Joshi, McCradden, and Goldenberg]{tonekaboni2019}
Sana Tonekaboni, Shalmali Joshi, Melissa~D. McCradden, and Anna Goldenberg.
\newblock What clinicians want: Contextualizing explainable machine learning for clinical end use.
\newblock In Finale Doshi-Velez, Jim Fackler, Ken Jung, David Kale, Rajesh Ranganath, Byron Wallace, and Jenna Wiens, editors, \emph{Proceedings of the 4th Machine Learning for Healthcare Conference}, volume 106 of \emph{Proceedings of Machine Learning Research}, pages 359--380. PMLR, 09--10 Aug 2019.
\newblock URL \url{https://proceedings.mlr.press/v106/tonekaboni19a.html}.

\bibitem[Tsai et~al.(2023)Tsai, Yeh, and Ravikumar]{tsai2023}
Che-Ping Tsai, Chih-Kuan Yeh, and Pradeep Ravikumar.
\newblock Faith-shap: The faithful shapley interaction index.
\newblock \emph{Journal of Machine Learning Research}, 24\penalty0 (94):\penalty0 1--42, 2023.
\newblock URL \url{http://jmlr.org/papers/v24/22-0202.html}.

\bibitem[\v{S}trumbelj et~al.(2009)\v{S}trumbelj, Kononenko, and {Robnik \v{S}ikonja}]{strumbelj2009}
E.~\v{S}trumbelj, I.~Kononenko, and M.~{Robnik \v{S}ikonja}.
\newblock Explaining instance classifications with interactions of subsets of feature values.
\newblock \emph{Data \& Knowledge Engineering}, 68\penalty0 (10):\penalty0 886--904, 2009.
\newblock ISSN 0169-023X.
\newblock \doi{https://doi.org/10.1016/j.datak.2009.01.004}.
\newblock URL \url{https://www.sciencedirect.com/science/article/pii/S0169023X09000056}.

\bibitem[\v{S}trumbelj and Kononenko(2014)]{strumbelj2014}
Erik \v{S}trumbelj and Igor Kononenko.
\newblock Explaining prediction models and individual predictions with feature contributions.
\newblock \emph{Knowledge and Information Systems}, 41:\penalty0 647--665, 12 2014.
\newblock ISSN 0219-1377, 0219-3116.
\newblock \doi{10.1007/s10115-013-0679-x}.
\newblock URL \url{http://link.springer.com/10.1007/s10115-013-0679-x}.

\bibitem[\v{S}trumbelj and Kononenko(2010)]{strumbelj2010}
Erik \v{S}trumbelj, Erik and Igor Kononenko.
\newblock An efficient explanation of individual classifications using game theory.
\newblock \emph{Journal of Machine Learning Research}, 11:\penalty0 1--18, 2010.
\newblock ISSN 1533-7928.
\newblock URL \url{http://jmlr.org/papers/v11/strumbelj10a.html}.

\bibitem[Weber(1988)]{weber1988}
Robert~J Weber.
\newblock Probabilistic values for games.
\newblock \emph{The Shapley Value. Essays in Honor of Lloyd S. Shapley}, page 101\textendash{}119, 1988.

\bibitem[Williamson and Feng(2020)]{williamson2020}
Brian Williamson and Jean Feng.
\newblock Efficient nonparametric statistical inference on population feature importance using shapley values.
\newblock In Hal~Daum\'{e} III and Aarti Singh, editors, \emph{Proceedings of the 37th International Conference on Machine Learning}, volume 119 of \emph{Proceedings of Machine Learning Research}, pages 10282--10291. PMLR, 13--18 Jul 2020.
\newblock URL \url{https://proceedings.mlr.press/v119/williamson20a.html}.

\bibitem[Yeh et~al.(2019)Yeh, Hsieh, Suggala, Inouye, and Ravikumar]{yeh2019}
Chih-Kuan Yeh, Cheng-Yu Hsieh, Arun~Sai Suggala, David~I. Inouye, and Pradeep Ravikumar.
\newblock On the (in)fidelity and sensitivity for explanations.
\newblock \emph{arXiv preprint arXiv:1901.09392}, 1 2019.
\newblock URL \url{http://arxiv.org/abs/1901.09392}.

\bibitem[Yeh et~al.(2022)Yeh, Lee, Liu, and Ravikumar]{yeh2022}
Chih-Kuan Yeh, Kuan-Yun Lee, Frederick Liu, and Pradeep Ravikumar.
\newblock Threading the needle of on and off-manifold value functions for shapley explanations.
\newblock \emph{arXiv:2202.11919 [cs]}, 2 2022.
\newblock URL \url{http://arxiv.org/abs/2202.11919}.

\bibitem[Zeiler and Fergus(2014)]{zeiler2014}
Matthew~D. Zeiler and Rob Fergus.
\newblock Visualizing and understanding convolutional networks.
\newblock In David Fleet, Tomas Pajdla, Bernt Schiele, and Tinne Tuytelaars, editors, \emph{Computer Vision -- ECCV 2014}, pages 818--833, Cham, 2014. Springer International Publishing.
\newblock ISBN 978-3-319-10590-1.

\end{thebibliography}

\end{document}